\documentclass[runningheads]{llncs}
\usepackage{graphicx}
\usepackage{amsmath,amssymb} % define this before the line numbering.
\usepackage{color}
\usepackage{enumerate}
\usepackage{subfigure}
\usepackage{booktabs}
\usepackage{enumitem}
\usepackage{array}
\usepackage[width=122mm,left=12mm,paperwidth=146mm,height=193mm,top=12mm,paperheight=217mm]{geometry}

\newcommand{\etal}{\textit{et al}.}
\newcommand{\ie}{\textit{i}.\textit{e}.}
\newcommand{\eg}{\textit{e}.\textit{g}.}
\newcommand{\etc}{\textit{etc}}

\begin{document}
% \linenumbers
\pagestyle{headings}
\mainmatter

\title{Stroke Controllable Fast Style Transfer with Adaptive Receptive Fields\thanks{Project page: \url{http://yongchengjing.com/StrokeControllable}}}

\titlerunning{Stroke Controllable Fast Style Transfer}

\authorrunning{Y. Jing et al.}

\author{Yongcheng Jing\inst{1,2} \and Yang Liu\inst{1,2} \and Yezhou Yang\inst{3} \and Zunlei Feng\inst{1,2}\and \\ Yizhou Yu\inst{4} \and Dacheng Tao\inst{5} \and Mingli Song\inst{1,2}}

% ORCID of Mingli Song: \orcidID{0000-0003-2621-6048}

%Please write out author names in full in the paper, i.e. full given and family names.
%If any authors have names that can be parsed into FirstName LastName in multiple ways, please include the correct parsing, in a comment to the volume editors:
%\index{Lastnames, Firstnames}
%(Do not uncomment it, because you may introduce extra index items if you do that, we will use scripts for introducing index entries...)

\institute{College of Computer Science and Technology, Zhejiang University, Hangzhou, China \\
\and Alibaba-Zhejiang University Joint Institute of Frontier Technologies, China \\
%\email{\{ycjing, lyng\_95, zunleifeng, brooksong\}@zju.edu.cn}\\
\and Arizona State University, Tempe, USA \\
%\email{yz.yang@asu.edu}\\
\and Deepwise AI Lab, Beijing, China \\
%\email{yizhouy@acm.org}\\
\and UBTECH Sydney AI Centre, SIT, FEIT, University of Sydney, Australia \\
%\email{dacheng.tao@sydney.edu.au}
}

\maketitle

\begin{abstract}

The Fast Style Transfer methods have been recently proposed to transfer a photograph to an artistic style in real-time. This task involves controlling the stroke size in the stylized results, which remains an open challenge. In this paper, we present a stroke controllable style transfer network that can achieve continuous and spatial stroke size control. By analyzing the factors that influence the stroke size, we propose to explicitly account for the receptive field and the style image scales. We propose a StrokePyramid module to endow the network with adaptive receptive fields, and two training strategies to achieve faster convergence and augment new stroke sizes upon a trained model respectively. By combining the proposed runtime control strategies, our network can achieve continuous changes in stroke sizes and produce distinct stroke sizes in different spatial regions within the same output image.

%our network can produce continuously varied stroke sizes in different output images or different spatial regions within the same output image.

\end{abstract}

%%%%%%%%% BODY TEXT
\section{Introduction}

Rendering a photograph with a given artwork style has been a long-standing research topic \cite{Gooch2001non,hertzmann2001image,rosin2012image,strothotte2002non}.
Conventionally, the task of style transfer is usually studied as a generalization of texture synthesis \cite{efros2001image,elad2017style,frigo2016split}.
Based on the recent progress in visual texture modelling \cite{gatys2015texture},
Gatys \textit{et al}. firstly propose an algorithm that exploits Convolutional Neural Network (CNN) to recombine the content of a given photograph and the style of an artwork, and reconstruct a visually plausible stylized image,
known as the process of Neural Style Transfer \cite{gatys2016image}.
Since the seminal work of Gatys \etal, Neural Style Transfer has been attracting wide attention from both academia and industry \cite{d2017coherent,chen2018stereoscopic,li2016combining,li2017demystifying,prisma}. However, the algorithm of Gatys \textit{et al}. is based on iterative image optimizations and leads to a slow optimization process for each pair of content and style.
To tackle this issue, several algorithms have been proposed to speed up the style transfer process, called
the Fast Style Transfer in the literature \cite{gatys2016controlling,lu2017decoder}.

\newcommand\y{2.8cm}

\begin{figure}[!t]
\setlength\tabcolsep{0.8 pt}
{\renewcommand{\arraystretch}{1}
%\begin{tabular}{>{\centering}n{\p} >{\centering}n{\p} >{\centering\arraybackslash}n{\p}}
%\begin{tabular}{>{\centering}m{2.34cm} >{\centering}m{2.34cm} >{\centering}m{2.34cm} >{\centering}m{2.34cm} >{\centering\arraybackslash}m{2.34cm}}
\begin{tabular}{ccccc}
\centering

\includegraphics[width=0.195\textwidth]{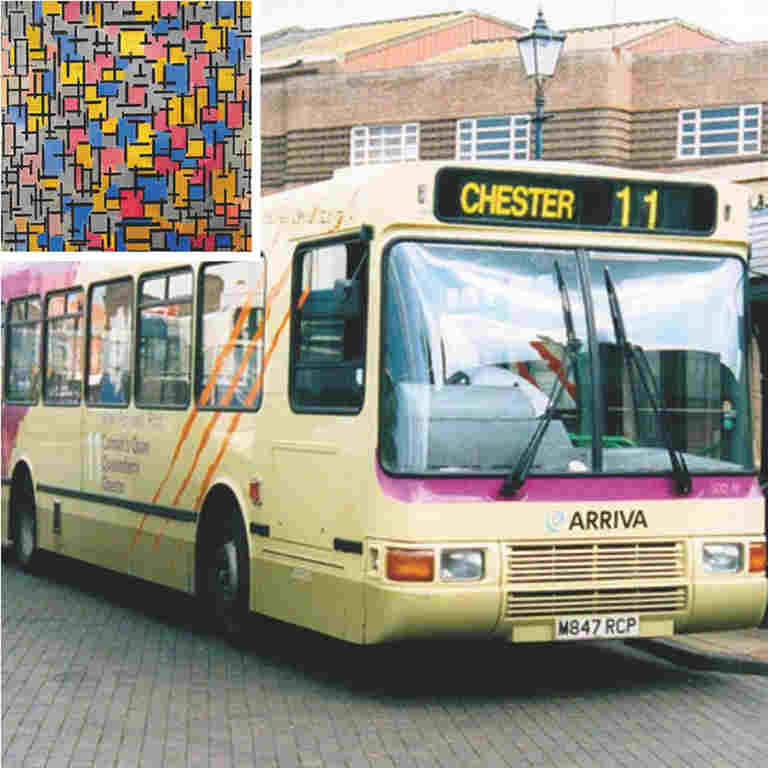}&\includegraphics[width=0.195\textwidth]{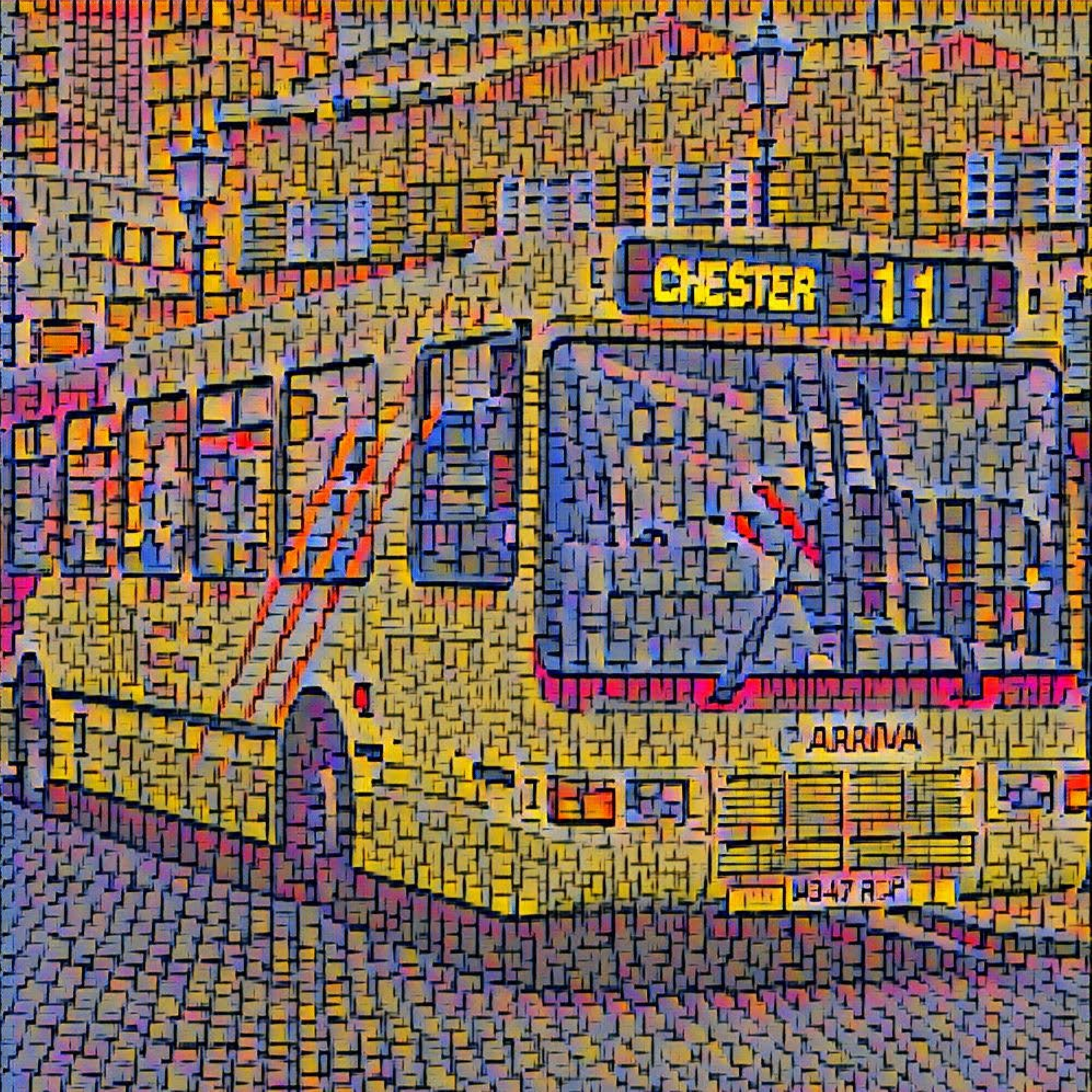}  & \includegraphics[width=0.195\textwidth]{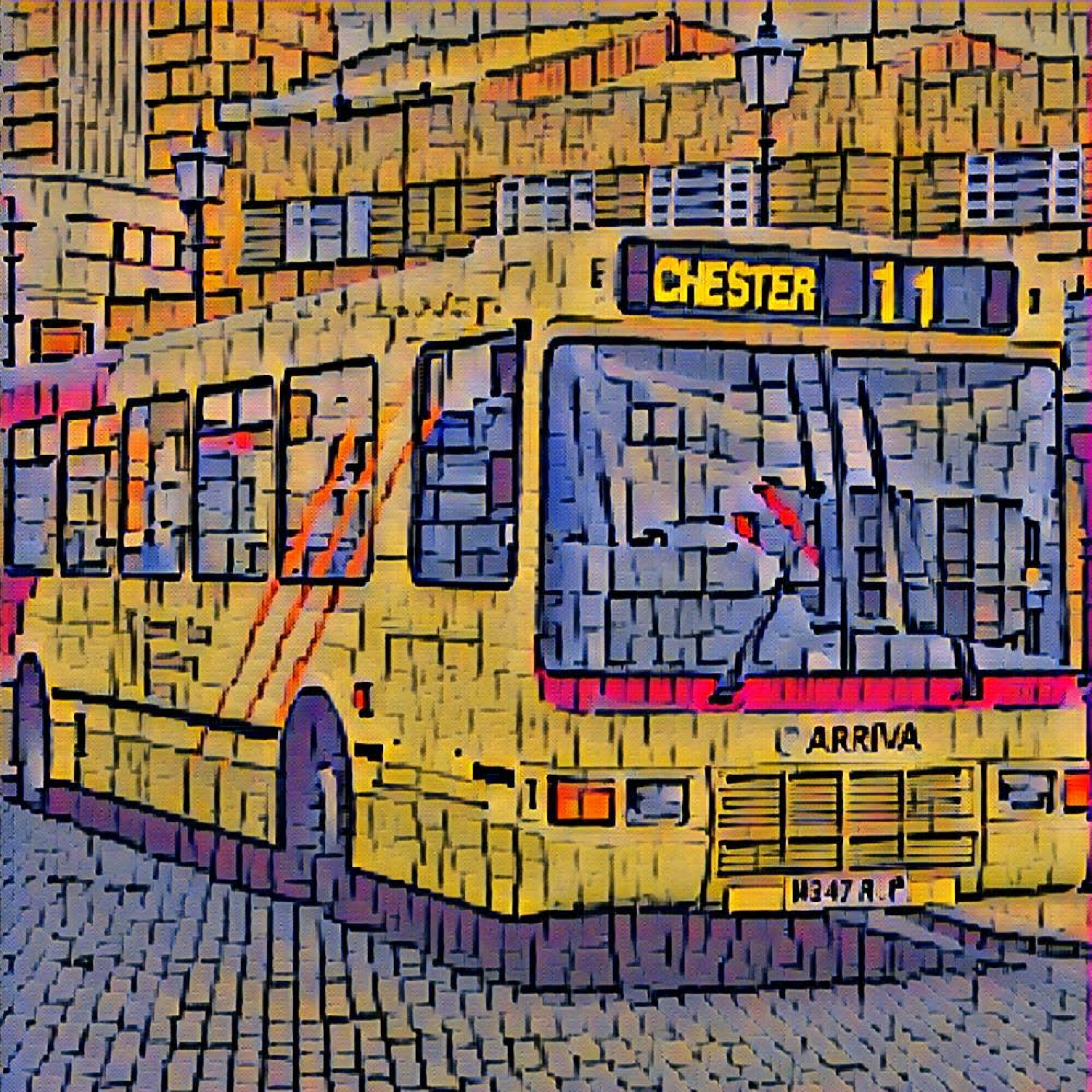} & \includegraphics[width=0.195\textwidth]{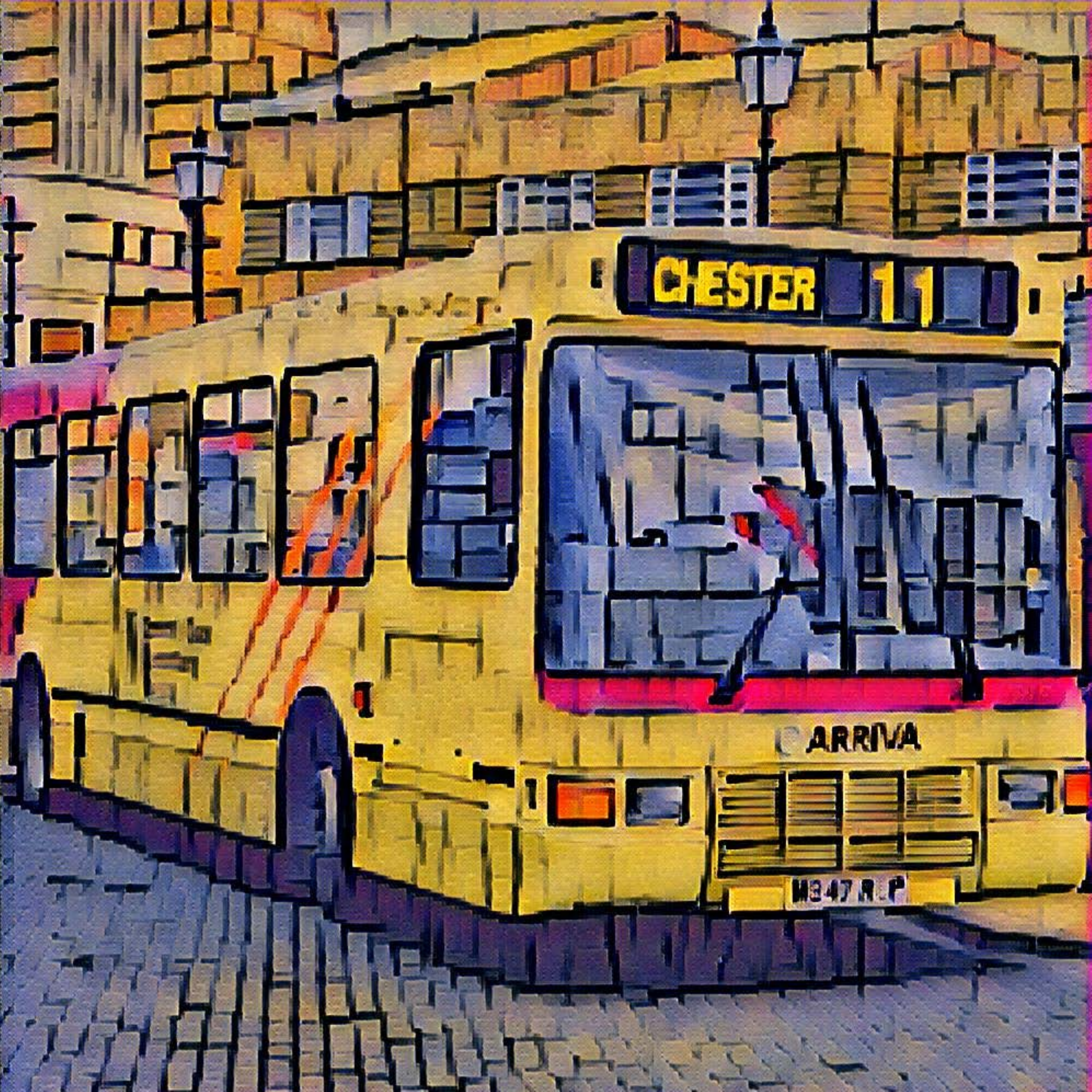} & \includegraphics[width=0.195\textwidth]{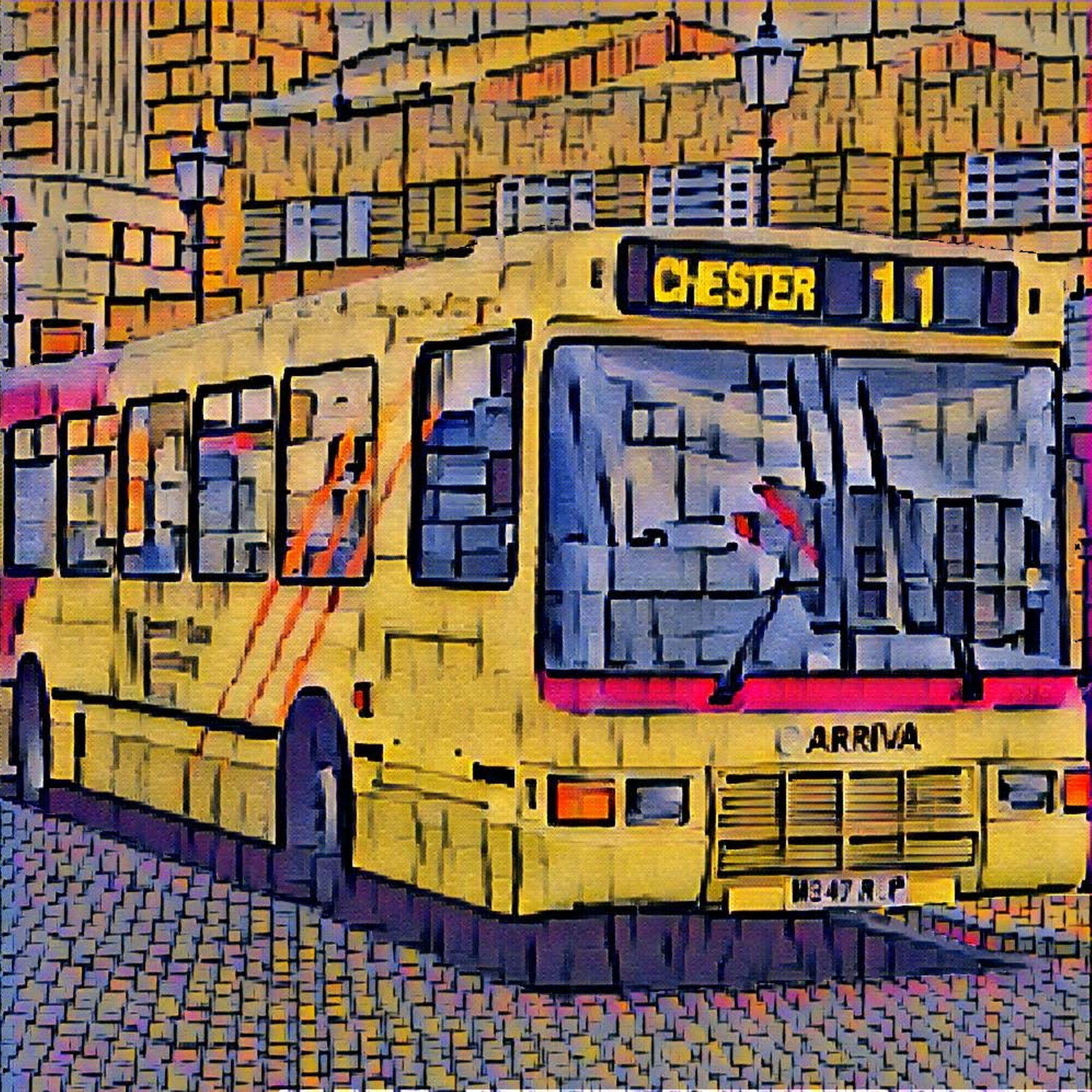}\\

\scriptsize{(a) Content\&Style}&\scriptsize{(b) Stroke Size \#1} & \scriptsize{(c) Stroke Size \#2}&\scriptsize{(d) Stroke Size \#3} & \scriptsize{(e) Mixed Strokes}
\end{tabular}
}
%\smallskip
\caption{Stylized results with different stroke sizes. All these results are produced by one single model in real-time using our proposed algorithm. }
%\myLcomment{draw the difference graph}
%We did not do any image pre- or post-processing before and after forwarding.
\label{fig:example} %% label for entire figure
\end{figure}
%\floatsep{1}

%However, the PSPM needs to train a separate model for each style, which loses the flexibility.

The current Fast Style Transfer approaches can be categorized into three classes,
Per-Style-Per-Model (PSPM) \cite{ulyanov2016texture,ulyanov2017improved,Johnson2016perceptual,li2016precomputed}, Multiple-Style-Per-Model (MSPM) \cite{zhang2017multi,dumoulin2016learned,li2017diverse,chen2017stylebank}, and Arbitrary-Style-Per-Model (ASPM)  \cite{huang2017arbitrary,li2017universal}.
The gist of  PSPM is to train a feed-forward style-specific generator and to produce
a corresponding stylized result with a forward pass.  MSPM improves the efficiency
by further incorporating multiple styles into one single generator.  ASPM aims at
transferring an arbitrary style through only one single model.
%, the quality is slightly behind others in some cases \cite{huang2017real}.  %\textcolor{red}{Regarding the quality, the PSPM is usually regarded as the gold-standard method for Fast Style Transfer.Thus, we build our system upon this category.}
% * <brooksong@ieee.org> 2017-11-09T04:55:04.901Z:
%
% > Thus, we build our system upon this category.
% 我觉得这话放在这里没啥用，可以删掉。如果后面需要的话，可以放在需要的地方。
%
% ^ <398372213@qq.com> 2017-11-12T01:06:37.931Z:
%
% 已去掉
%
% ^ <brooksong@ieee.org> 2017-11-13T01:43:32.849Z.

There is a trade-off between efficiency and quality
for all such  Fast Style Transfer algorithms \cite{huang2017arbitrary,li2017universal}.
In terms of quality, PSPM is usually regarded to produce more appealing stylized results \cite{ulyanov2017improved,huang2017arbitrary}.
However, PSPM is not flexible in terms of controlling perceptual factors (\eg, style-content tradeoff, color control, spatial control). Among these perceptual factors, strokes are one of the most important geometric primitives to characterize an artwork,
as shown in Figure~\ref{fig:example}. In reality, for the same texture, different artists have their own way to place different sizes of strokes as a reflection of their unique ``styles'' (\eg, Monet and Pollock).
To achieve different stroke sizes with PSPM, one possible solution is to train multiple models,
which is time and space consuming. Another solution is to resize the input image to different scales,
which will inevitably hurt the quality of stylization. None of these solutions, however, can
achieve continuous stroke size control or produce distinct stroke sizes in different spatial regions
without trading off quality and efficiency.
%To achieve different stroke sizes, one possible solution is to re-train a new model with a suitable scale of style image for each desired stroke size. However, it is time and space consuming to train separate models for different stroke sizes. Another solution is to first resize the content image according to the desired size of the brush stroke. After achieving the stylized result of the same size, we resize the result back to the size of the original content image. Due to the upsampling and downsampling process, the quality of the stylized result will inevitably decrease.
%Extra post-processing (\eg super-resolution) will bring additional time cost and affect the stylization efficiency.

%Another stroke related problem is that stylized result produced by a single feed-forward pass seems to have the same stroke size across the whole image.
%achieve continuous and spatial stroke size control without trading off quality and efficiency.

In this paper, we propose a stroke controllable Fast Style Transfer algorithm that can incorporate multiple stroke sizes into one single model and achieves flexible continuous stroke size control and spatial stroke size control.
By analyzing the factors that influence the stroke size in stylized results,
we propose to explicitly account for both the receptive field and the style image scale.
To this end, we propose a \emph{StrokePyramid} module to endow the network with adaptive
receptive fields and different stroke sizes are learned with different receptive fields.
We then introduce a progressive training strategy to make the network converge faster
and an incremental training strategy to learn new stroke sizes upon a trained model.
By combining two proposed runtime control techniques which are continuous stroke size
control and spatial stroke size control, our network can produce distinct stroke sizes in different
outputs or different spatial regions within the same output image.

In summary, our work has three primary contributions: 1) We analyze the factors that influence the stroke size in stylized results,
and  propose that both the receptive field and the style image scale should be considered for stroke size control in most cases.
2) We propose a stroke controllable style transfer network and two corresponding training strategies in order to achieve faster convergence and augment new stroke sizes upon a trained model respectively.
3) We present two runtime control strategies to empower our single model with the ability of producing continuous changes in stroke size and distinct stroke sizes in different spatial regions within the same output image.
To the best of our knowledge, this is the first style transfer network that  achieves continuous stroke size control and spatial stroke size control.

\section{Related Work}
\label{sect:relatedwork}
We briefly review here perceptual factors in Fast Style Transfer as well as the involving regulating receptive field in neural networks.

\textbf{Controlling perceptual factors in Fast Style Transfer.} Stroke size control belongs to the domain of controlling perceptual factors during stylization. In this field, several significant works are recently presented. However, there are few efforts devoted to controlling stroke size during Fast Style Transfer. In \cite{gatys2016controlling}, Gatys \textit{et al}. study the color control and spatial control for Fast Style Transfer. Lu \textit{et al}. further extend Gatys \etal's work to meaningful spatial control by incorporating semantic content, achieving the so-called Fast Semantic Style Transfer \cite{lu2017decoder}. Another related work is Wang \etal's algorithm which aims to learn large brush strokes for high-resolution images \cite{wang2016multimodal}. They find that current Fast Style Transfer algorithms fail to paint large strokes in high-resolution images and propose a coarse-to-fine architecture to solve this problem. Note that the work in \cite{wang2016multimodal} is intrinsically different from this paper as one single pre-trained model in \cite{wang2016multimodal} still produces one stroke size for the same input image. A concurrent work in \cite{zhang2017multi} also explores the issue of stroke size control. Compared with \cite{zhang2017multi}, our work has the benefits of flexible continuous and spatial stroke size control.
% * <brooksong@ieee.org> 2017-11-09T06:14:52.376Z:
%
% > the studied problem
%
% 有些vague
%
% ^ <398372213@qq.com> 2017-11-12T02:22:51.987Z:
%
% 已修改。
%
% ^ <brooksong@ieee.org> 2017-11-13T01:43:44.573Z.

%Note that we do not study the situation when the test image has very high resolution but we believe our proposed algorithm can act as an embedded module in \cite{wang2016multimodal}.

\textbf{Regulating receptive field in neural networks.} The receptive field is one of the basic concepts in convolutional neural networks, which refers to a region of the input image that one neuron is responsive to. It can affect the performance of the networks and becomes a critical issue in many tasks (\eg, semantic segmentation \cite{zhang2018context}, image parsing). To regulate the receptive field, \cite{yu2015multi} proposes the operation of dilated convolution (also called atrous convolution in \cite{chen2017deeplab}), which supports the expansion of receptive field by setting different dilation values and is widely used in many generation tasks like \cite{he2018deep,fan2018decouple}. Another work in \cite{dai2017deformable} further proposes a deformable convolution which augments the sampling locations in regular convolution with additional offsets. Furthermore, Wei \textit{et al}. \cite{wei2017learning} propose a learning-based receptive field regulating method which is to inflate or shrink feature maps automatically.

\section{Pre-analysis}
\label{sect:analysis}
% * <398372213@qq.com> 2017-11-12T01:13:24.747Z:
%
% > Pre-analysis
% 这个章节是根据您之前的architecture那一节中的comment加的，我也同时在您那里的comment中做了response。文章总体逻辑为 1. 首先解释笔触和哪些因素有关，笔触和感受野大小是有关的（文章第一个揭示的，之前工作中没有，所以是文章中一个贡献），并用实验证明，对原因进行分析。 2. 根据这一影响因素，得出用可变感受野来学习多笔触（大感受野学习大笔触，小感受野学习小笔触）会让实验结果更好的结论，由此设计一个拥有可变感受野的网络架构。所以黎叔，我觉得这一部分还是比较重要的。之前和architecture混在了一起，您那里给出的comment放在intro或实验。试了下放在intro会让intro过长而且看起来混入了文章细节，放在实验可能看到architecture那一节会让人觉得很疑惑为什么这样设计。所以和宋师兄和雷师兄讨论后，我仿照禽兽哥"Semi-supervised Node Splitting for Random Forest Construction"这篇和我逻辑有些像的cvpr文章，在architecture之前加入了一个“Pre-analysis”。
%
% ^ <brooksong@ieee.org> 2017-11-13T01:42:57.612Z.

We start by reviewing the concept of the stroke size. Consider an image in style transfer as a composition of a series of small stroke textons, which are referred as the fundamental geometric micro-structures in images \cite{julesz1981textons,zhu2005textons}. The stroke size of an image can be defined as the average scale of the composed stroke textons.

In the deep neural network based Fast Style Transfer, three factors are found to influence the stroke size, namely the scale of the style image \cite{wang2016multimodal}, the receptive field in the loss network \cite{gatys2016controlling}, and the receptive field in the generative network.
%We will firstly explain the first two factors and then focus our effort on the third one.
% * <398372213@qq.com> 2017-11-12T03:05:35.207Z:
%
% > there are three affecting factors
% 三个因素中，第一个之前文章有表达过类似的意思（有篇文章中说拿256style图训练1024内容图测试效果会很差）但没有直接指出笔触变化，而且也没有分析原因。所以我这里写的稍微多一点，分析清楚一些。然后第二个因素有文章详细讨论过，所以一句话简单提一下给个reference。然后第三个因素是这篇文章第一个指出来的，所以篇幅最长，作为后面提出的架构的inspiration。
%
% ^ <brooksong@ieee.org> 2017-11-13T01:43:00.512Z.

The objective style is usually learned by matching the style image's gram-based statistics \cite{gatys2016image} in style transfer algorithms, which are computed over the feature maps from the pre-trained VGG network \cite{simonyan2014very}. These gram-based statistics are scale-sensitive, \ie, they contain the scale information of the given style image.
One reason for this characteristic is that the VGG features vary with the image scale.
We also find that for other style statistics (\eg, BN-based statistics in \cite{li2017demystifying}),
it reaches the same conclusion. Therefore, given the same content image, generative networks
trained with different scales of the style image can produce different stroke sizes.

% * <brooksong@ieee.org> 2017-11-13T02:37:28.075Z:
%
% 再起一段，每段一个factor
%
% ^ <398372213@qq.com> 2017-11-13T03:36:08.494Z:
%
% 已修改。
%
% ^ <brooksong@ieee.org> 2017-11-13T03:56:34.983Z.

Although the stroke in stylized results usually becomes larger with the increase of the style image scale, this is infeasible when the style image is scaled to a high resolution (\eg, $3000 \times 3000$ pixels \cite{gatys2016controlling}). The reason for this problem is that a neuron in pre-trained VGG loss network can only affect a region with the receptive field size in the input image. When the stroke texton is much larger than the fixed receptive field in VGG loss network, there is no visual difference between a large and larger stroke texton in a relatively small region.
% * <brooksong@ieee.org> 2017-11-13T02:32:09.869Z:
%
% > It is because in style transfer algorithms, the objective style is usually learned by matching the style image's gram-based statistics, which are computed over the feature maps from the pre-trained VGG network \cite{simonyan2014very}.
% 这个it is because是对应上一句的还是下一句的？跟前面还是跟后面是因果关系？
%
% ^ <398372213@qq.com> 2017-11-13T03:38:20.714Z:
%
% it is because 和前面是因果关系，解释前面，然后it is because后面跟的两句话都是解释的，因为放在一句话的话会非常长。
%
% ^ <brooksong@ieee.org> 2017-11-13T03:56:23.137Z:
%
% 我加了for this problem，这样逻辑勉强能连上。
%
% ^ <398372213@qq.com> 2017-11-13T06:10:33.040Z:
%
% 黎叔，您开始的comment说的it is because in style transfer algorithms, the objective style is usually .... 这一段，我把第一句话挪到最后了，然后把it is because 去掉了。读起来逻辑可能顺一些，想麻烦黎叔再帮我看一下。
%
% ^.
\newcolumntype{?}[1]{!{\vrule width #1}}
\begin{figure}[!t]
\setlength\tabcolsep{1 pt}
{\renewcommand{\arraystretch}{0.8}
%\begin{tabular}{>{\centering}n{\p} >{\centering}n{\p} >{\centering\arraybackslash}n{\p}}
%\begin{tabular}{>{\centering}m{1.98cm} >{\centering}m{1.98cm} >{\centering}m{1.98cm} ?{0.2mm} >{\centering}m{1.98cm} >{\centering}m{1.98cm} >{\centering\arraybackslash}m{1.98cm}}
\begin{tabular}{ccc ?{0.2mm} ccc}
\centering

\includegraphics[width=0.16\textwidth]{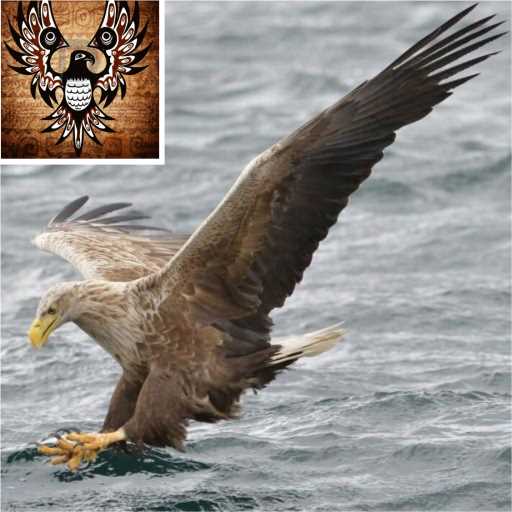}& \includegraphics[width=0.16\textwidth]{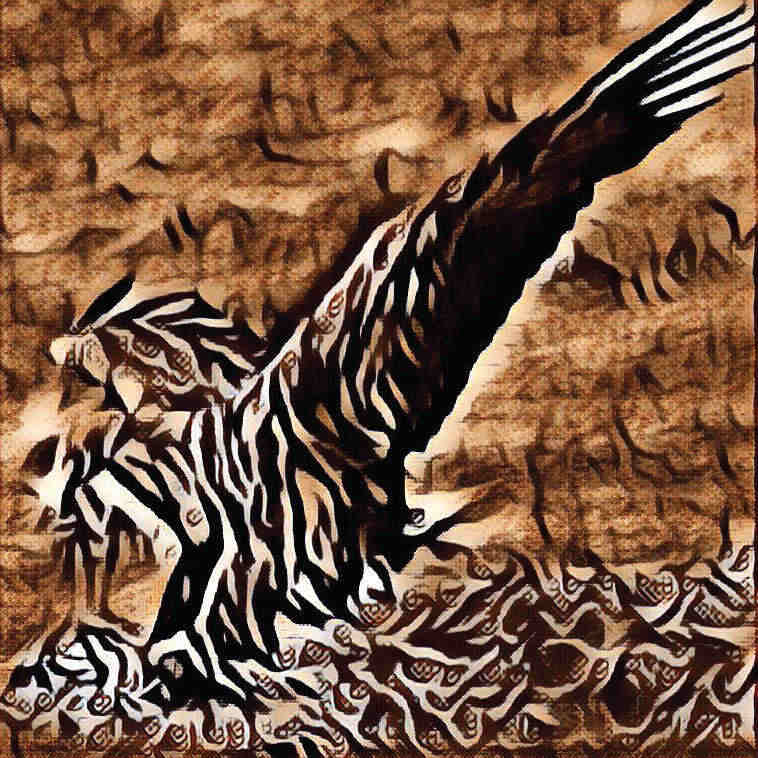} & \includegraphics[width=0.16\textwidth]{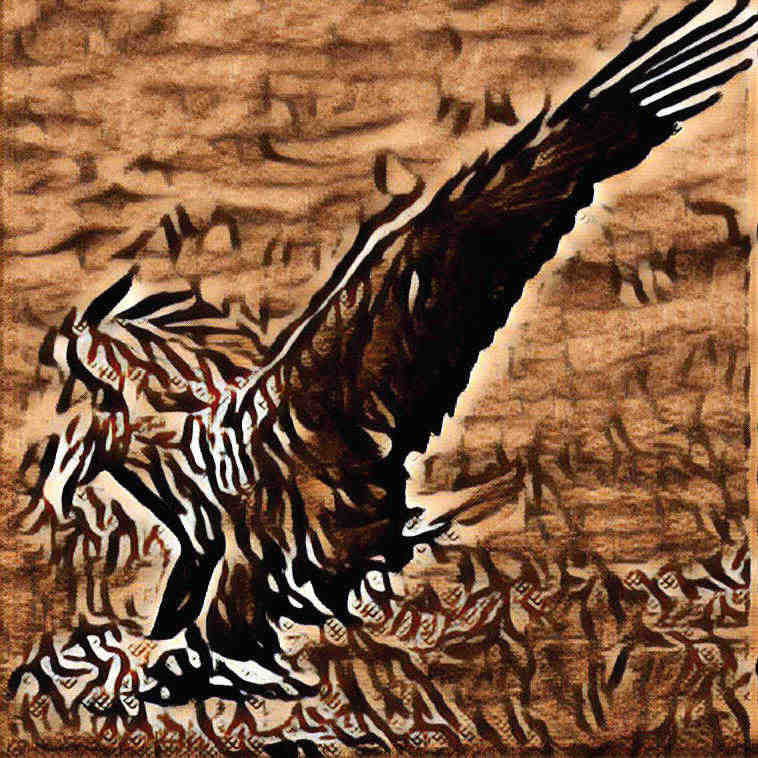} &\includegraphics[width=0.16\textwidth]{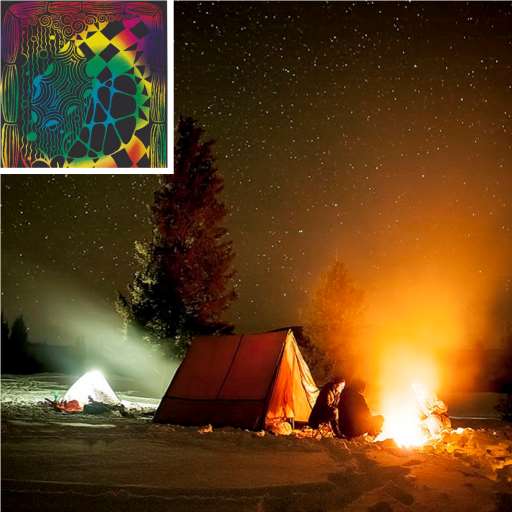}&\includegraphics[width=0.16\textwidth]{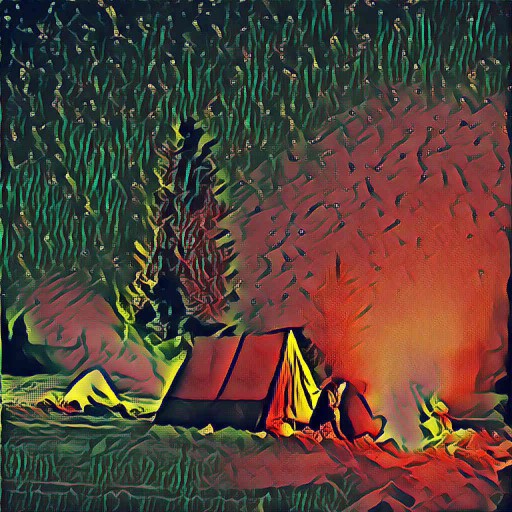}&\includegraphics[width=0.16\textwidth]{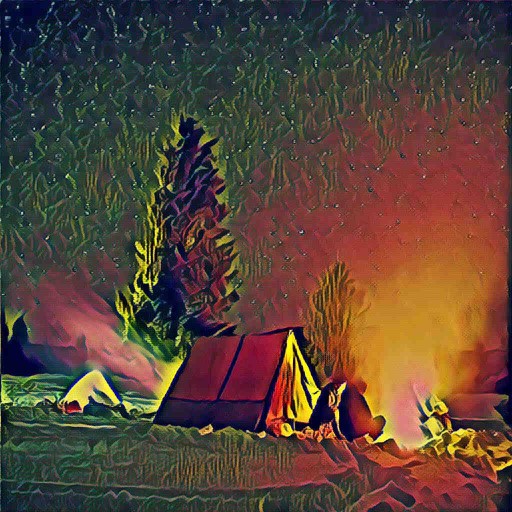}\\
%\smallskip
\scriptsize{Content \& Style}&\scriptsize{LRF Result} & \scriptsize{SRF Result} & \scriptsize{Content \& Style} & \scriptsize{LRF Result} & \scriptsize{SRF Result}
\end{tabular}
}
%\vspace{-0.18cm}
%\vspace{-0.01cm}
\caption{Results of learning the same size of large strokes with large and small receptive fields, respectively. LRF represents the result produced with a large receptive field and SRF represents the result produced with a small receptive field. Content images are credited to flickr users \emph{Kevin Robson} and \emph{b togol}.}
\label{fig:analysis} %% label for entire figure
\end{figure}

\begin{figure}[!t]
  \centering
    %\label{fig:subfig:a} %% label for first subfigure
    \includegraphics[width=0.8\textwidth]{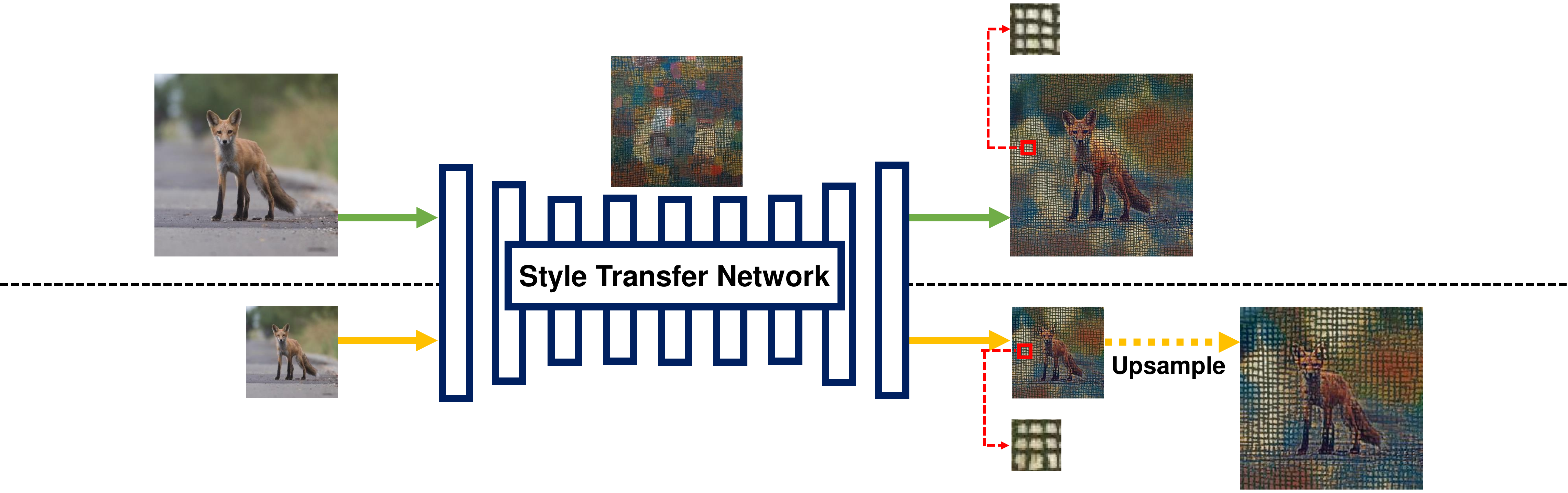}
  \caption{The feed-forward process of Fast Style Transfer. For the same size of regions in the outputs of both small and large input images respectively, their stroke sizes are almost the same. The content image is credited to flickr user \emph{BillChenSF}.}
  \label{fig:explain} %% label for entire figure
\end{figure}

Apart from these above two factors, we further find that the receptive field size in the generative network also has influence on the stroke size. In Figure~\ref{fig:analysis}, we change the receptive field size in the generative network and other factors remain the same. It is noticeable that a larger stroke size is produced with a larger receptive field for some styles. To explain this result, we interpret the training process of the generative network as teaching the convolutional kernels to paint a pre-defined size of stroke textons in each region with the size of receptive field. Therefore, given two different sizes of input images, the kernels of a trained network paint almost the same size of stroke textons in the same size of regions, as shown in Figure~\ref{fig:explain}. In particular, when the receptive field in a generative network is smaller than the stroke texton, the kernels can only learn to paint a part of the whole stroke texton in each region, which influences the stroke size. Hence, for a large stroke size, the network needs larger receptive fields to learn the global stroke configuration. For a small stroke size, the network only needs to learn local features.

To sum up, both the scale of the style image and the receptive field in the generative network should generally be considered for stroke size control. As the style image is not high-resolution in most cases, the influence of the receptive field in the loss network is not considered in this work.

%the influence of the receptive field in VGG loss network is not considered in this work.

%Figure~\ref{fig:explain} verifies our hypothesis by demonstrating that given the same size of regions in the results of both small and large input images, the kernels of a trained network paint almost the same size of stroke textons.
% * <brooksong@ieee.org> 2017-11-13T02:44:32.533Z:
%
% >  The results are produced when the network converges.
% 没有信息，删掉
%
% ^ <398372213@qq.com> 2017-11-13T03:14:33.681Z:
%
% 已删掉。
%
% ^ <brooksong@ieee.org> 2017-11-13T03:43:09.391Z.
% * <brooksong@ieee.org> 2017-11-13T02:41:45.772Z:
%
% > Except for these two factors
% 用except还是besides，还是apart from?
%
% ^ <398372213@qq.com> 2017-11-13T03:28:17.102Z:
%
% 改成了apart from。
%
% ^ <brooksong@ieee.org> 2017-11-13T03:43:04.778Z.

%(pre-trained on the ImageNet dataset \cite{russakovsky2015imagenet})
 %is usually learned by matching the second-order feature statistics of the style image \cite{gatys2016image}, which is obtained by computing Gram matrix over the feature maps from the pre-trained VGG network \cite{simonyan2014very}

%%%%%
%%%%%

\section{Proposed Approach}
\label{sect:network}

\subsection{Problem Formulation}

Assume that $\mathcal{T}_i \in \mathbb{T}$ denotes the stroke size of an image, $\mathbb{T}$ denotes the set of all stroke sizes, and $I^{\mathcal{T}_i}$ represents an image $I$ with the stroke size $\mathcal{T}_i$. The problem studied in this paper is to incorporate different stroke sizes $\mathcal{T}_i \in \mathbb{T}$ into the feed-forward fast neural style transfer model. Firstly, we formulate the feed-forward stylization process as:

\begin{equation}
g(I_c) = I_o,  \quad I_o \sim p(I_o| I_c, I_s),
\label{eg:previous}
\end{equation}
where $g$ is the trained generator. And the target statistic $p(I_o)$ of the output image $I_o$ is characterized by two components, which are the semantic content statistics derived from the input image $I_c$, and the visual style statistics derived from the style image $I_s$.

Our feed-forward style transfer process for producing multiple stroke sizes can then be modeled as:

\begin{equation}
g'(I_c, \mathcal{T}_i) = I^{\mathcal{T}_i}_{o},  \quad I^{\mathcal{T}_i}_{o} \sim p(I^{\mathcal{T}_i}_{o}| I_c, I_s, \mathcal{T}_i)\ (\mathcal{T}_i \in \mathbb{T}).
\label{eg:now}
\end{equation}
We aim to enable one single generator $g'$ to produce stylized results with multiple stroke sizes $\mathcal{T}_i \in \mathbb{T}$ for the same content image $I_c$.

%Our proposed algorithm has the benefits of both two aforementioned possible solutions, \ie exploit one single generator $g'$ and do not need to resize the test image $I_c$.

\subsection{Network Architecture}

%\myLcomment{Need to explain why the other two did not work}
%%%%%%
\begin{figure}[!t]
  \centering
    %\label{fig:subfig:a} %% label for first subfigure
    \includegraphics[width=\textwidth]{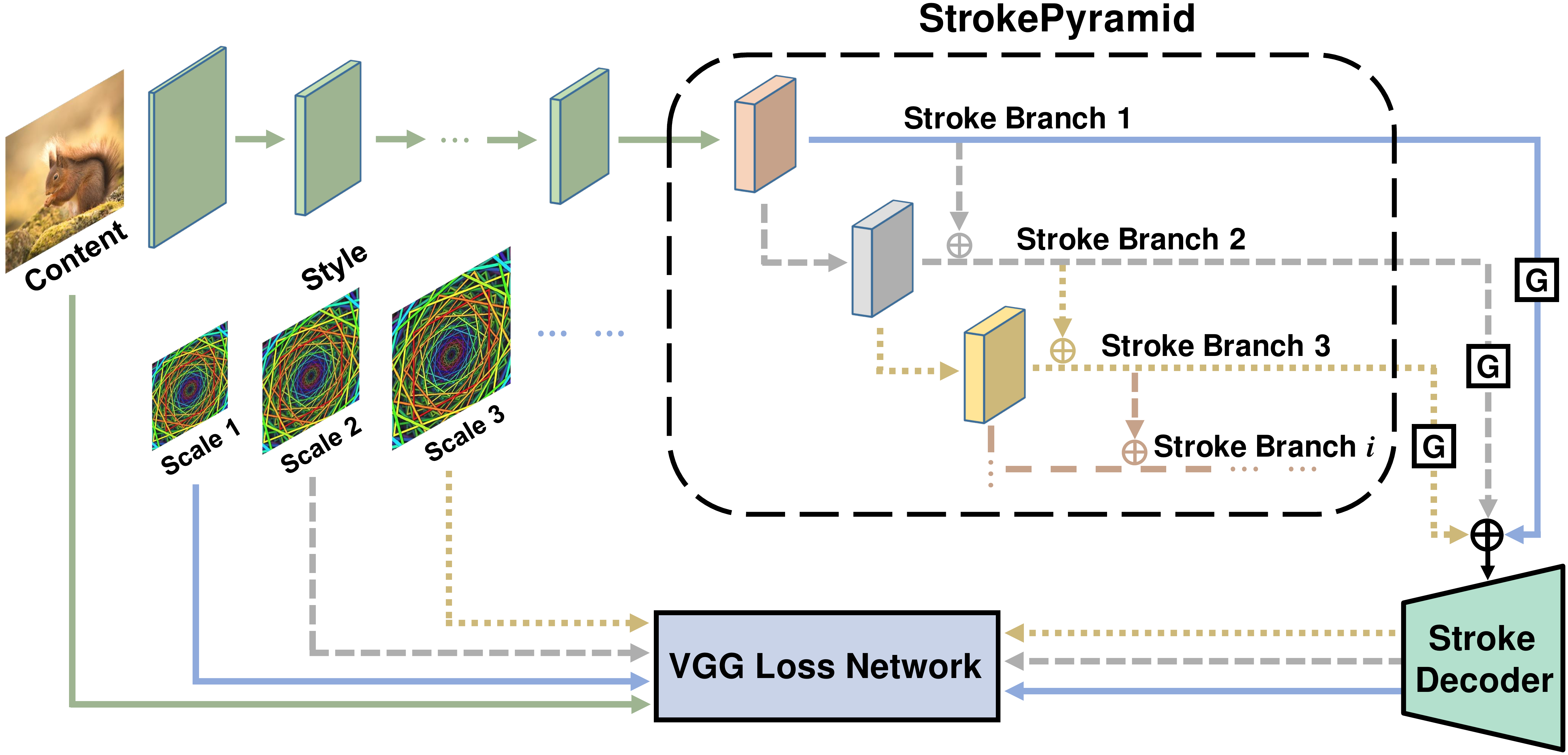}
  \caption{An overview of our network architecture with the \emph{StrokePyramid}. It consists of several stroke branches with gating functions. Each stroke branch corresponds to a specific stroke size.}
  \label{fig:arch} %% label for entire figure
\end{figure}

Based on the analysis in Section~\ref{sect:analysis}, to incorporate different stroke sizes into one single model, we propose to design a network with adaptive receptive fields and each receptive field is used to learn a corresponding size of stroke. The network architecture of our proposed approach is depicted in Figure~\ref{fig:arch}.

%In addition, the network should encourage content consistency between adjacent stroke size control results which should differ only in stroke sizes. Motivated by this consideration, we propose to augment a new stroke size based on the previously augmented stroke sizes. Our network architecture is depicted in Figure~\ref{fig:arch}.

Our network consists of three components. At the core of our network, a \emph{StrokePyramid} module is proposed to decompose the network into several stroke branches. Each branch has a larger receptive field than the previous branch through progressively growing convolutional filters. In this way, our network also encourages \emph{stroke consistency} (which refers to the consistency of stroke orientation, configuration, \etc) between adjacent stroke size control results which should differ only in stroke sizes. By handling different stroke branches, the \emph{StrokePyramid} can regulate the receptive field in the generative network. With different receptive fields, the network learns to paint strokes with different sizes. In particular, to better preserve the desired size of strokes, larger strokes are learned with larger receptive fields, as explained in Section~\ref{sect:analysis}. During the testing phase, given a signal which indicates the desired stroke size, the \emph{StrokePyramid} automatically adapts the receptive field in the network and the stylized result with a corresponding stroke size can be produced.

%For each size of the receptive field, the network learns to paint strokes with a corresponding size.

In addition to the \emph{StrokePyramid}, there are two more components in the network, namely the pre-encoder and the stroke decoder. The pre-encoder module refers to the first few layers in the network and is shared among different stroke branches to learn both the semantic content of a content image and the basic appearances of a style. The stroke decoder module takes the feature maps from the \emph{StrokePyramid} as input and decodes the stroke feature into the stylized result with a corresponding stroke size. To determine which stroke feature to decode, we augment a gating function $G$ in each stroke branch. The gating function $G$ is defined as

\begin{equation}
G(\mathcal{F}^{\mathcal{B}_{s_i}}) = a_i \mathcal{F}^{\mathcal{B}_{s_i}}, \quad \sum \limits_{i}{a_i} = 1 \ (0 \leq a_i \leq 1),
%a_i =  \left\{
%\begin{aligned}
%0,  i \neq k \\
%1,  i = k
%\end{aligned}
%\right.
%,
\end{equation}
where $\mathcal{F}^{\mathcal{B}_{s_i}}$ is the output feature map of the branch $\mathcal{B}_{s_i}$ in the \emph{StrokePyramid}, which corresponds to the stroke size $\mathcal{T}_i$. For the selection of $a$, at the training stage, $a_i$ is binary. More specifically, $a_i = 1$ when $i=k$ (\ie, the selected stroke branch to be trained is $\mathcal{B}_{s_k}$). Otherwise, $a_i = 0$ when $i \neq k$. At the testing stage, $a_i$ can be fractional, which is the basis of our continuous stroke size control.

%Assume that the selected feature map to be decoded is from the branch $\mathcal{B}_{s_k}$.

All the stroke features from the \emph{StrokePyramid} need to go through the gating function and then be fed into the stroke decoder $Dec$ to be decoded into the output result $I^{\mathcal{T}_k}_{o}$ with the desired stroke size:

%\begin{equation}
%G(\mathcal{F}^{\mathcal{B}_{s_i}}) =  \left\{
%\begin{aligned}
%0\cdot \mathcal{F}^{\mathcal{B}_{s_i}},  i \neq k \\
%1\cdot \mathcal{F}^{\mathcal{B}_{s_i}},  i = k
%\end{aligned}
%\right.
%,
%\end{equation}

\begin{equation}
Dec(\sum \limits_{i}{G(\mathcal{F}^{\mathcal{B}_{s_i}})}) = I^{\mathcal{T}_k}_{o}.
\end{equation}

\subsection{Loss Function}

\textbf{Semantic loss.} The semantic loss is defined to preserve the semantic information in the content image, which is formulated as the Euclidean distance between the content image $I_c$ and the output stylized image $I_o$ in the feature space of the VGG network \cite{gatys2016image}.
%We follow \cite{gatys2016image} to formulate the semantic loss, which is the Euclidean distance in the feature %space of the VGG network \cite{gatys2015texture} (pre-trained on the ImageNet dataset \cite{russakovsky2015imagenet}).

% * <brooksong@ieee.org> 2017-11-09T07:51:05.633Z:
%
% > We follow \cite{gatys2016image}
% 又来了，为啥总是写follow xxx，都是follow那你自己还剩下什么？
%
% ^ <398372213@qq.com> 2017-11-12T01:30:16.480Z:
%
% 已修改。
%
% ^ <brooksong@ieee.org> 2017-11-13T01:42:33.085Z.

%
%As the gram-based style statistic is scale-sensitive, for the same content image $I_c$, a single trained generator $g$ can only produce the result with a single stroke size.

Assume that $\mathcal{F}^{l}(I) \in \mathbb{R}^{C \times H \times W}$ represents the feature map at layer $l$ in VGG network with a given image $I$, where $C$, $H$ and $W$ denote the number of channels, the height and width of the feature map respectively. The semantic content loss is then defined as:

%\begin{equation}
%\mathcal{L}_{c} =  \sum\limits_{l \in \{l_c\}}|| \mathcal{F}^{l}(I_c) - \mathcal{F}^{l}(I_o) ||^2.
%\end{equation}
%\sum\mathop{}_{l \in \{l_c\}}\lVert F^l(O_i) - F^l(I)\rVert^2
\begin{equation}
\mathcal{L}_{c} =  \sum\mathop{}_{l \in \{l_c\}}\lVert \mathcal{F}^{l}(I_c) - \mathcal{F}^{l}(I_o) \rVert^2,
\end{equation}
where $\{l_c\}$ represents the set of VGG layers used to compute the content loss.

\textbf{Stroke loss.} The visual style statistics can be well represented by the correlations between filter responses of the style image $I_s$ in different layers of pre-trained VGG network. These feature correlations can be obtained by computing the Gram matrix over the feature map at a certain layer in VGG network. As the gram-based statistic is scale-sensitive, representations of different stroke sizes can be obtained by simply resizing the given style image.

By reshaping $\mathcal{F}^{l}(I)$ into $\mathcal{F}^{l}(I)' \in \mathbb{R}^{C \times (H \times W)}$, the Gram matrix $\mathcal{G}(\mathcal{F}^{l}(I)') \in \mathbb{R}^{C \times C}$ over feature map $\mathcal{F}^{l}(I)'$ can be computed as:
\begin{equation}
\mathcal{G}(\mathcal{F}^{l}(I_s)')= [\mathcal{F}^{l}(I_s)'] [\mathcal{F}^{l}(I_s)']^T.
\label{eg:gram}
\end{equation}
The stroke loss for size $\mathcal{T}_k$ can be therefore defined as:

\begin{equation}
\mathcal{L}_{\mathcal{T}_k} =  \sum\mathop{}_{l \in \{l_s\}}\lVert \mathcal{G}(\mathcal{F}^{l}(\mathcal{R}(I_s, \mathcal{T}_k))') - \mathcal{G}(\mathcal{F}^{l}(I_o^{\mathcal{B}_{s_k}})') \rVert^2,
\end{equation}
where $\mathcal{R}$ represents the function that resizes the style image to an appropriate scale according to the desired stroke size $\mathcal{T}_k$, and $I_o^{\mathcal{B}_{s_k}}$ represents the output of the $k$-th stroke branch. $\{l_s\}$ is the set of VGG layers used for style loss.
%to calculate the style loss.

The total loss for stroke branch $\mathcal{B}_{s_k}$ is then  written as:

\begin{equation}
\mathcal{L}_{\mathcal{B}_{s_k}} = \alpha \mathcal{L}_{c} + \beta_k \mathcal{L}_{\mathcal{T}_k} + \gamma \mathcal{L}_{tv}, %+ \gamma \mathcal{L}_{reg}.
\label{eg:totalloss}
\end{equation}
where $\alpha$, $\beta$ and $\gamma$ are balancing factors. $\mathcal{L}_{tv}$ is a total variation regularization loss to encourage smoothness in the generated images.

\subsection{Training Strategies}
\label{sect:trainstrategy}

\textbf{Progressive training.} To train different stroke branches in one single network, we propose a progressive training strategy. This training strategy stems from the intuition that the training of the latter stroke branch benefits from the knowledge of the previously learned branches. Taken this into consideration, the network learns different stroke sizes with different stroke branches progressively. Assume that the number of the stroke sizes to be learned is $K$. For every $K$ iterations, the network firstly updates the first stroke branch in order to learn the smallest size of stroke. Then, based on the learned knowledge of the first branch, the network uses the second stroke branch to learn the second stroke size with a corresponding scale of the style image. In particular, since the second stroke branch grows the convolutional filters on the basis of the first stroke branch, the updated components in the previous iteration are also adjusted. Similarly, the following stroke branches are updated with the same progressive process. In the next $K$ iterations, the network repeats the above progressive process, since we need to ensure that the network preserves the previously learned stroke sizes.

%The first stroke branch learns the smallest size of strokes and also encodes some semantic information. Then the network progressively updates the following stroke branches to learn the remaining stroke sizes. As the only difference between different stroke branches is the learned stroke sizes, it will be easier for the latter stroke branch to learn a new stroke size on the basis of the previously learned knowledge. In other words, these stroke branches can mutually improve the training of each other.

\textbf{Incremental training.} We also propose a flexible incremental training strategy to efficiently augment new stroke sizes upon a trained model. Given a new desired stroke size, instead of learning from scratch, our algorithm incrementally learns the new stroke size by adding one single layer as a new stroke branch in the \emph{StrokePyramid}. The position of the augmented layer depends on the previously learned stroke sizes and their corresponding receptive fields. By fixing other network components and only updating the augmented layer, the network learns to paint a new size of strokes on the basis of the previously learned stroke features and thus can reach convergence quickly.

\subsection{Runtime Control Strategies}
\label{sect:runtime}
%\paragraph{Stroke Capabilities}\myLcomment{discussion of incremental learning for different stroke sizes?}

%\myLcomment{Use texture synthesis to produce intermediate adding stroke size}

\textbf{Continuous stroke size control.} One of the advantages of our algorithm over previous approaches is that our algorithm can endow one single model with the ability of finer continuous stroke size control. We propose a stroke interpolation strategy to exploit our architecture to interpolate between trained stroke sizes in the feature embedding space, instead of training with tons of style image scales.

Given a content image $I_c$, we assume that $\mathcal{F}^{\mathcal{B}_{s_m}}$ and $\mathcal{F}^{\mathcal{B}_{s_n}}$ are two output feature maps in the \emph{StrokePyramid}, which can be decoded into the stylized results with two stroke sizes $I^{\mathcal{T}_m}_{o}$ and $I^{\mathcal{T}_n}_{o}$ respectively. The interpolated feature $\mathcal{F}^{\mathcal{B}_{\widetilde s}}$ can then be obtained by controlling the gating functions in Figure~\ref{fig:arch} to interpolate between output feature maps in the \emph{StrokePyramid}:

\begin{equation}
\mathcal{F}^{\mathcal{B}_{\widetilde s}} = a_m \mathcal{F}^{\mathcal{B}_{s_m}} + (1-a_m) \mathcal{F}^{\mathcal{B}_{s_n}}.
\end{equation}
By gradually changing the value of $a_m$ and feeding the obtained $\mathcal{F}^{\mathcal{B}_{\widetilde s}}$ into the stroke decoder module, stylized results with arbitrary intermediate stroke sizes $I^{\mathcal{\widetilde T}}_{o}$ can be produced.
%\begin{equation}
%f(I^{\mathcal{\widetilde T}}_{c}) = a f(I^{\mathcal{T}_m}_{c}) + (1-a) f(I^{\mathcal{T}_n}_{c})
%\end{equation}
%I^{\mathcal{\widetilde T}}_{c}
%Then fetch two representations from the encoders and do the interpolation.

To our knowledge, none of previous approaches considers this much finer continuous stroke size control. However, from our point of view, there may be some possible solutions which can be derived from current approaches: 1) Directly interpolate between stylized results with different stroke sizes in the pixel space. 2) Design a network with different encoders but a shared decoder, and train each encoder and shared decoder jointly with different style image scales. Then interpolate between two representations from the encoders. 3) Rescale the style image and use ASPM methods to produce the corresponding results.

However, our algorithm outperforms these solutions in the following aspects correspondingly: 1) We manipulate the interpolation in the feature embedding space to achieve perceptually superior results \cite{Johnson2016perceptual,dosovitskiy2016generating}. 2) Our stroke representations are obtained with different receptive fields in the \emph{StrokePyramid}. As explained and verified in Section~\ref{sect:analysis} and Figure~\ref{fig:analysis}, our stroke representations are perceptually better than those obtained with the same receptive field. In addition, the
results of our proposed \emph{StrokePyramid} are more consistent in stroke orientations and configurations during stroke size control. The comparison results can be found in the supplementary material\footnote{\url{https://yongchengjing.com/pdf/strokeControllable\_supp.pdf}}. 3) ASPM compromises on visual quality and is generally not effective at producing fine strokes and details. Our algorithm outperforms ASPM in terms of quality and also stylization speed. %which will be verified in our experiments.
%subsequent
%To achieve a finer stroke size control, we propose a stroke interpolation algorithm to exploit our architecture to achieve continuous changes in stroke sizes. The idea of our algorithm is to interpolate between the output feature maps in the \emph{StrokePyramid} to achieve the interpolated stroke representations. Then, arbitrary intermediate stroke sizes can be produced by feeding the achieved stroke representations into the stroke decoder module.

%In addition, our stroke decoder is shared among different stroke branches, which has better generalization ability.

\textbf{Spatial stroke size control.} Previously, in the community of Fast Style Transfer, stylized results usually have almost the same stroke size across the whole image, which is impractical in the real case. Our algorithm supports mixed stroke sizes in different spatial regions and also with only one single model. In this way, the contrast information in stylized results can be enhanced.

Our spatial stroke size control is achieved by feeding masked content image through different corresponding stroke branches by controlling the gating functions, and then combining these stylized results. The mask can be obtained either by manual labelling or forwarding the content image through a pre-trained semantic segmentation network, \eg, DeepLabv2 \cite{chen2017deeplab}. By further combining our continuous stroke size control strategy, our algorithm provides practitioners a much finer control over the stylized results.

%We believe that it is promising to incorporate semantic segmentation network as a module in our framework to automatically determine stroke sizes for different spatial regions.

\section{Experiment}
\label{sect:experiment}

\subsection{Implementation Details}
%\paragraph{Training Details.}\myLcomment{data, style, optimizer, batch size layer, pretrained loss network, weight?}
Our proposed network is trained on MS-COCO dataset \cite{lin2014microsoft}. All the images are cropped and resized to $512 \times 512$ pixels before training. We adopt the Adam optimizer \cite{kingma2014adam} during training. The pre-trained VGG-19 network \cite{simonyan2014very} is selected as the loss network and $\{ relu1\_1, relu2\_1, relu3\_1, relu4\_1, relu5\_1 \}$ are used as the style layers and $relu4\_2$ is used as the content layer. By default, the number of initially learned stroke sizes is set to 3 to ensure the ability of stroke decoder, and the scales are 256, 512, and 768 for different stroke sizes for all styles in our experiment. More information can be found in the supplementary material.
%For a fair comparison, the parameters of the existing algorithms are set to be the default values according to their published literature.
%Our network architecture is based on \cite{Johnson2016perceptual} but uses fewer channels to further reduce the computational complexity. Our implementation is based on TensorFlow \cite{abadi2016tensorflow}.

%To further enable the model with the ability of learning multiple styles, conditional instance normalization operator \cite{dumoulin2016learned} is added instead of batch or instance normalization in our network.

%%%%%%%%%%%%%%%%%
\newcommand\p{2.8cm}
\newcommand\n{0.22}

\begin{figure}[!t]
\setlength\tabcolsep{2 pt}
{\renewcommand{\arraystretch}{0.3}
%\begin{tabular}{>{\centering}n{\p} >{\centering}n{\p} >{\centering\arraybackslash}n{\p}}
\begin{tabular}{>{\centering}m{0.1cm} >{\centering}m{\p} >{\centering}m{\p} >{\centering}m{\p} >{\centering\arraybackslash}m{\p}}
%\begin{tabular}{ccccc}
\centering

&\includegraphics[width=\n\textwidth]{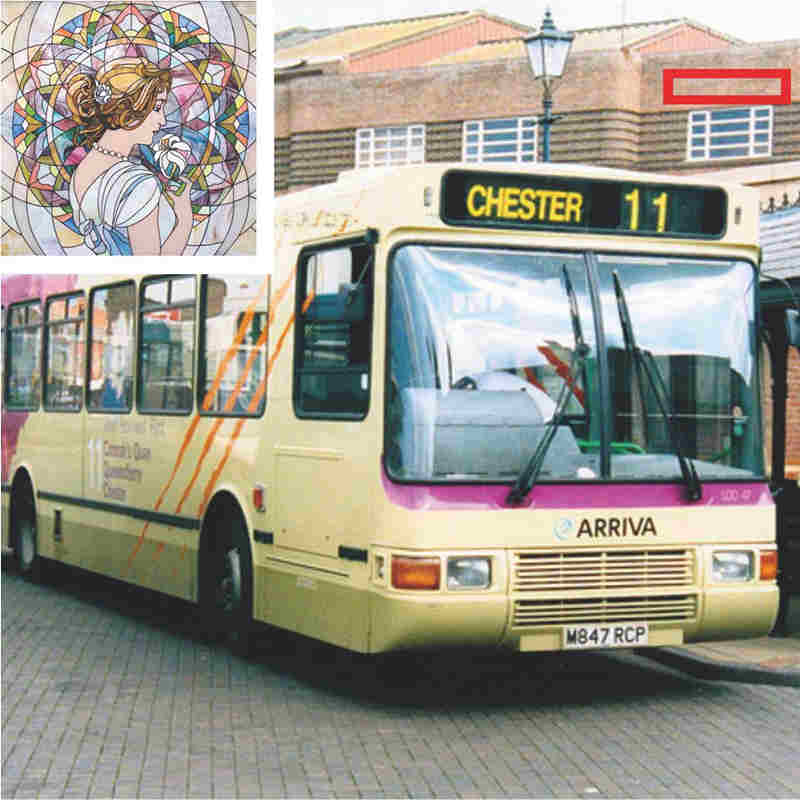} & \includegraphics[width=\n\textwidth]{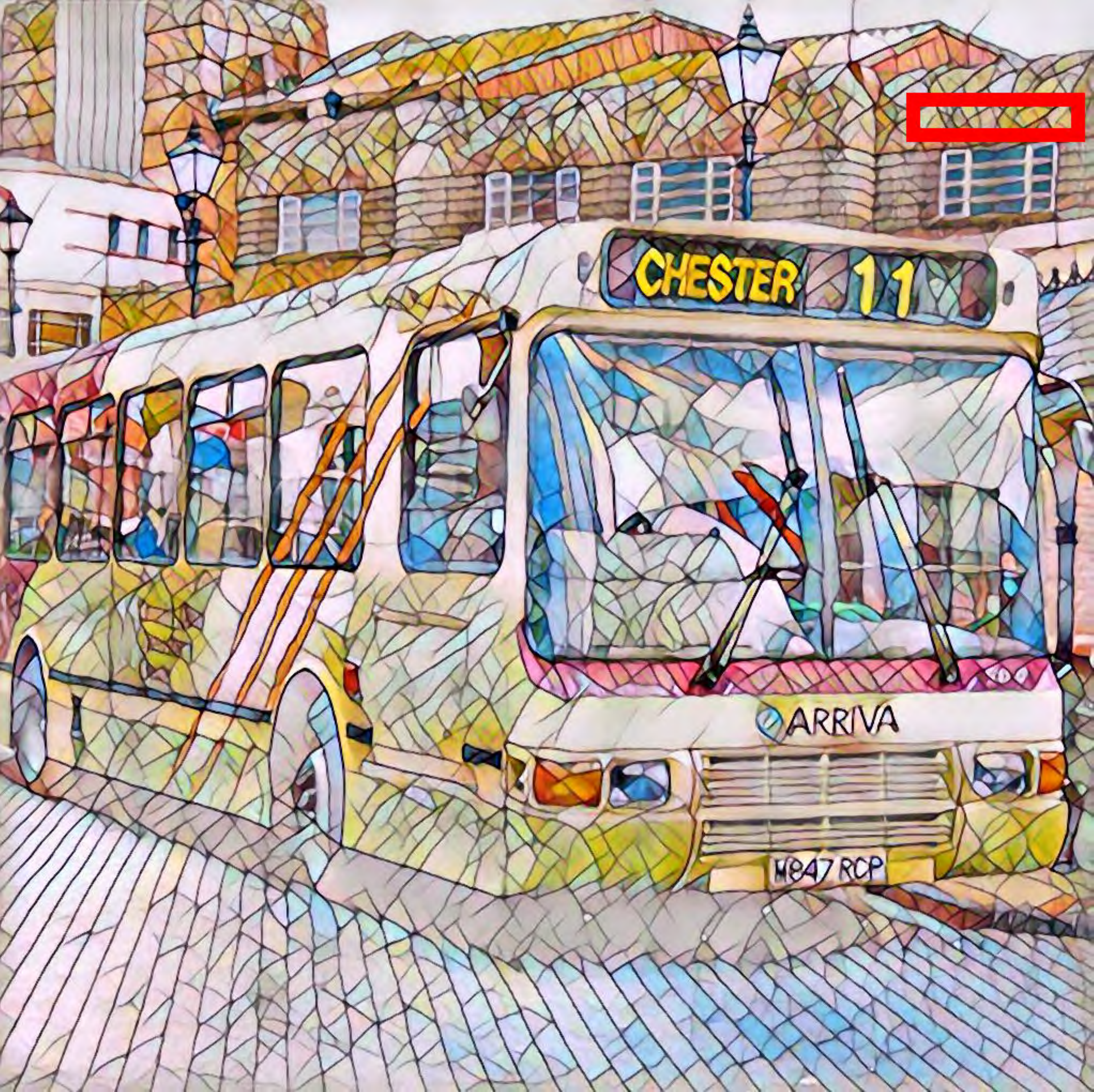}&\includegraphics[width=\n\textwidth]{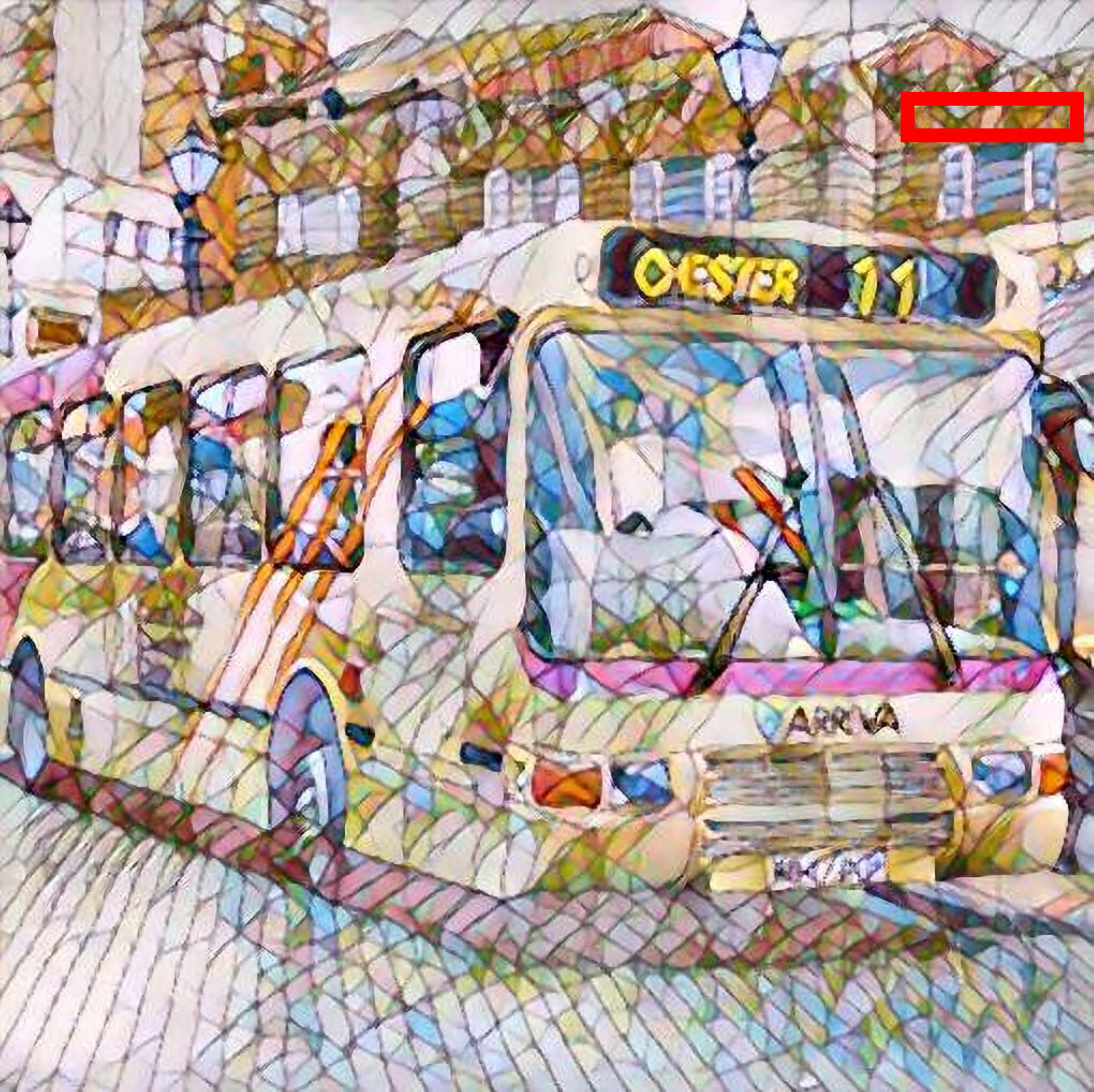}  & \includegraphics[width=\n\textwidth]{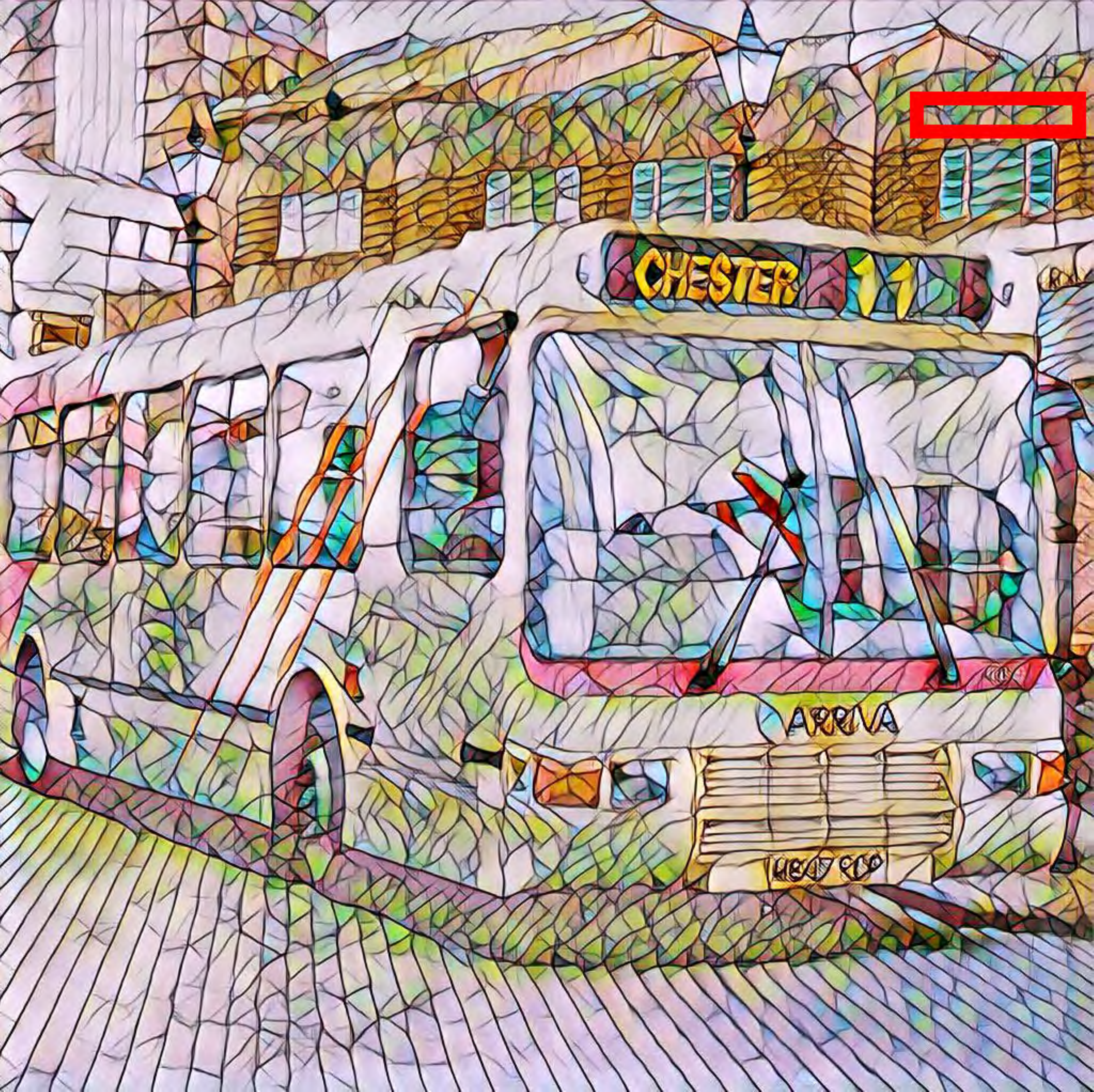}\\

%&&&&\\
%\smallskip
%\vspace{5cm}

&\includegraphics[width=\n\textwidth]{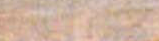}&\includegraphics[width=\n\textwidth]{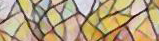} & \includegraphics[width=\n\textwidth]{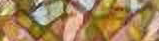} & \includegraphics[width=\n\textwidth]{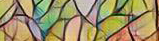} \\

%\smallskip

&&&&\\
 &\scriptsize{(a) Content \& Style\\}& \scriptsize{(b) Training a separate generator}&\scriptsize{(c) Image resizing + forwarding + SR} & \scriptsize{(d) Our proposed approach}\\

\end{tabular}
}
\caption{Quality comparison of our proposed algorithm and aforementioned two possible stroke control solutions in Section~\ref{sect:network}. SR represents the image super-resolution technique \cite{kim2016accurate}. The images in the second line represent the zoom regions in the red frames.}
%\myLcomment{draw the difference graph}
\label{fig:compareprvious} %% label for entire figure
\end{figure}

\subsection{Qualitative Evaluation}

\textbf{Comparison with previous solutions.} Sample results of our algorithm and two aforementioned possible solutions are shown in Figure~\ref{fig:compareprvious} (Figure~\ref{fig:compareprvious}(b) is produced by \cite{Johnson2016perceptual}). Our algorithm achieves comparable results with the first possible solution in Figure~\ref{fig:compareprvious}(b) regarding quality while preserving the flexibility of the second possible solution in Figure~\ref{fig:compareprvious}(c). Figure~\ref{fig:qualitativeresult} shows sample results of our algorithm and other Fast Style Transfer algorithms. Compared with \cite{Johnson2016perceptual,ulyanov2017improved}, our results with different stroke sizes are more consistent in stroke orientations and stroke configurations (the positions of the blue strokes in Figure~\ref{fig:qualitativeresult}). The stroke orientations and configurations in \cite{Johnson2016perceptual,ulyanov2017improved}'s results are more random, since they use different encoder-decoder pairs to learn different stroke sizes separately. By contrast, our \emph{StrokePyramid} can encourage \emph{stroke consistency} between adjacent stroke size control results which should differ only in stroke sizes. Compared with \cite{wang2016multimodal}, our algorithm can exploit one single trained model to achieve continuous and spatial stroke size control. Also, our model size is much smaller than \cite{wang2016multimodal}, which is 0.99 MB \emph{vs} 32.2 MB. Compared with other single-model stroke size control algorithms \cite{huang2017arbitrary,li2017universal}, our results capture finer strokes and more details. Also, our results seem to be superior in terms of visual quality. More explanations and comparison results can be found in the supplementary material.

%For the results of \cite{Johnson2016perceptual,ulyanov2017improved}, we train a separate scale-specific generator for each stroke size. It can be noticed that our algorithm achieves competitive results against \cite{Johnson2016perceptual,ulyanov2017improved} but exploits only one single pre-trained model. Our algorithm also needs less training time due to our progressive and incremental training strategy. Similarly, our algorithm is comparable to \cite{wang2016multimodal} regarding quality, but \cite{wang2016multimodal} needs to train several separate stroke-specific subnets. \cite{huang2017arbitrary} and \cite{li2017universal} belong to the category of ASPM algorithms and can control the stroke size by just feeding different scales of style images. However, \cite{huang2017arbitrary,li2017universal} are not effective at producing some fine textures and details for some styles.

%Although \cite{li2017universal} captures finer textures, the details are not well preserved in some cases (Figure~\ref{fig:qualitativeresult}, the sixth row).

\textbf{Runtime user controls.} In Figure~\ref{fig:strokeinterpolation}, we show sample results of our proposed continuous stroke size control strategy. Our network is trained with three scales of the style image as default and we do the stroke interpolation between them to obtain totally six stroke sizes. The test content image is never seen during training. We also demonstrate the results of \cite{huang2017arbitrary,li2017universal} for comparison, as explained in Section~\ref{sect:runtime}. We compare the results of different algorithms both globally and locally. Globally, our algorithm seems to achieve superior performance in terms of visual quality. Locally, compared with \cite{huang2017arbitrary,li2017universal}, our algorithm is more effective at producing fine strokes and preserving details. In addition, as shown in Figure~\ref{fig:difference}, the absolute differences of our adjacent continuous stroke size control results have a much clearer stroke contour, which indicates that most strokes in our results increase or decrease in size together during continuous stroke size control. We have also produced a sample video to demonstrate our continuous stroke size control in the supplementary material. Figure~\ref{fig:mix} demonstrates the results of our spatial stroke size control strategy. Our spatial stroke size control is realized with only one single model. Compared with Figure~\ref{fig:mix}(c), controlling the stroke size in different spatial regions can enhance the contrast of stylized images and make AI-Created Art much closer to Human-Created Art.
%globally our results are more consistent in image content between adjacent stroke size control results, which only differ in stroke sizes.
% Although \cite{li2017universal} produces finer strokes than \cite{huang2017arbitrary}, our algorithm still preserves much finer structures and details.

\newcommand\x{0.142}
\newcommand\q{1.74cm}

\begin{figure*}[!t]
\setlength\tabcolsep{0.8pt}
%\setlength\extrarowheight{1pt}
%{\renewcommand{\arraystretch}{1}
%%\begin{tabular}{m{3cm} >{\centering}m{\q} >{\centering}m{\q} m{2.5cm} >{\centering}m{\q} >{\centering}m{\q} >{\centering\arraybackslash}m{\q}}
%%\begin{tabular}{m{2.8cm} >{\centering}m{\q} >{\centering}m{\q} >{\centering}m{\q} >{\centering}m{\q} >{\centering}m{\q} >{\centering\arraybackslash}m{\q}}
%\begin{tabular}{m{4.585cm}>{\centering}m{1.747cm} >{\centering}m{0.2mm}?{0.3mm}>{\centering}m{0.2mm} >{\centering\arraybackslash}m{1.74cm}}
%%\begin{tabular}{cc}
%%\centering
%%\toprule
%%& \textbf{\small{Group \uppercase\expandafter{\romannumeral1}}} & \textbf{\small{Group \uppercase\expandafter{\romannumeral2}}} & \textbf{\small{Group %\uppercase\expandafter{\romannumeral3}}} & \textbf{\small{Group \uppercase\expandafter{\romannumeral4}}} & \textbf{\small{Group %\uppercase\expandafter{\romannumeral5}}} & \textbf{\small{Group \uppercase\expandafter{\romannumeral6}}} & \textbf{\small{Group %\uppercase\expandafter{\romannumeral7}}}& \textbf{\small{Group \uppercase\expandafter{\romannumeral8}}}\\
%&\textbf{\tiny{Content}} \vspace{-0.8mm} & && \textbf{\tiny{Style}} \vspace{-0.8mm}\\
%
%&\includegraphics[width=\x\textwidth]{figs/exp/fangzi.jpg} &&& \includegraphics[width=\x\textwidth]{figs/exp/style_niao_new.jpg} \\
%
%%\toprule
%
%\end{tabular}
%}

%\smallskip
%%%%%%%%%%%%%%%%%%%%%%%%
%\smallskip

{\renewcommand{\arraystretch}{0.5}
\begin{tabular}{m{2.14cm}>{\centering}m{3cm} m{1.212cm} ?{0.3mm} m{1.67cm} >{\centering\arraybackslash}m{2cm}}
%\centering
%\toprule
%& \textbf{\small{Group \uppercase\expandafter{\romannumeral1}}} & \textbf{\small{Group \uppercase\expandafter{\romannumeral2}}} & \textbf{\small{Group %\uppercase\expandafter{\romannumeral3}}} & \textbf{\small{Group \uppercase\expandafter{\romannumeral4}}} & \textbf{\small{Group %\uppercase\expandafter{\romannumeral5}}} & \textbf{\small{Group \uppercase\expandafter{\romannumeral6}}} & \textbf{\small{Group %\uppercase\expandafter{\romannumeral7}}}& \textbf{\small{Group \uppercase\expandafter{\romannumeral8}}}\\
 & \textbf{\fontsize{6pt}{\baselineskip}\selectfont{Multiple Models}}& &&\textbf{\fontsize{6pt}{\baselineskip}\selectfont{Single Model}} \\

%\toprule

\end{tabular}
%%%%%%%%%%%%%%%%%%%%%%%%
\vspace{-4mm}
%\smallskip
%\vspace{-0.8mm}
%\begin{tabular}{>{\centering}m{2.8cm} >{\centering}m{\q} >{\centering}m{\q} >{\centering}m{\q} >{\centering}m{0.1mm}>{\centering}m{0.1mm} >{\centering}m{\q} >{\centering}m{\q} >{\centering\arraybackslash}m{\q}}
%\centering
%
%
%& \textbf{\footnotesize{Stroke Size \uppercase\expandafter{\romannumeral1}}} & \textbf{\footnotesize{Stroke Size \uppercase\expandafter{\romannumeral2}}} & \textbf{\footnotesize{Stroke Size \uppercase\expandafter{\romannumeral3}}}& &&\textbf{\footnotesize{Stroke Size \uppercase\expandafter{\romannumeral1}}} & \textbf{\footnotesize{Stroke Size \uppercase\expandafter{\romannumeral2}}} & \textbf{\footnotesize{Stroke Size \uppercase\expandafter{\romannumeral3}}} \\
%
%\end{tabular}
%%%%%%%%%%%%%%%%%%%%

\begin{tabular}{>{\centering}m{1cm} >{\centering}m{\q} >{\centering}m{\q} >{\centering}m{\q} >{\centering}m{0.2mm}?{0.3mm}>{\centering}m{0.2mm} >{\centering}m{\q} >{\centering}m{\q} >{\centering\arraybackslash}m{\q}}
%\centering

%\toprule
&&&&&&&& \\%[0.0ex]
&&&&&&&& \\%[0.0ex]
%& \textbf{\footnotesize{Stroke Size \uppercase\expandafter{\romannumeral1}}} & \textbf{\footnotesize{Stroke Size \uppercase\expandafter{\romannumeral2}}} & \textbf{\footnotesize{Stroke Size \uppercase\expandafter{\romannumeral3}}}& &&\textbf{\footnotesize{Stroke Size \uppercase\expandafter{\romannumeral1}}} & \textbf{\footnotesize{Stroke Size \uppercase\expandafter{\romannumeral2}}} & \textbf{\footnotesize{Stroke Size \uppercase\expandafter{\romannumeral3}}} \\

&\textbf{\tiny{\cite{Johnson2016perceptual}}} & \textbf{\tiny{\cite{ulyanov2017improved}}}&\textbf{\tiny{\cite{wang2016multimodal}}} && & \textbf{\tiny{\cite{huang2017arbitrary}}} & \textbf{\tiny{\cite{li2017universal}}} & \textbf{\tiny{Ours}} \\
%\tiny{Stroke Size \#1}

%\baselineskip
%\vspace{-1cm}
%\smallskip

\textbf{\tiny{SS \#1:}} & \includegraphics[width=\x\textwidth]{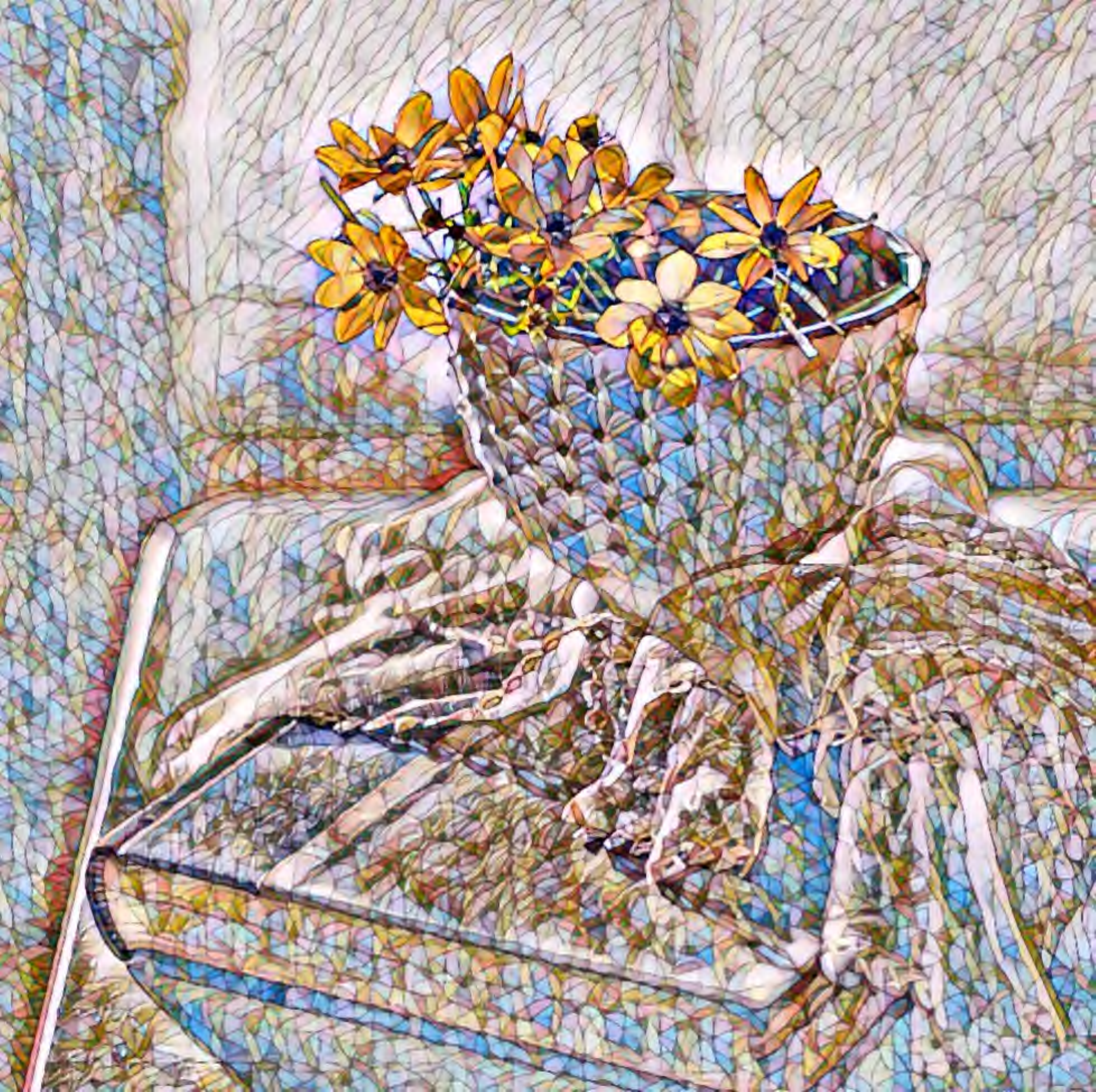}  & \includegraphics[width=\x\textwidth]{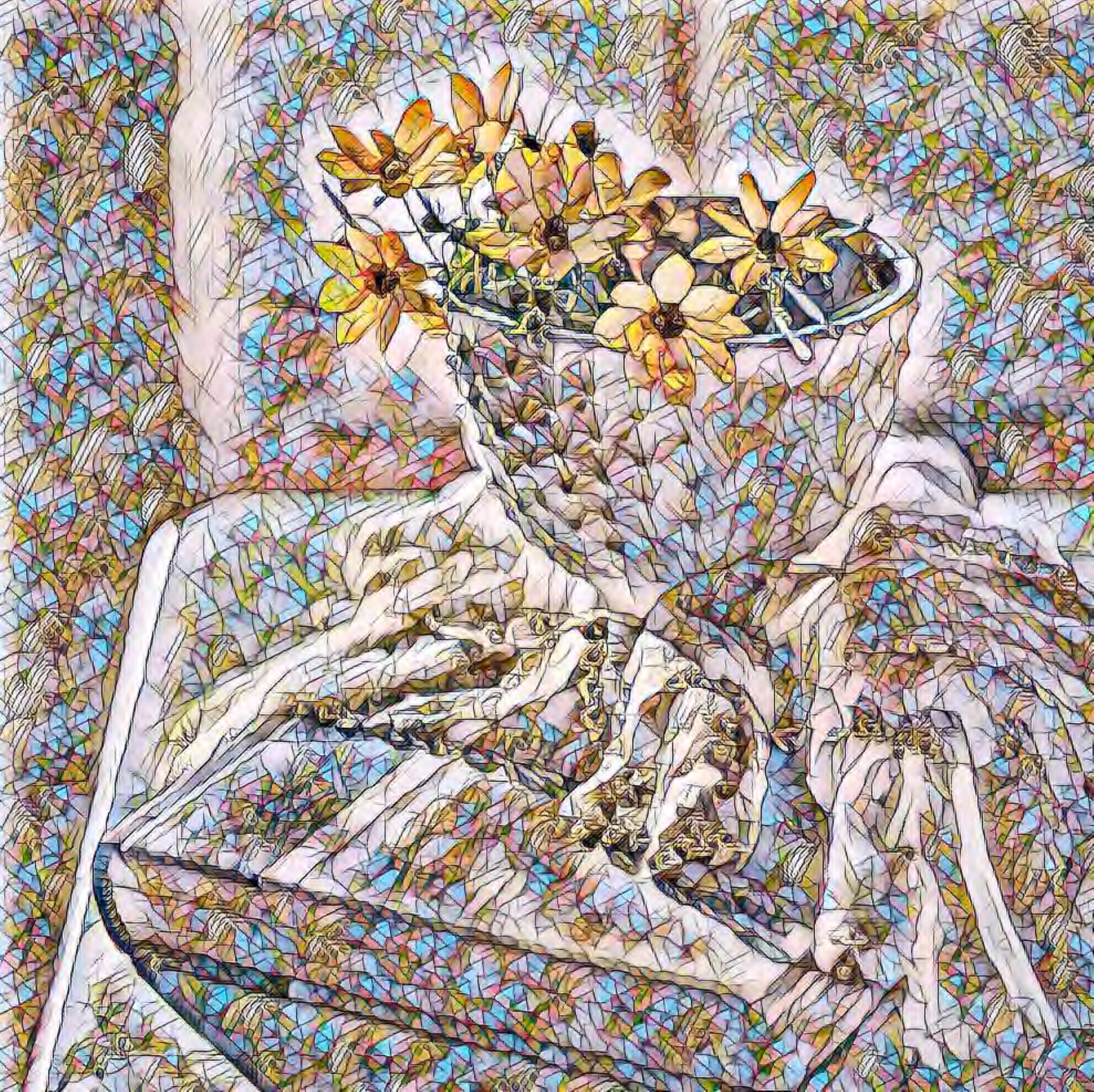}  & \includegraphics[width=\x\textwidth]{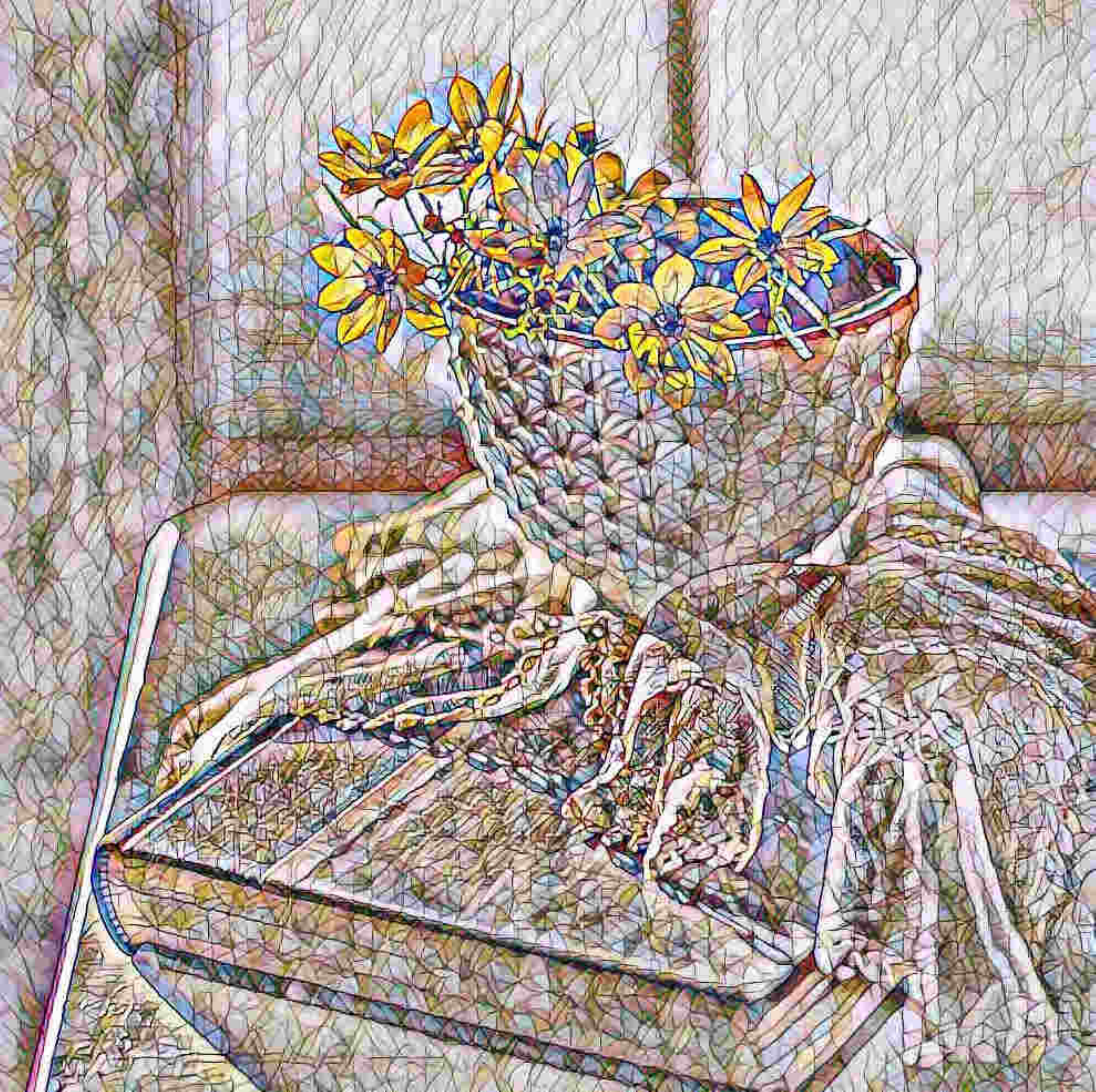} & && \includegraphics[width=\x\textwidth]{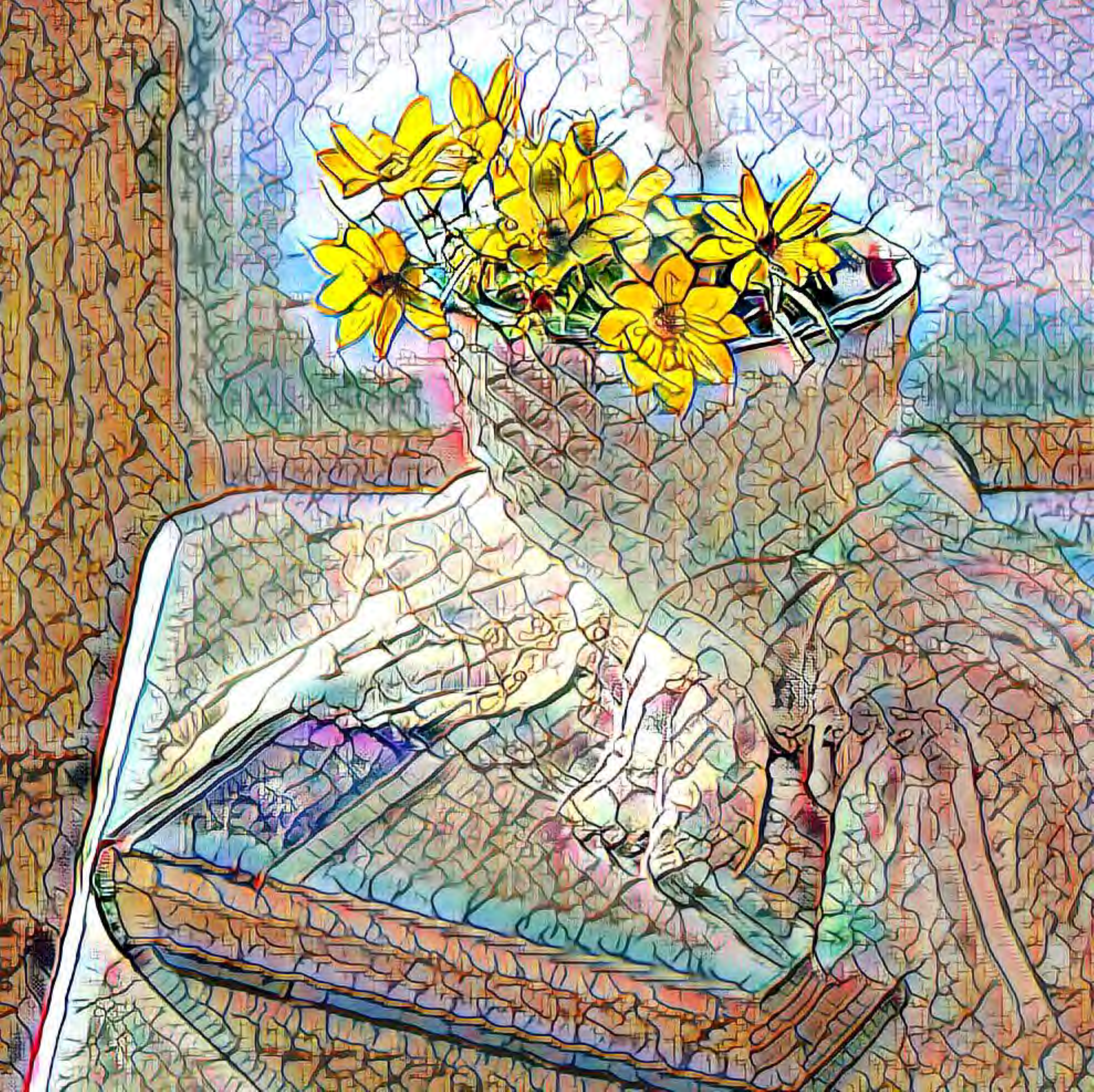} & \includegraphics[width=\x\textwidth]{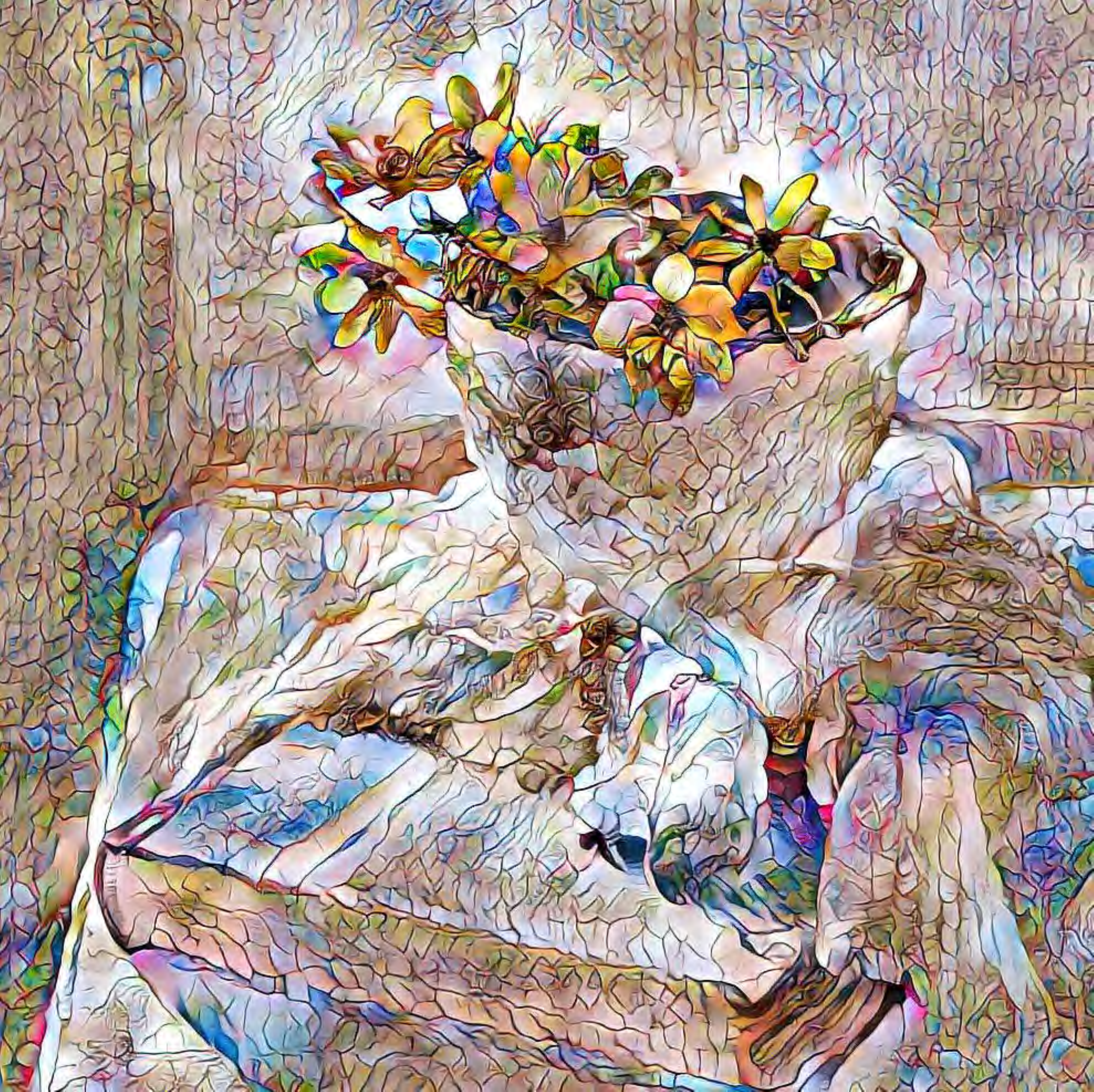}& \includegraphics[width=\x\textwidth]{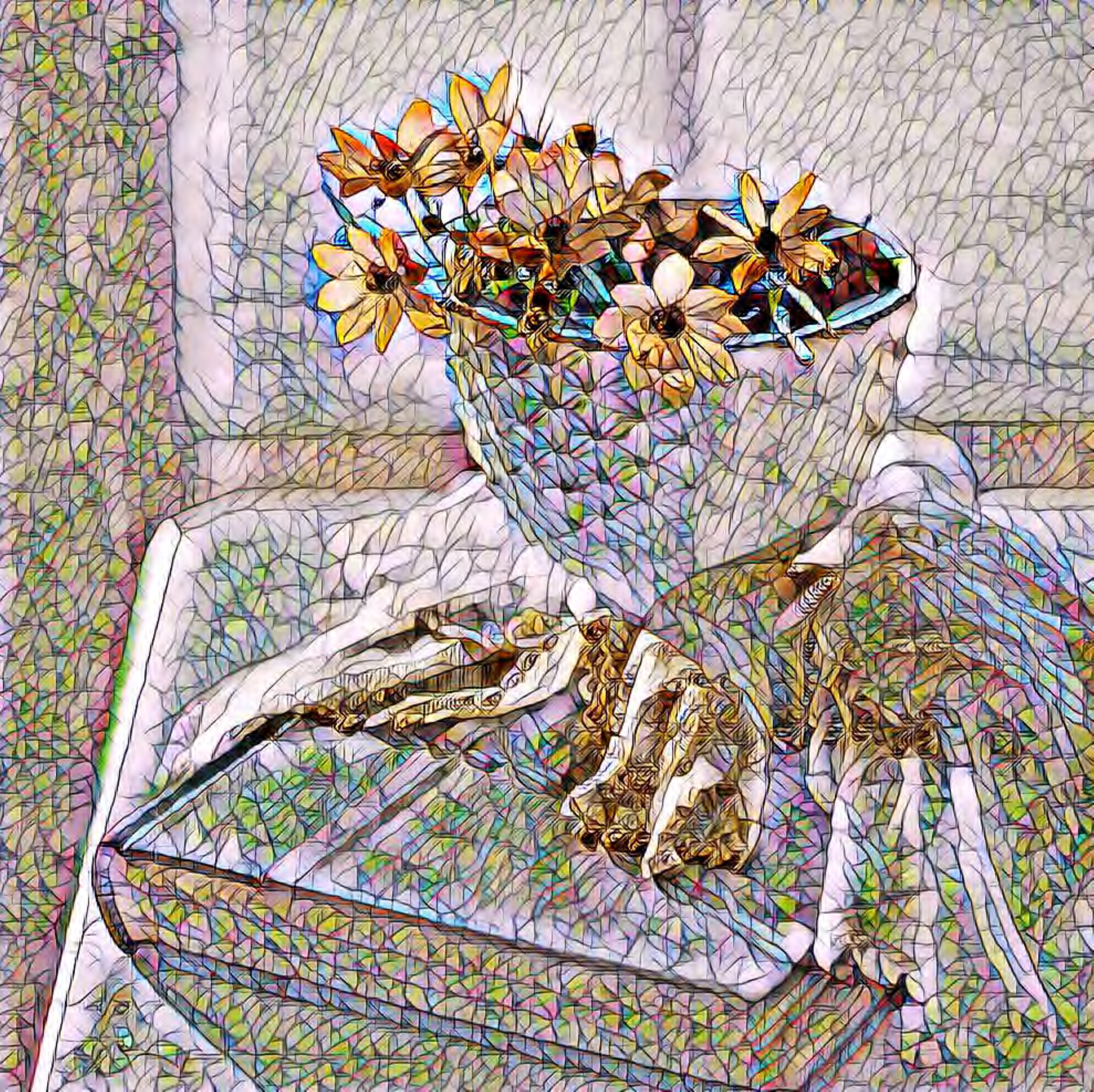} \\

\textbf{\tiny{SS \#2:}} & \includegraphics[width=\x\textwidth]{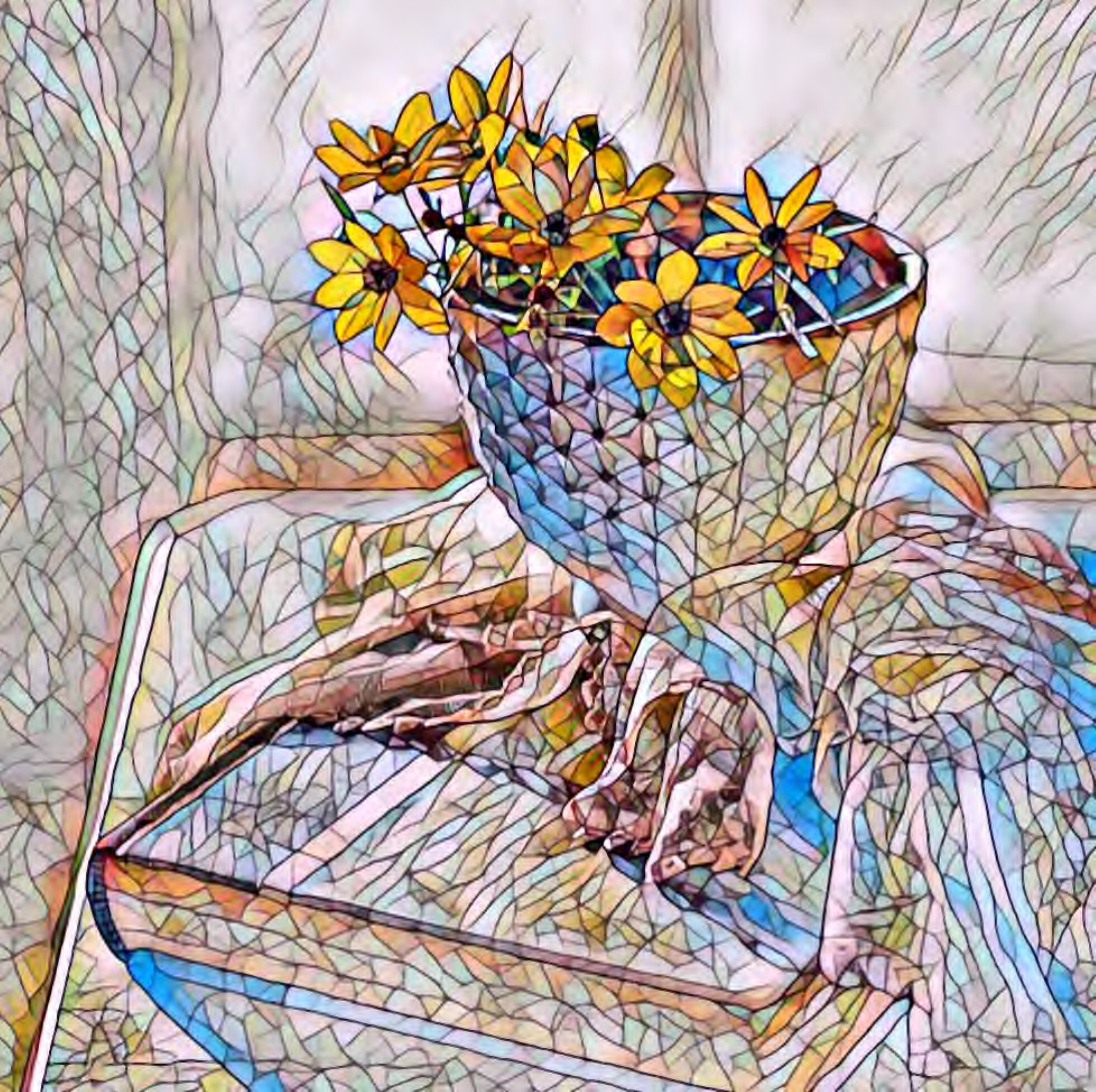}  & \includegraphics[width=\x\textwidth]{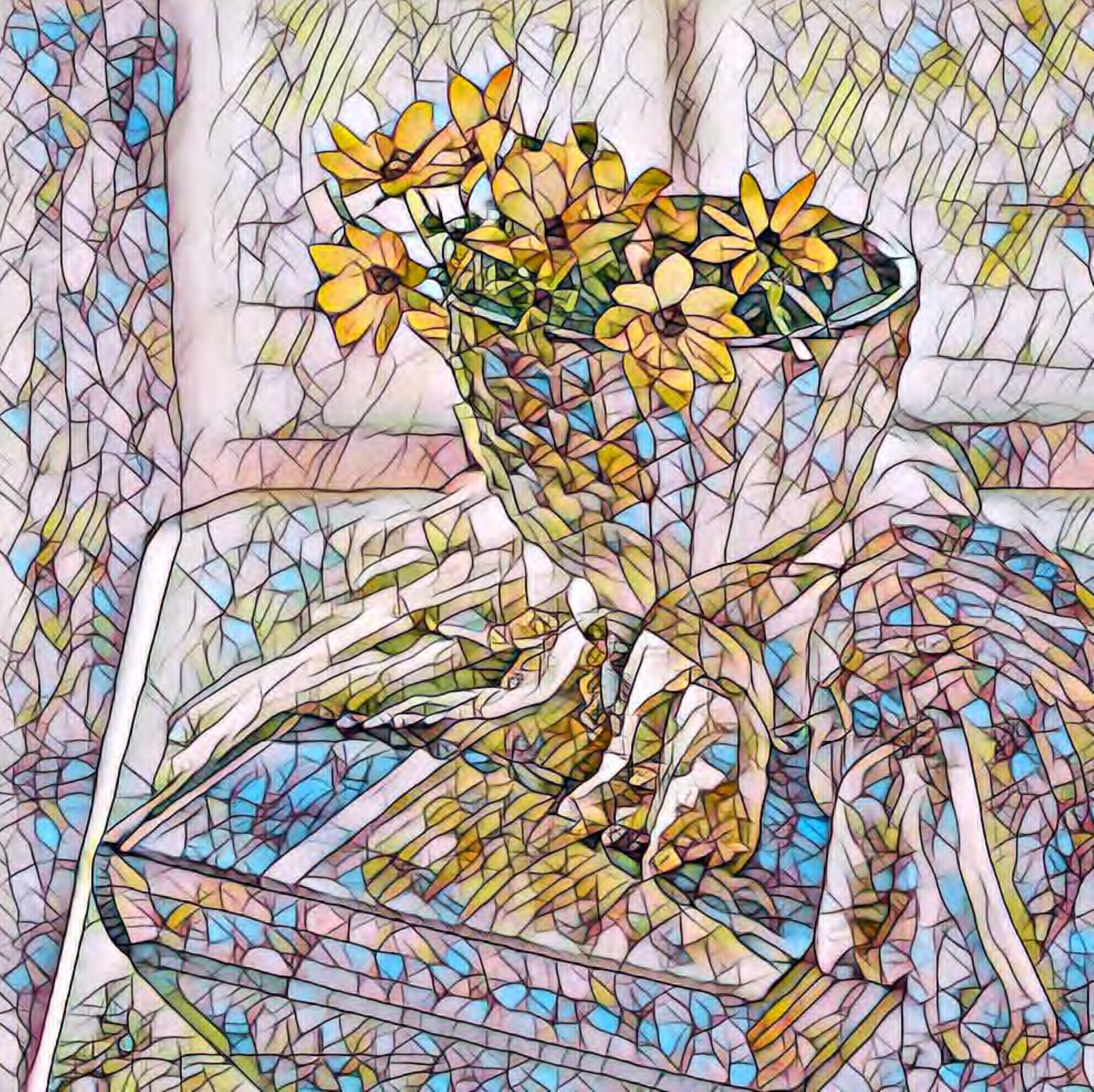}  & \includegraphics[width=\x\textwidth]{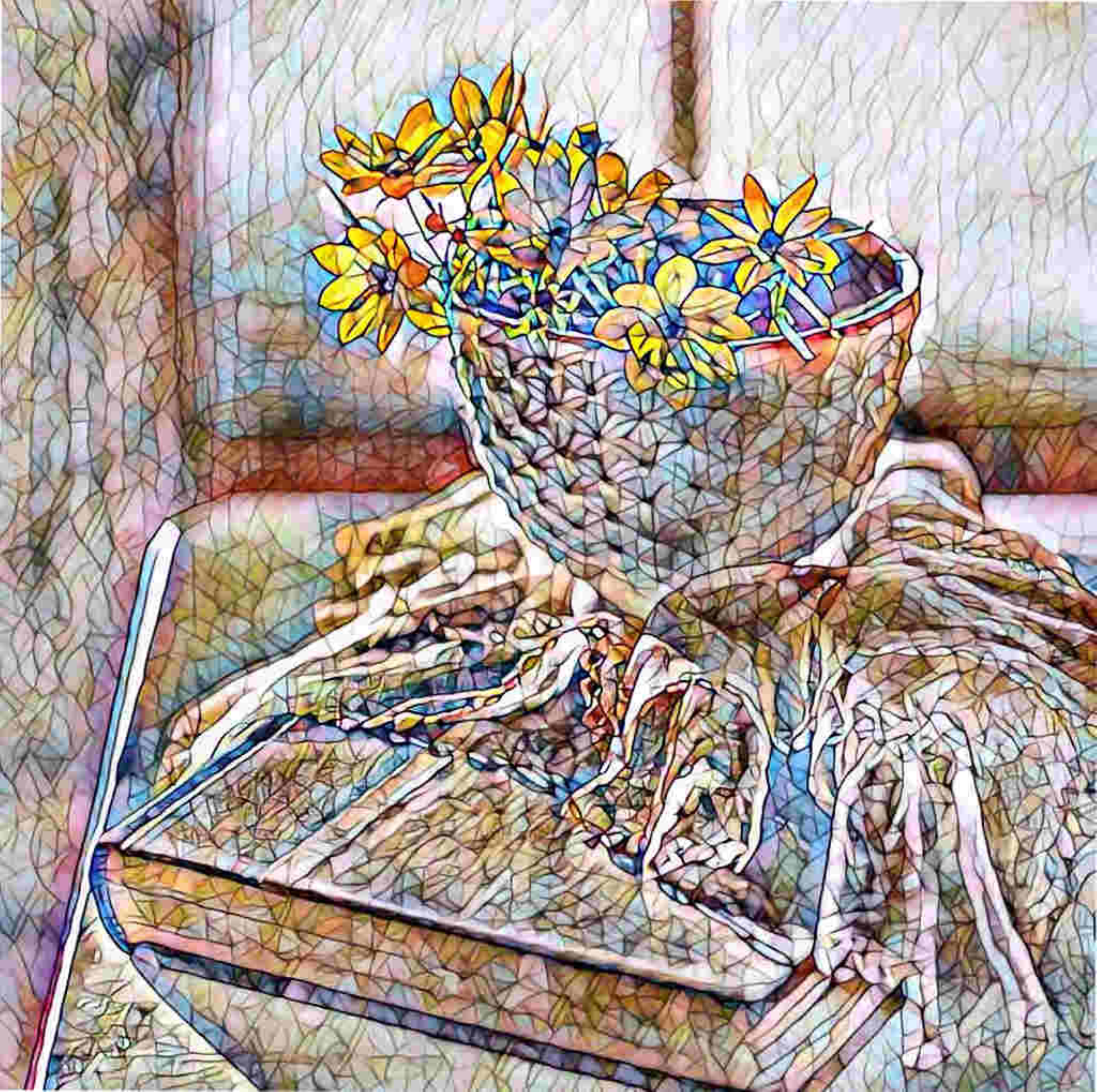} & && \includegraphics[width=\x\textwidth]{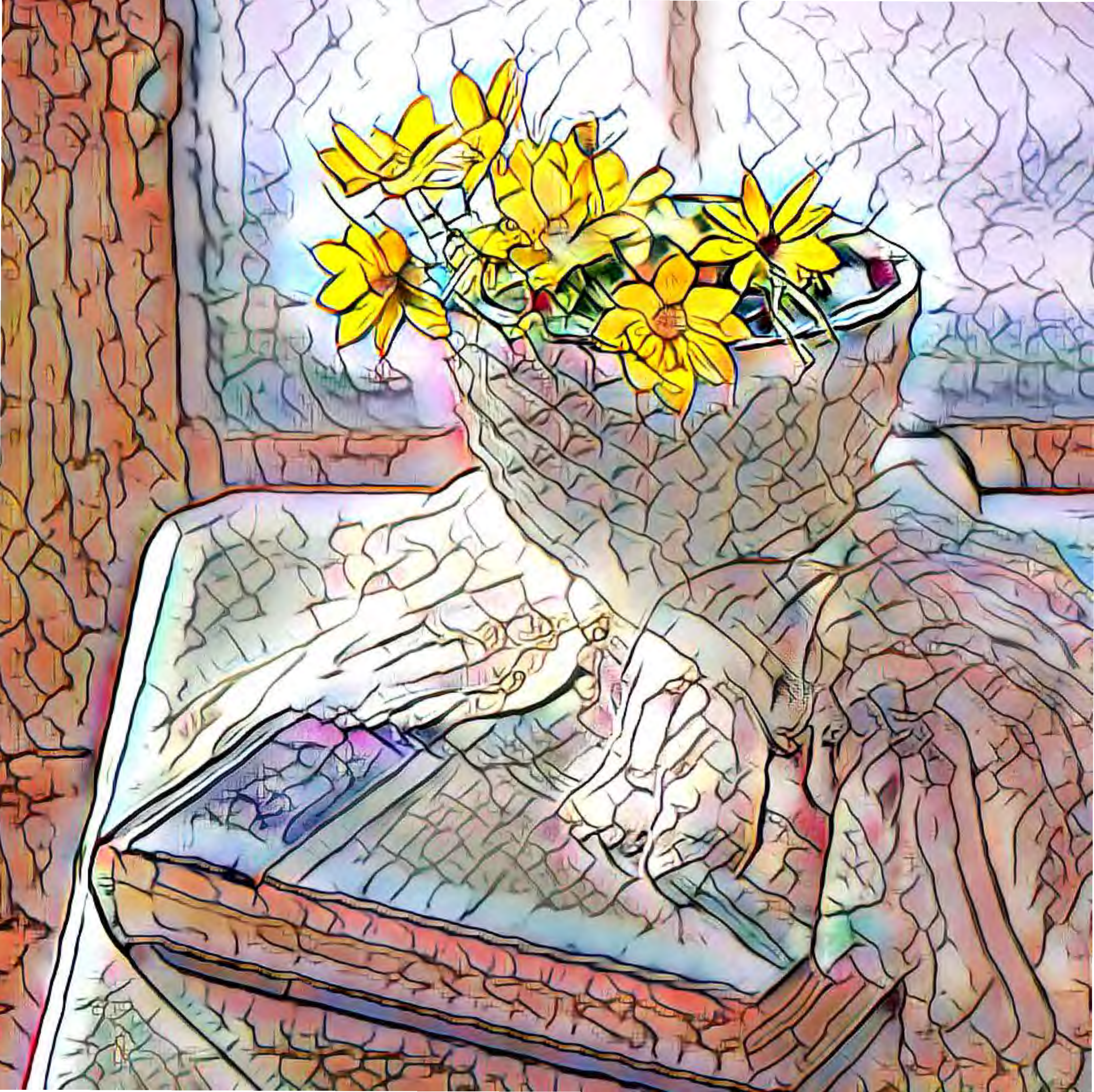} & \includegraphics[width=\x\textwidth]{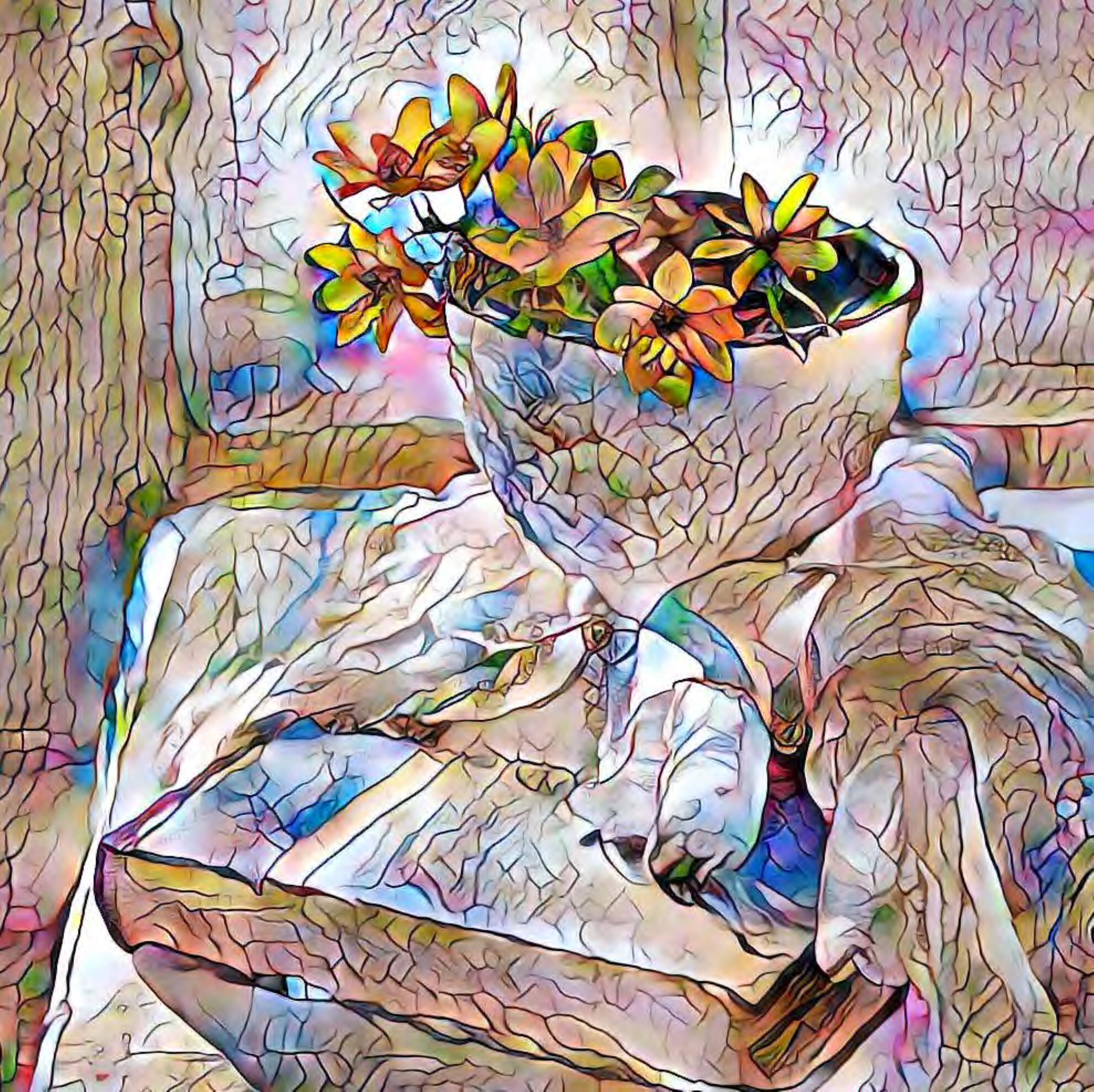}& \includegraphics[width=\x\textwidth]{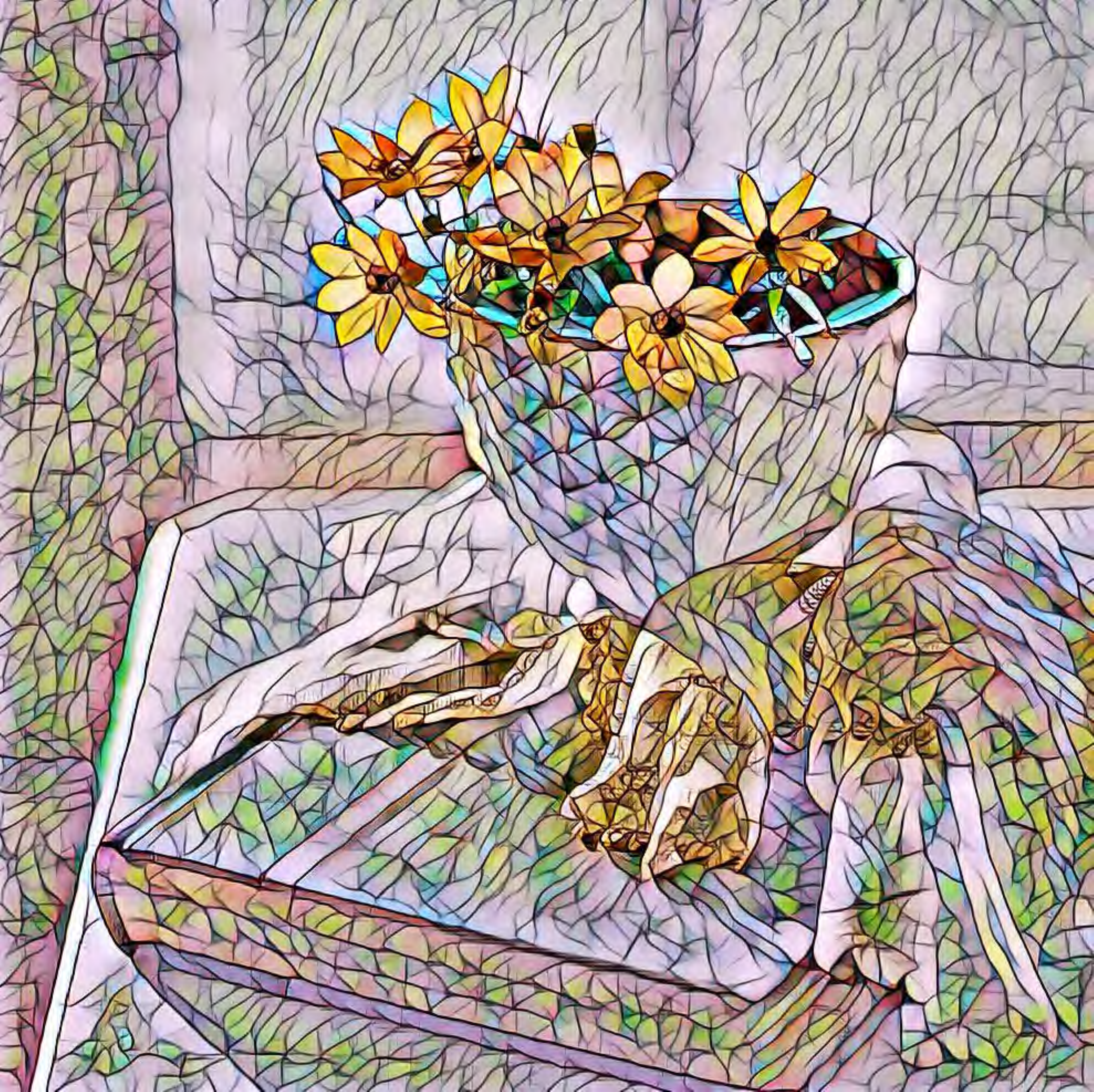} \\

\textbf{\tiny{SS \#3:}} & \includegraphics[width=\x\textwidth]{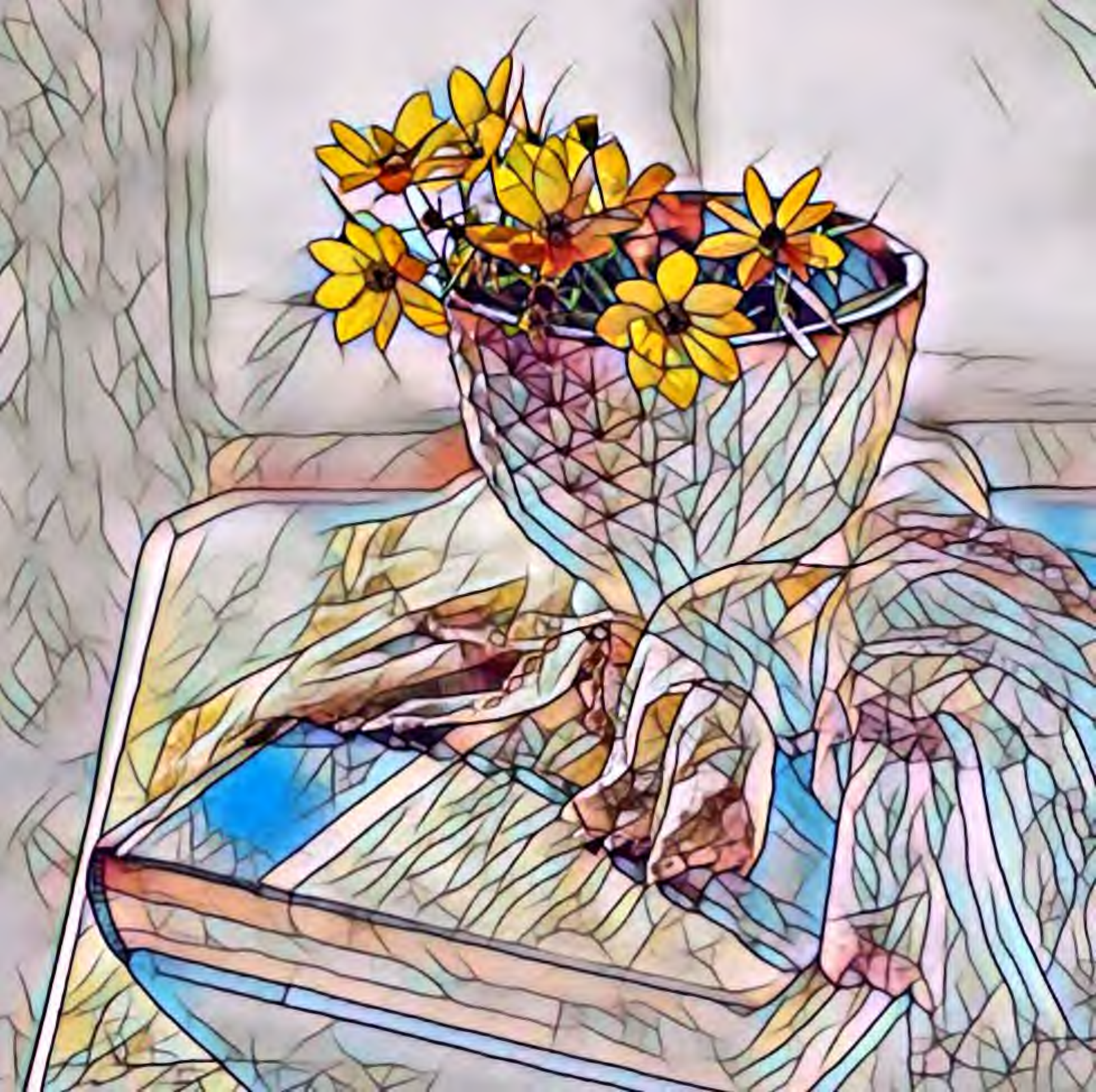}  & \includegraphics[width=\x\textwidth]{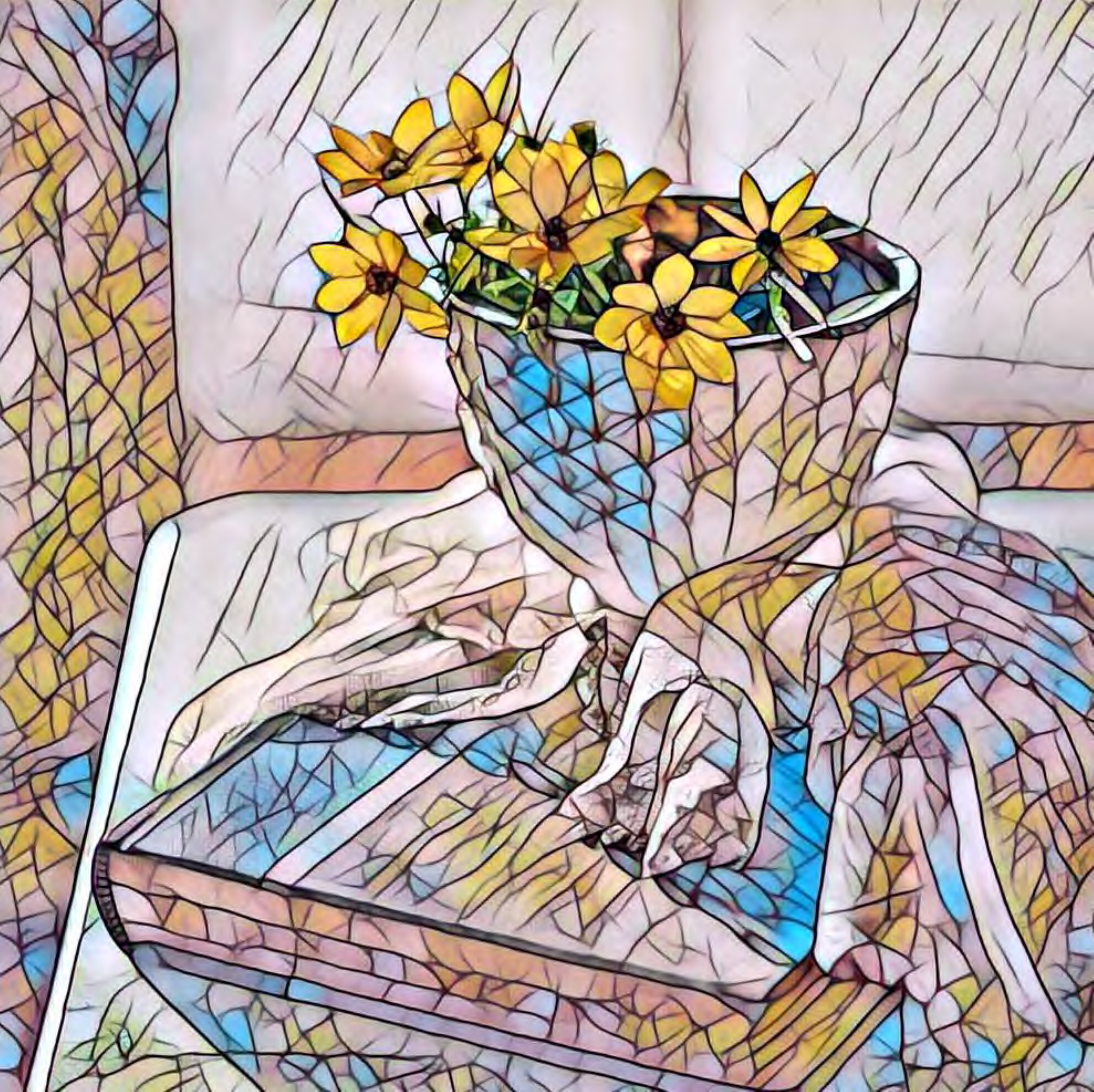}  & \includegraphics[width=\x\textwidth]{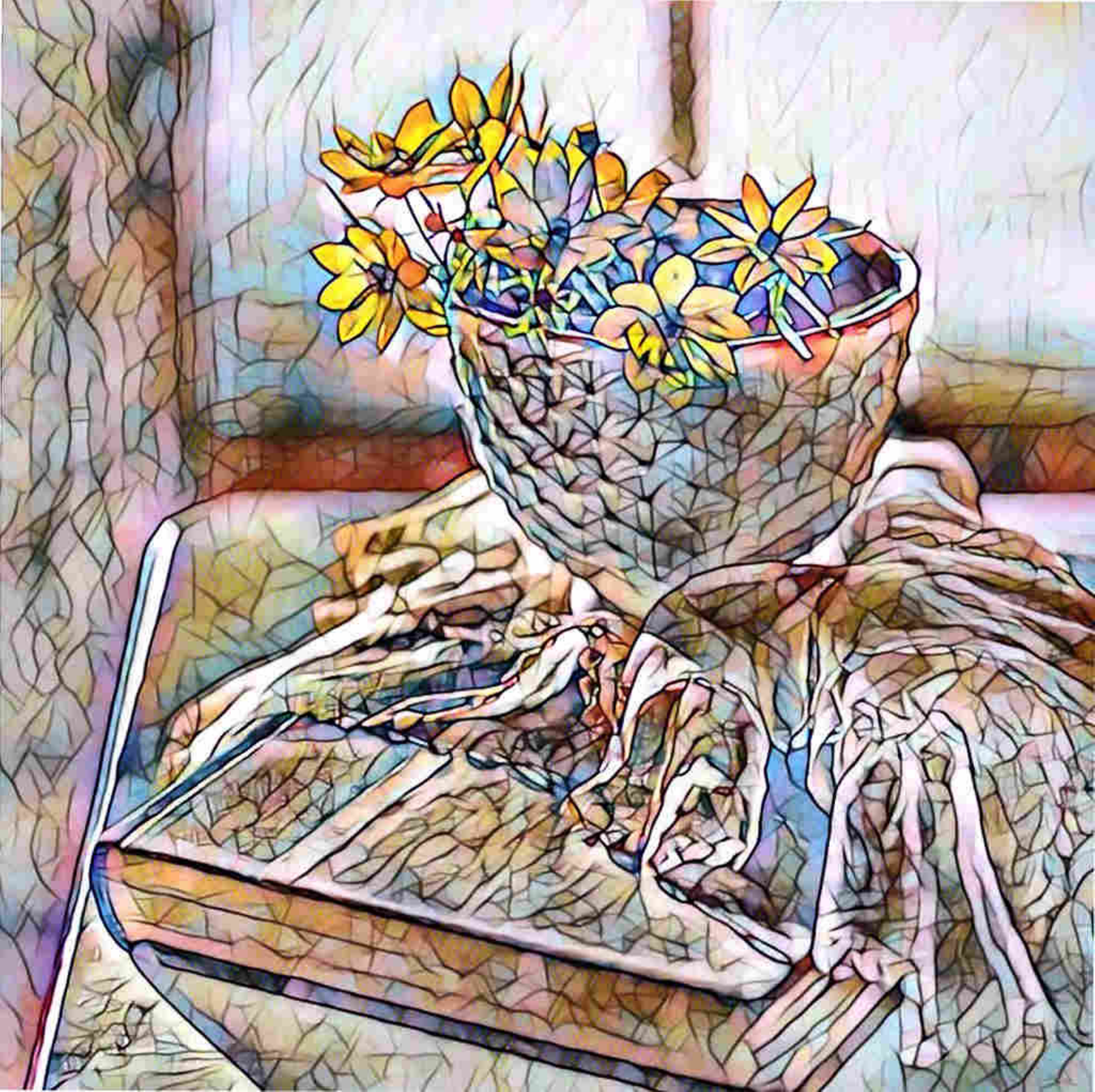} & && \includegraphics[width=\x\textwidth]{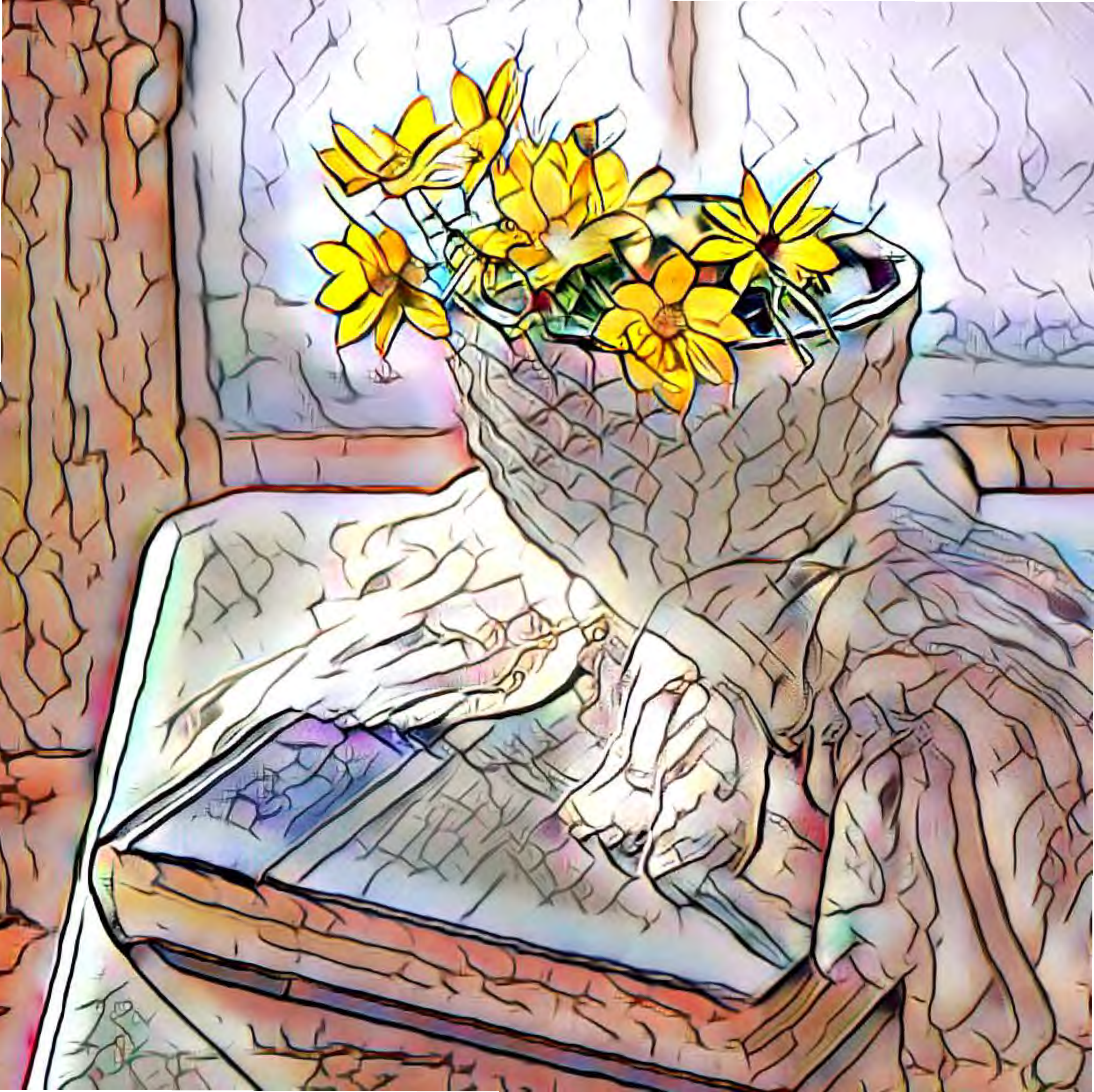} & \includegraphics[width=\x\textwidth]{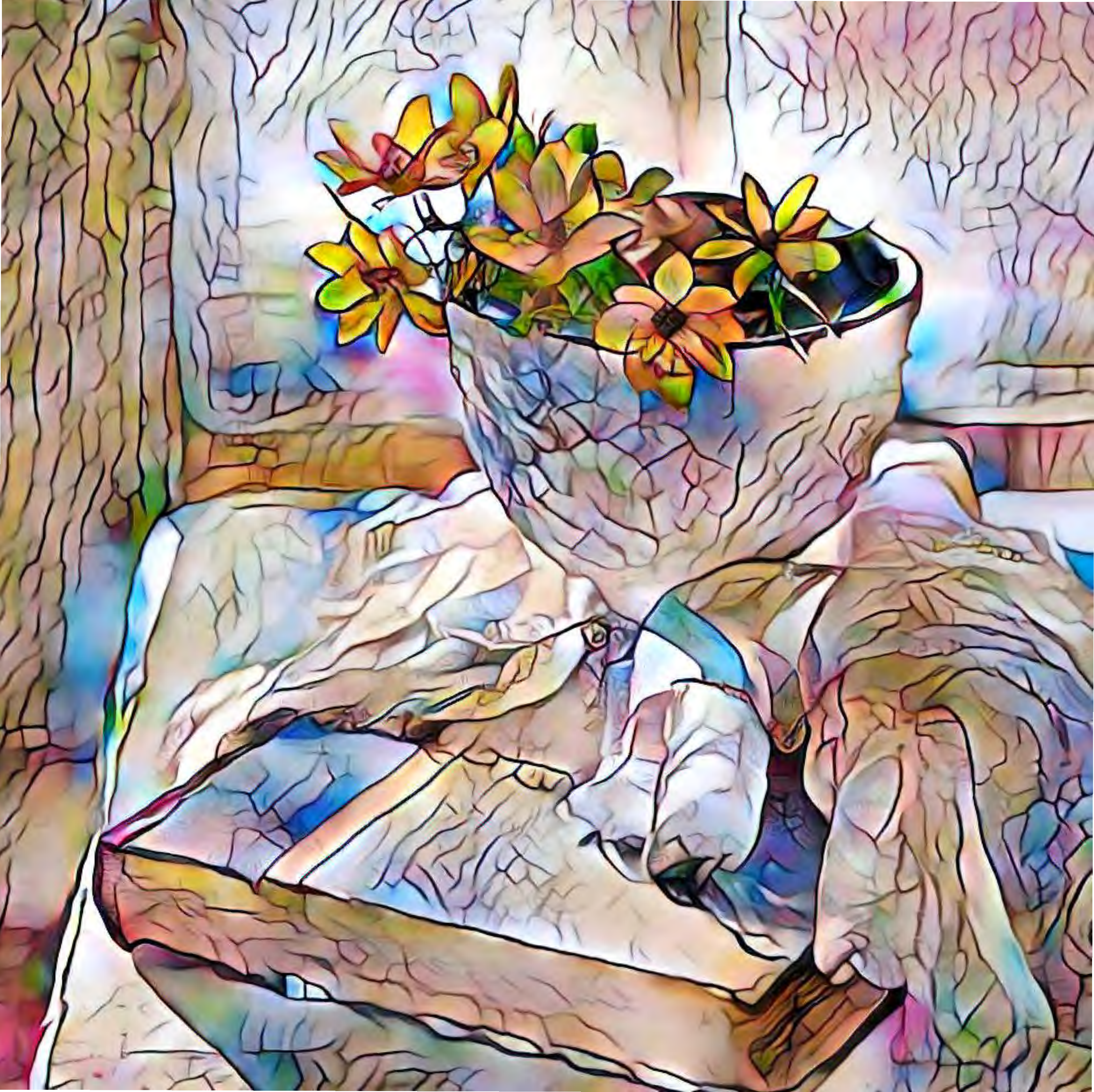}& \includegraphics[width=\x\textwidth]{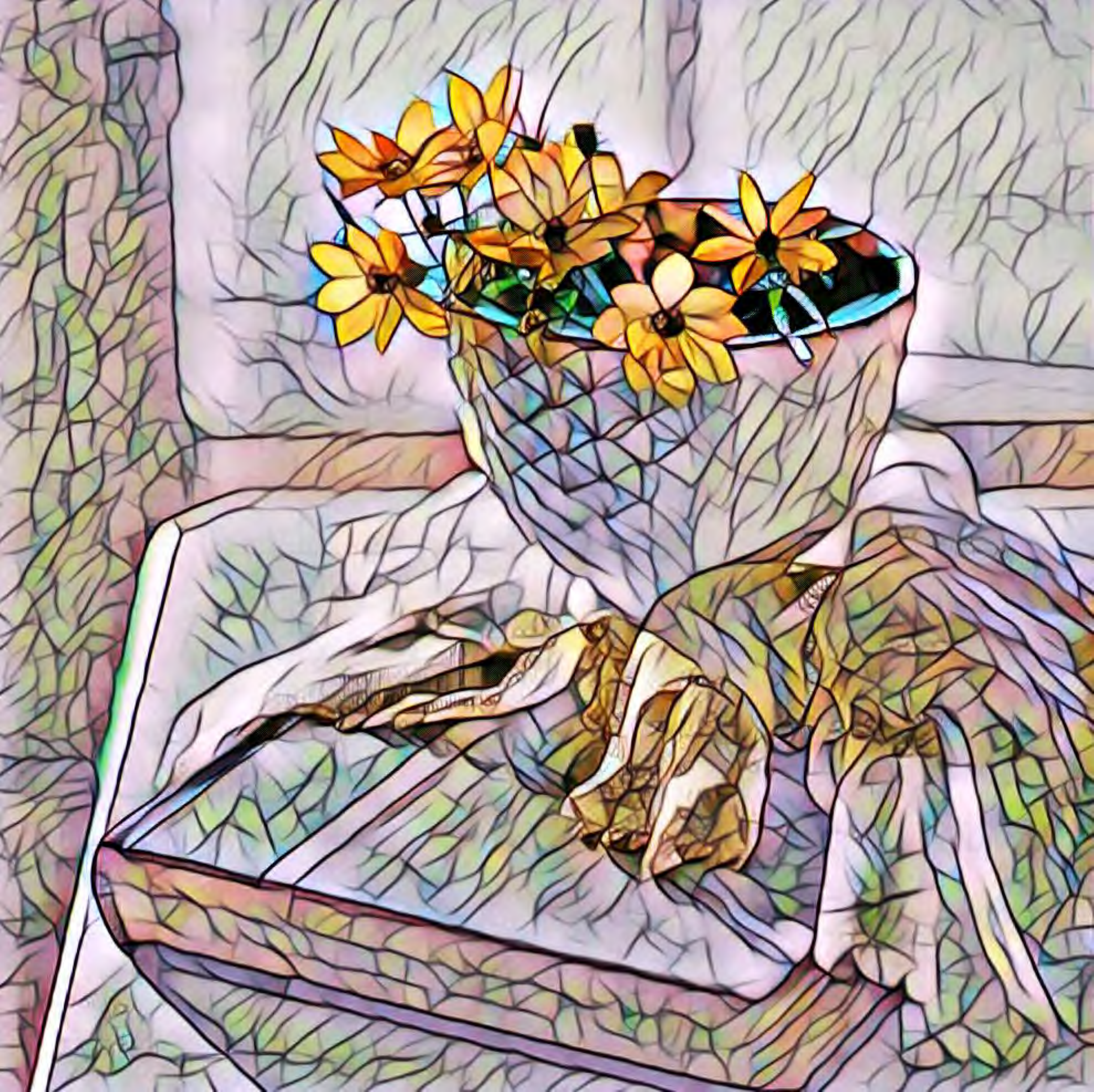} \\

\end{tabular}

%%%%%%%%%%%%%%%
}
%\smallskip
\caption{Some example results of different stroke sizes (SS) produced by our algorithm and other Fast Style Transfer algorithms. Each column represents the results of the same algorithm. The style image is the same with Figure~\ref{fig:compareprvious}.}
  \label{fig:qualitativeresult} %% label for entire figure
\end{figure*}
%%%%

%\newcommand\z{2.3cm}
%\newcommand\m{0.193}
%%%%%%%%
%\begin{figure*}[!t]
%\setlength\tabcolsep{1.6 pt}
%%\setlength\extrarowheight{1pt}
%{\renewcommand{\arraystretch}{1}
%\begin{tabular}{>{\centering}m{\z} >{\centering}m{\z} >{\centering}m{\z} >{\centering}m{\z} >{\centering\arraybackslash}m{\z}}
%\centering
%
%\includegraphics[width=\m\textwidth]{figs/000_11.pdf} &\includegraphics[width=\m\textwidth]{figs/015_11.pdf} & \includegraphics[width=\m\textwidth]{figs/030_11.pdf}& \includegraphics[width=\m\textwidth]{figs/042_11.pdf}& \includegraphics[width=\m\textwidth]{figs/059_11.pdf}\\
%\toprule
%\includegraphics[width=\m\textwidth]{figs/000d.png} &\includegraphics[width=\m\textwidth]{figs/015d.png} & \includegraphics[width=\m\textwidth]{figs/030d.png}& \includegraphics[width=\m\textwidth]{figs/042d.png}& \includegraphics[width=\m\textwidth]{figs/059d.png}\\
%
%
%%\footnotesize{(a) }&\footnotesize{(a) Relu1\_1} & \footnotesize{(b) Relu2\_1} & \footnotesize{(c) Relu3\_1} & \footnotesize{(d) Relu4\_1} & \footnotesize{(f) Relu5\_1} \\
%
%\end{tabular}
%}
%  \caption{Results of stroke interpolation. We zoom in on the same region (red frame) to observe the variations of stroke sizes.}
%  \label{fig:strokeinterpolation} %% label for entire figure
%\end{figure*}

\newcommand\w{1.87cm}
\newcommand\e{0.15}

%\newcommand\w{1.62cm}
%\newcommand\e{0.135}

%%%%%%%
\begin{figure}[!t]
\setlength\tabcolsep{0pt}
{\renewcommand{\arraystretch}{0.2}
%\begin{minipage}{\textwidth}
%\begin{tabular}{ccccccc}

\begin{tabular}{m{0.8cm} >{\centering}m{\w}  >{\centering}m{\w} >{\centering}m{\w} >{\centering}m{\w} >{\centering}m{\w} >{\centering\arraybackslash}m{\w}}

\textbf{\tiny{\cite{huang2017arbitrary}:}}&\includegraphics[width=\e\textwidth]{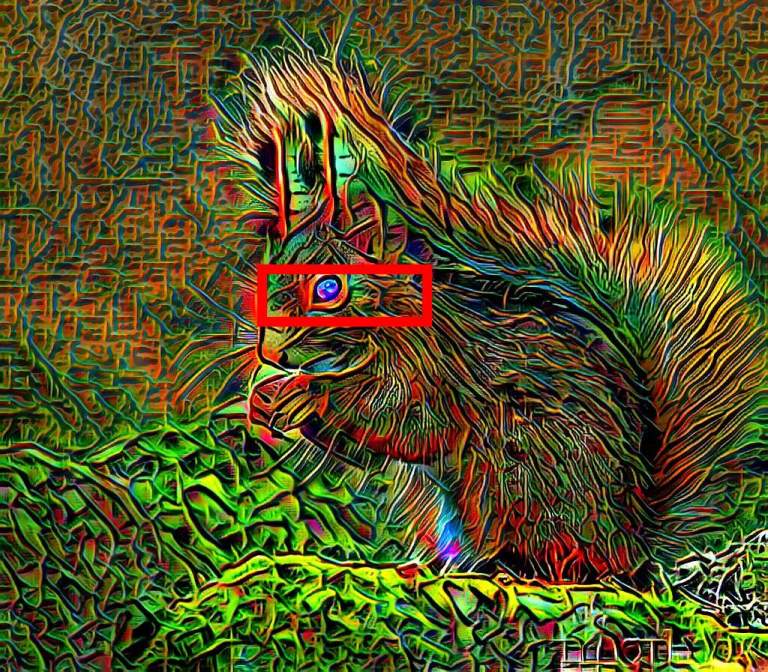} &\includegraphics[width=\e\textwidth]{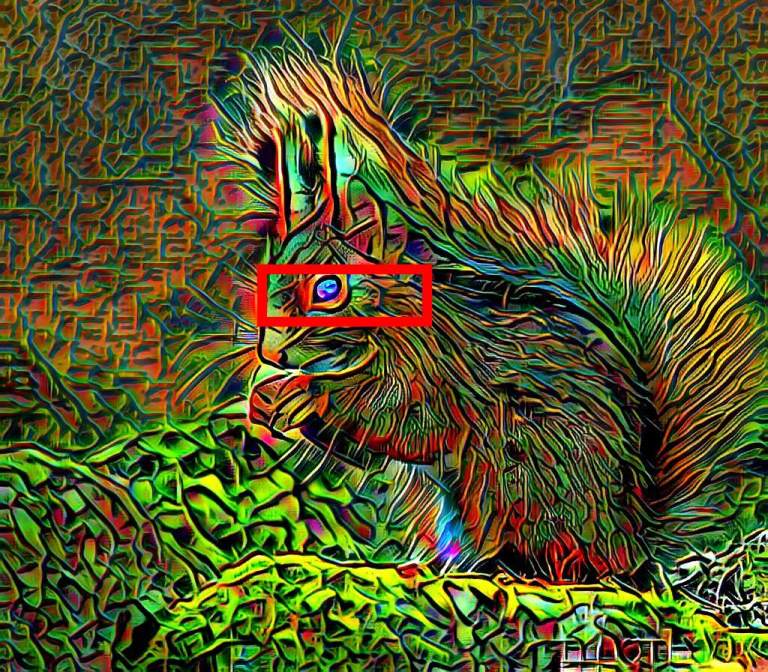} & \includegraphics[width=\e\textwidth]{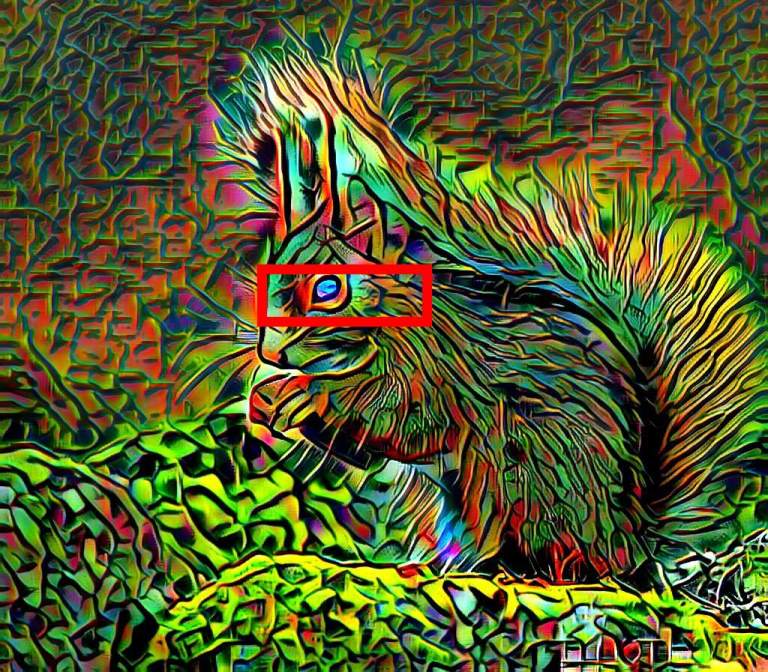}& \includegraphics[width=\e\textwidth]{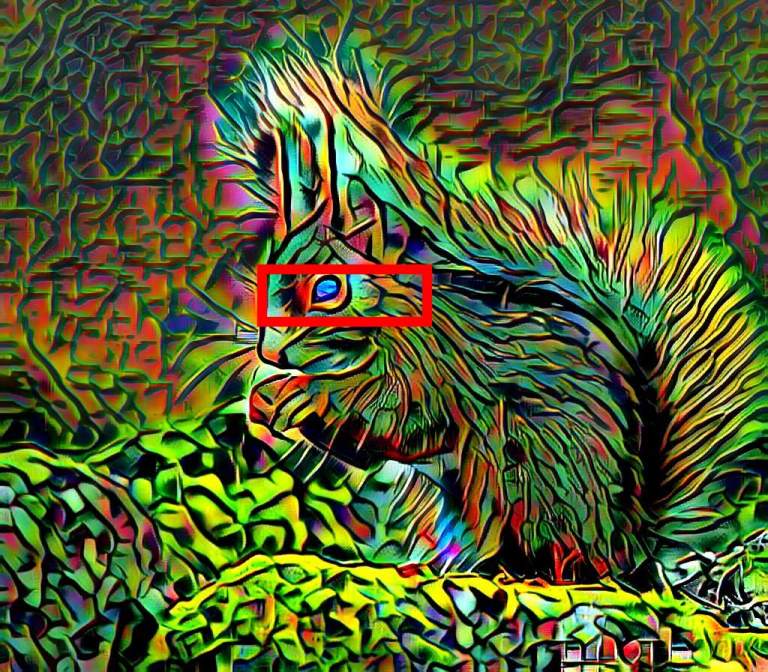}& \includegraphics[width=\e\textwidth]{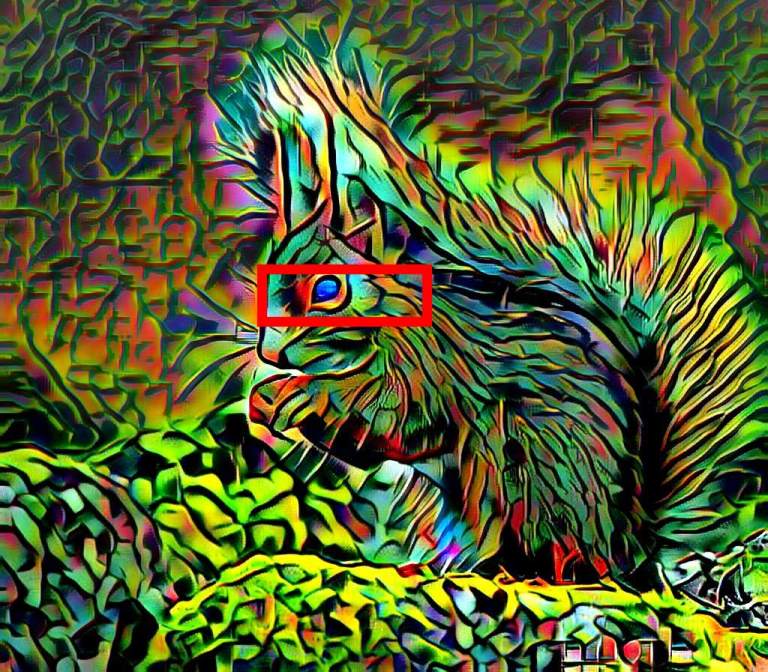}& \includegraphics[width=\e\textwidth]{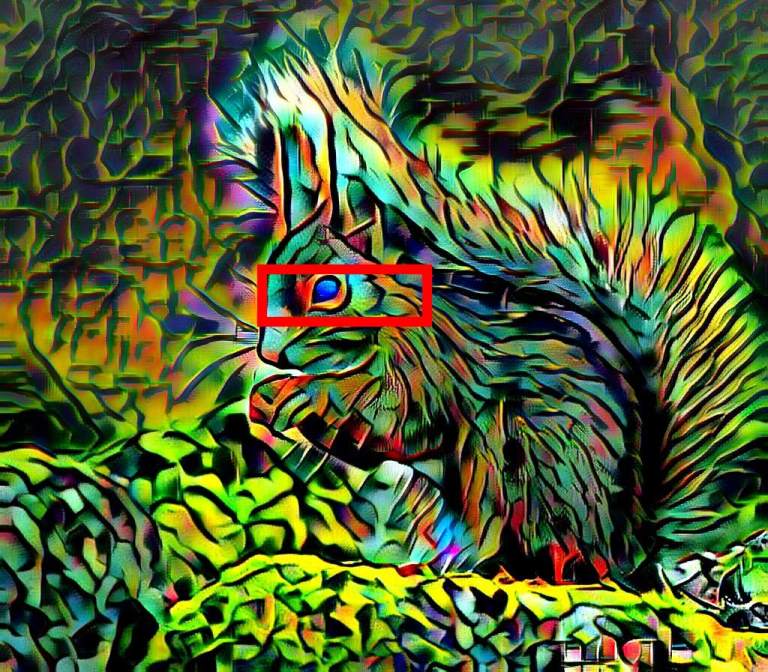}\\

&\includegraphics[width=\e\textwidth]{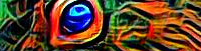} &\includegraphics[width=\e\textwidth]{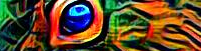}&\includegraphics[width=\e\textwidth]{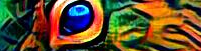}&\includegraphics[width=\e\textwidth]{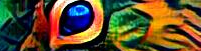}
&\includegraphics[width=\e\textwidth]{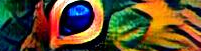}&\includegraphics[width=\e\textwidth]{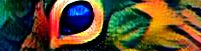}\\

\end{tabular}

\begin{tabular}{m{0.8cm} >{\centering}m{\w}  >{\centering}m{\w} >{\centering}m{\w} >{\centering}m{\w} >{\centering}m{\w} >{\centering\arraybackslash}m{\w}}

\textbf{\tiny{\cite{li2017universal}:}}&\includegraphics[width=\e\textwidth]{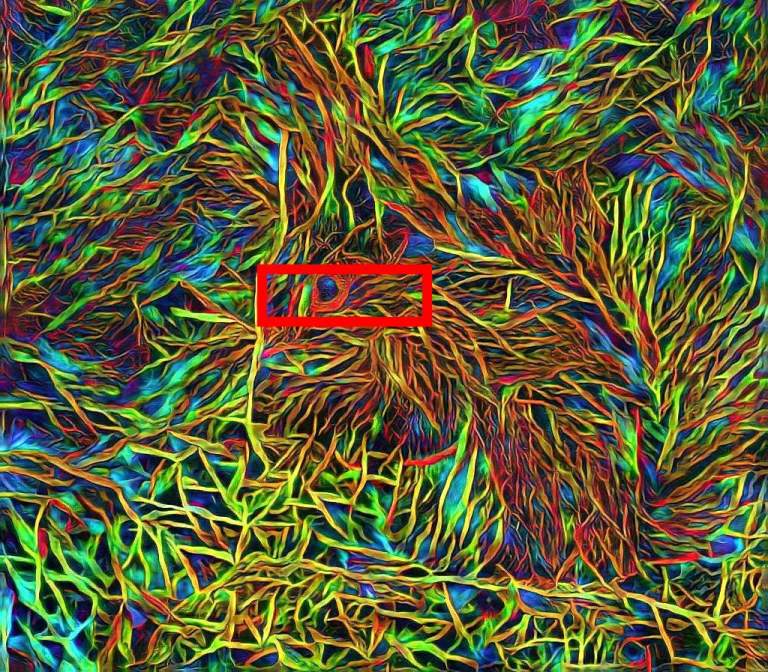} &\includegraphics[width=\e\textwidth]{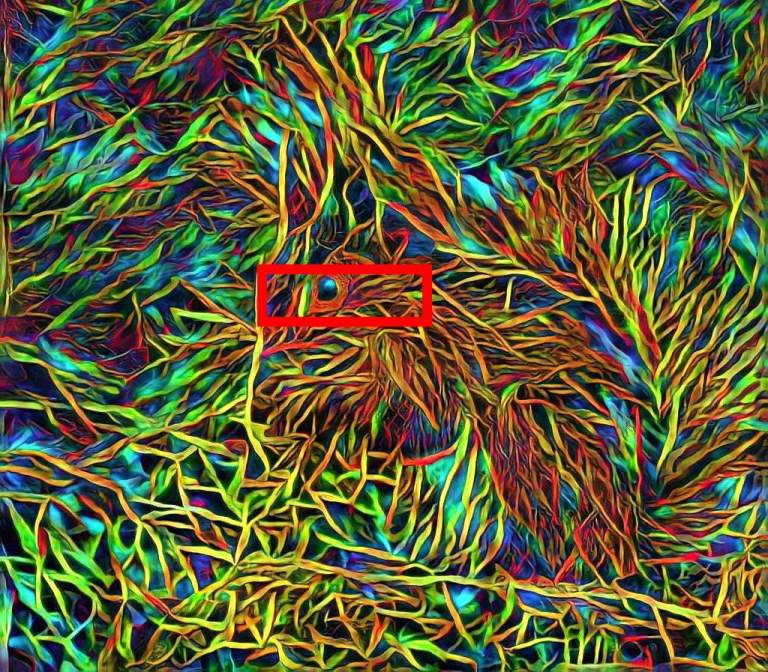} & \includegraphics[width=\e\textwidth]{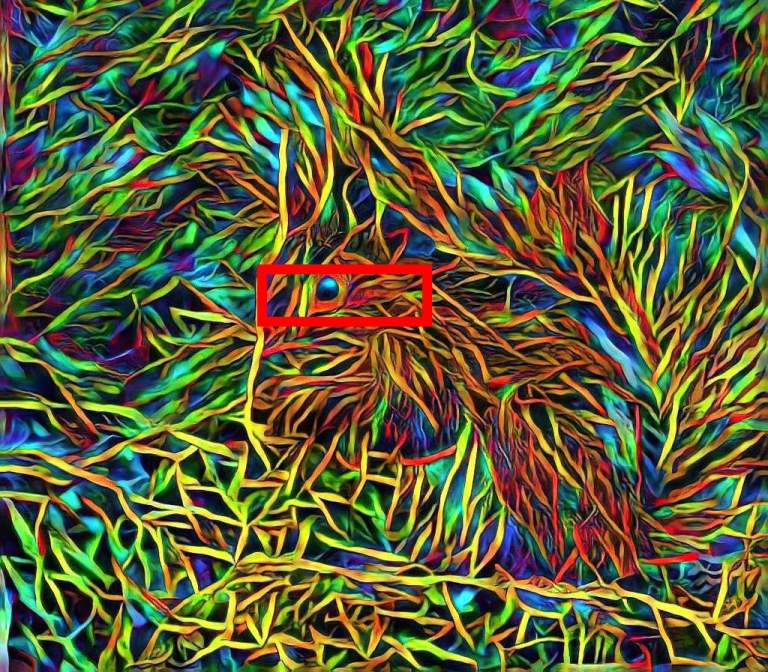}& \includegraphics[width=\e\textwidth]{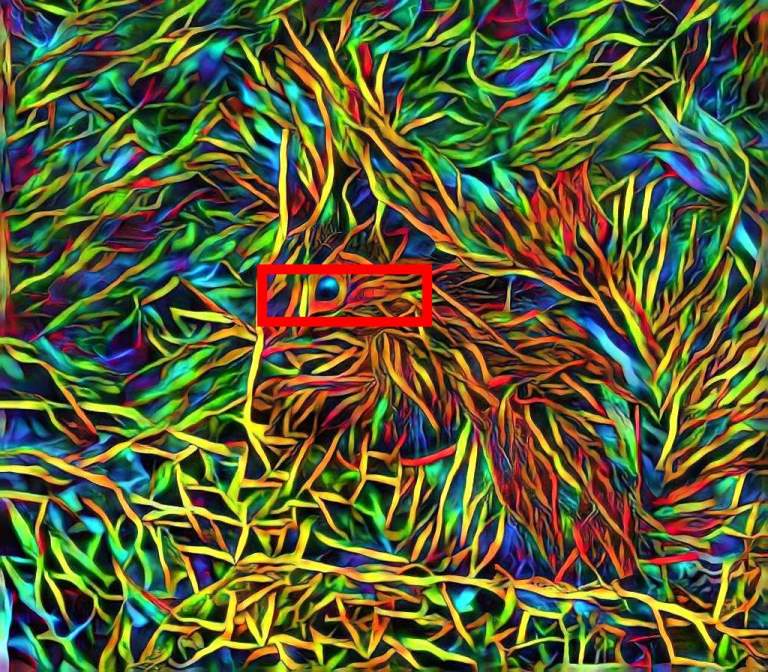}& \includegraphics[width=\e\textwidth]{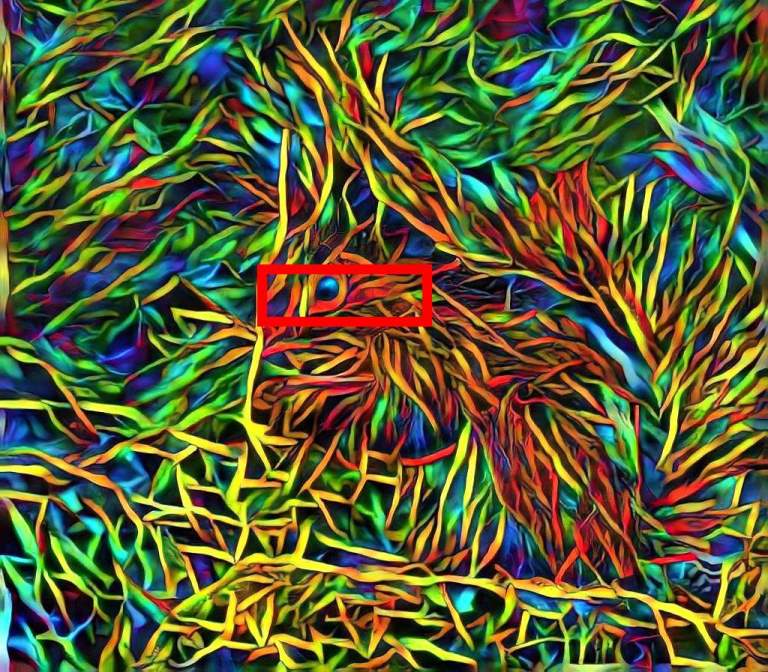}& \includegraphics[width=\e\textwidth]{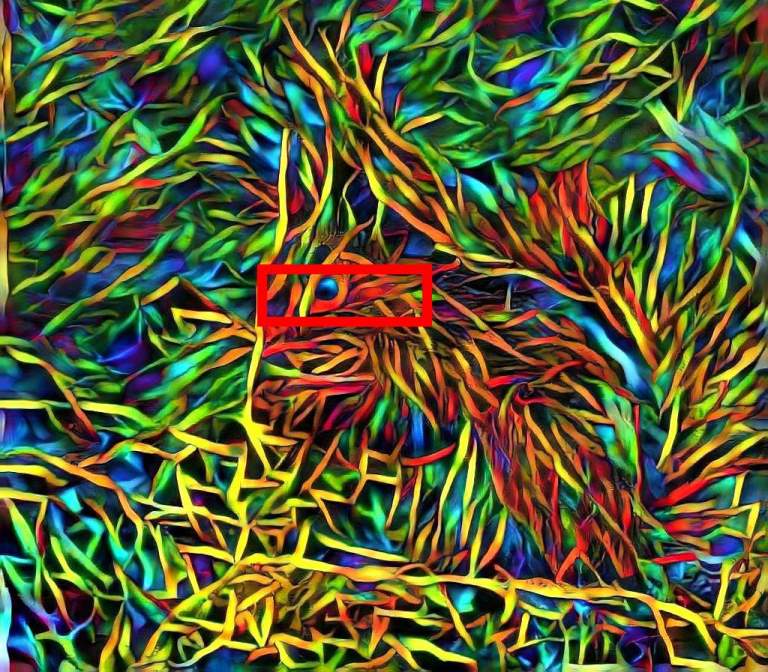}\\

&\includegraphics[width=\e\textwidth]{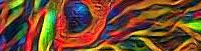} &\includegraphics[width=\e\textwidth]{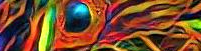}&\includegraphics[width=\e\textwidth]{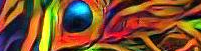}&\includegraphics[width=\e\textwidth]{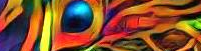}
&\includegraphics[width=\e\textwidth]{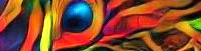}&\includegraphics[width=\e\textwidth]{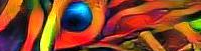}\\

\end{tabular}

\begin{tabular}{m{0.8cm} >{\centering}m{\w}  >{\centering}m{\w} >{\centering}m{\w} >{\centering}m{\w} >{\centering}m{\w} >{\centering\arraybackslash}m{\w}}

\textbf{\tiny{Ours:}}&\includegraphics[width=\e\textwidth]{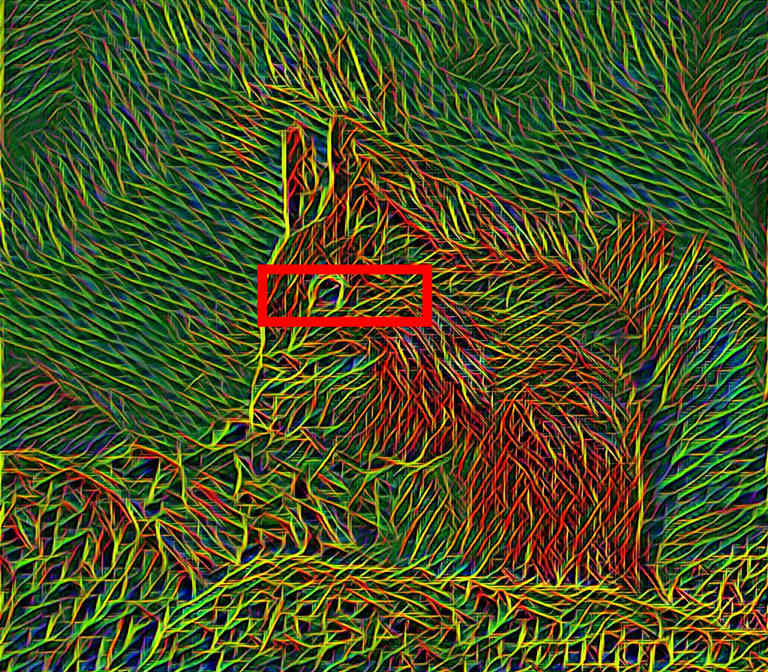} &\includegraphics[width=\e\textwidth]{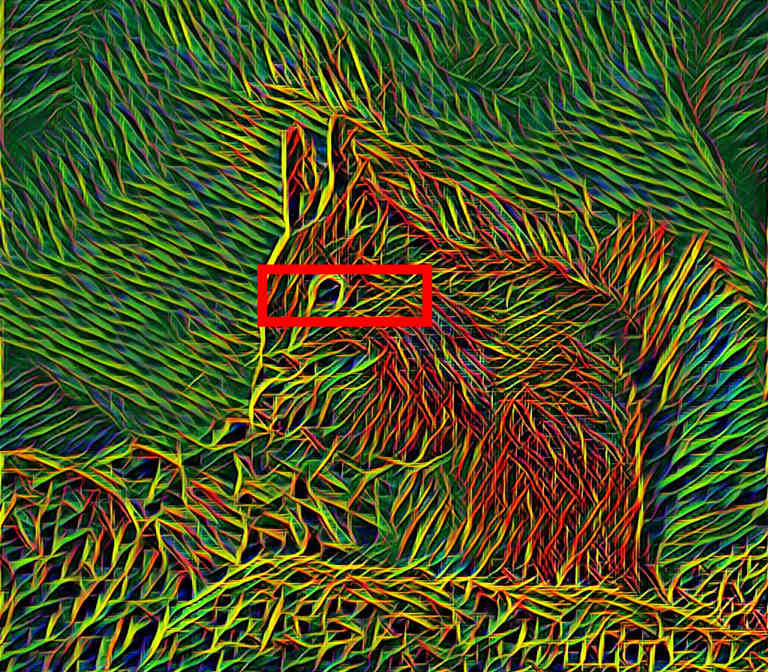} & \includegraphics[width=\e\textwidth]{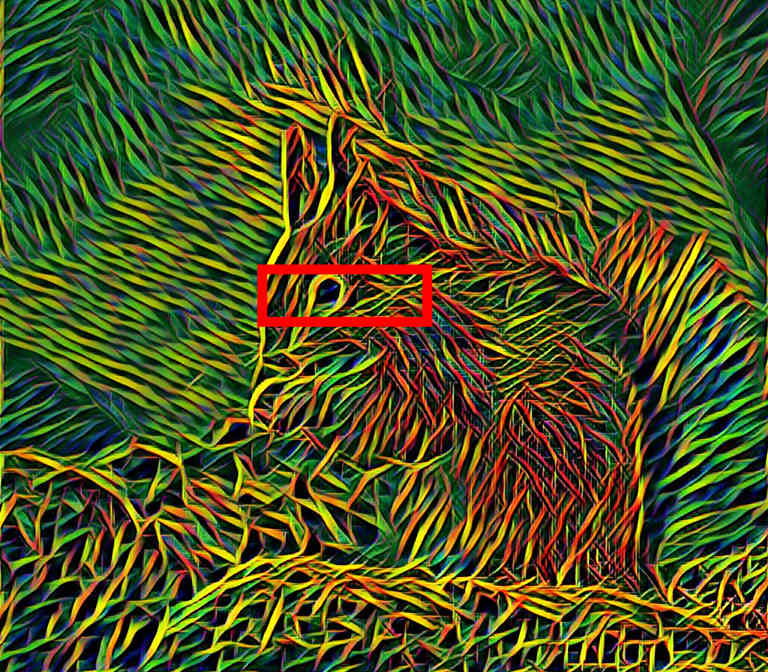}& \includegraphics[width=\e\textwidth]{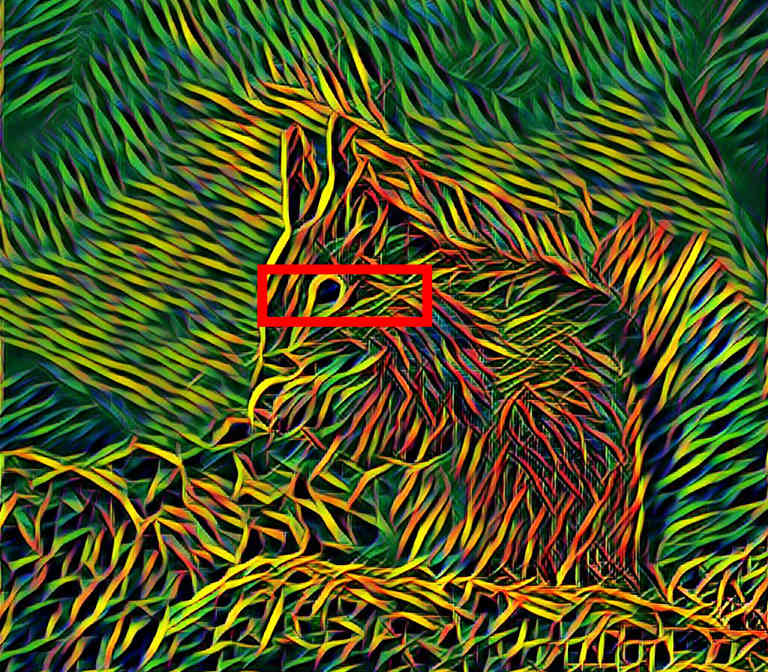}& \includegraphics[width=\e\textwidth]{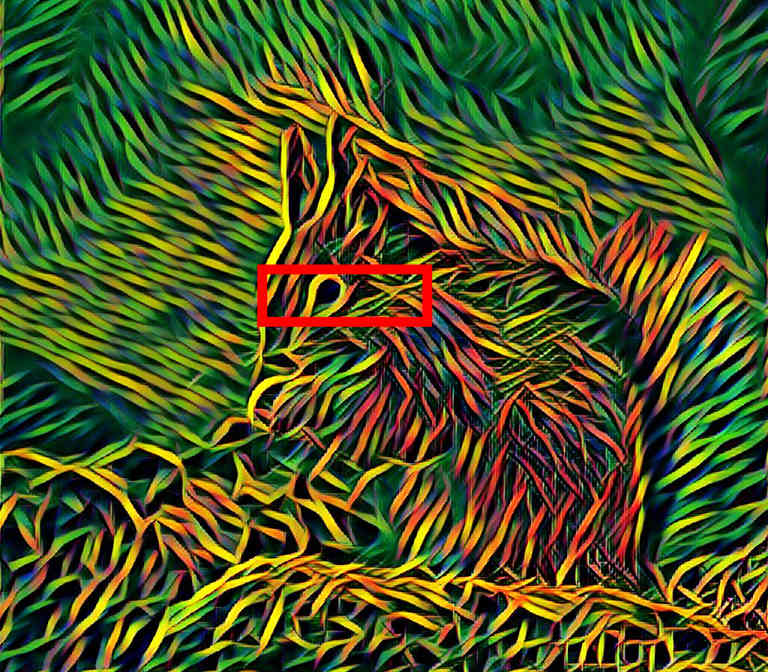}& \includegraphics[width=\e\textwidth]{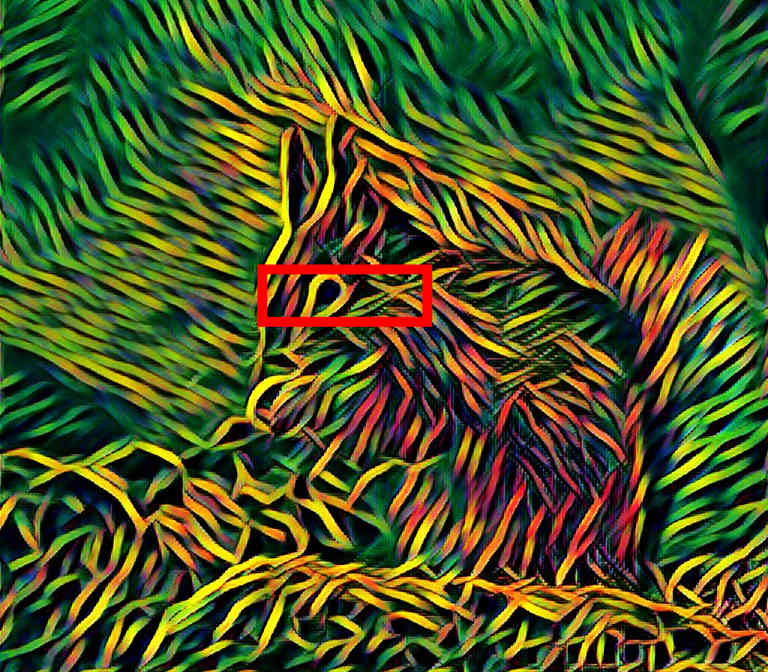}\\

%\textbf{\tiny{Possible:}}&\includegraphics[width=\e\textwidth]{figs/interpolation/possible/1_rectangle.jpg} &\includegraphics[width=\e\textwidth]{figs/interpolation/possible/2_rectangle.jpg} & \includegraphics[width=\e\textwidth]{figs/interpolation/possible/3_rectangle.jpg}& \includegraphics[width=\e\textwidth]{figs/interpolation/possible/4_rectangle.jpg}& \includegraphics[width=\e\textwidth]{figs/interpolation/possible/5_rectangle.jpg}& \includegraphics[width=\e\textwidth]{figs/interpolation/possible/6_rectangle.jpg}\\

&\includegraphics[width=\e\textwidth]{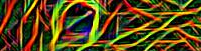} &\includegraphics[width=\e\textwidth]{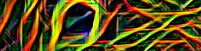}&\includegraphics[width=\e\textwidth]{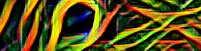}&\includegraphics[width=\e\textwidth]{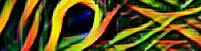}
&\includegraphics[width=\e\textwidth]{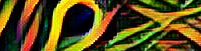}&\includegraphics[width=\e\textwidth]{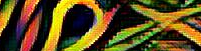} \\

%&\includegraphics[width=\e\textwidth]{figs/interpolation/possible/1_detail.png} &\includegraphics[width=\e\textwidth]{figs/interpolation/possible/2_detail.png}&\includegraphics[width=\e\textwidth]{figs/interpolation/possible/3_detail.png}&\includegraphics[width=\e\textwidth]{figs/interpolation/possible/4_detail.png}
%&\includegraphics[width=\e\textwidth]{figs/interpolation/possible/5_detail.png}&\includegraphics[width=\e\textwidth]{figs/interpolation/possible/6_detail.png} \\

%\textbf{\tiny{Pixel Interpolation}}:&\includegraphics[width=\e\textwidth]{figs/interpolation/pixel/1_rectangle.jpg} &\includegraphics[width=\e\textwidth]{figs/interpolation/pixel/2_rectangle.jpg} & \includegraphics[width=\e\textwidth]{figs/interpolation/pixel/3_rectangle.jpg}& \includegraphics[width=\e\textwidth]{figs/interpolation/pixel/4_rectangle.jpg}& \includegraphics[width=\e\textwidth]{figs/interpolation/pixel/5_rectangle.jpg}& \includegraphics[width=\e\textwidth]{figs/interpolation/pixel/6_rectangle.jpg}\\

%&\includegraphics[width=\e\textwidth]{figs/interpolation/pixel/1_detail.png} &\includegraphics[width=\e\textwidth]{figs/interpolation/pixel/2_detail.png}&\includegraphics[width=\e\textwidth]{figs/interpolation/pixel/3_detail.png}&\includegraphics[width=\e\textwidth]{figs/interpolation/pixel/4_detail.png}
%&\includegraphics[width=\e\textwidth]{figs/interpolation/pixel/5_detail.png}&\includegraphics[width=\e\textwidth]{figs/interpolation/pixel/6_detail.png} \\

\end{tabular}

}
  \caption{Results of continuous stroke size control. We zoom in on the same region (red frame) to observe the variations of stroke sizes. Our algorithm produces finer strokes and details. The content and style image can be found in Figure~\ref{fig:arch}.}
  \label{fig:strokeinterpolation} %% label for entire figure
\end{figure}

\begin{figure}[!t]
\setlength\tabcolsep{0 pt}
{\renewcommand{\arraystretch}{0.8}
%\begin{tabular}{>{\centering}n{\p} >{\centering}n{\p} >{\centering\arraybackslash}n{\p}}
%\begin{tabular}{>{\centering}m{1.98cm} >{\centering}m{1.98cm} >{\centering}m{1.98cm} ?{0.2mm} >{\centering}m{1.98cm} >{\centering}m{1.98cm} >{\centering\arraybackslash}m{1.98cm}}
%\begin{tabular}{cc?{0.2mm}cc?{0.2mm}cc}

\begin{tabular}{m{0.8cm} >{\centering}m{\w}  >{\centering}m{\w}?{0.2mm} >{\centering}m{\w} >{\centering}m{\w} ?{0.2mm}>{\centering}m{\w} >{\centering\arraybackslash}m{\w}}

&\includegraphics[width=\e\textwidth]{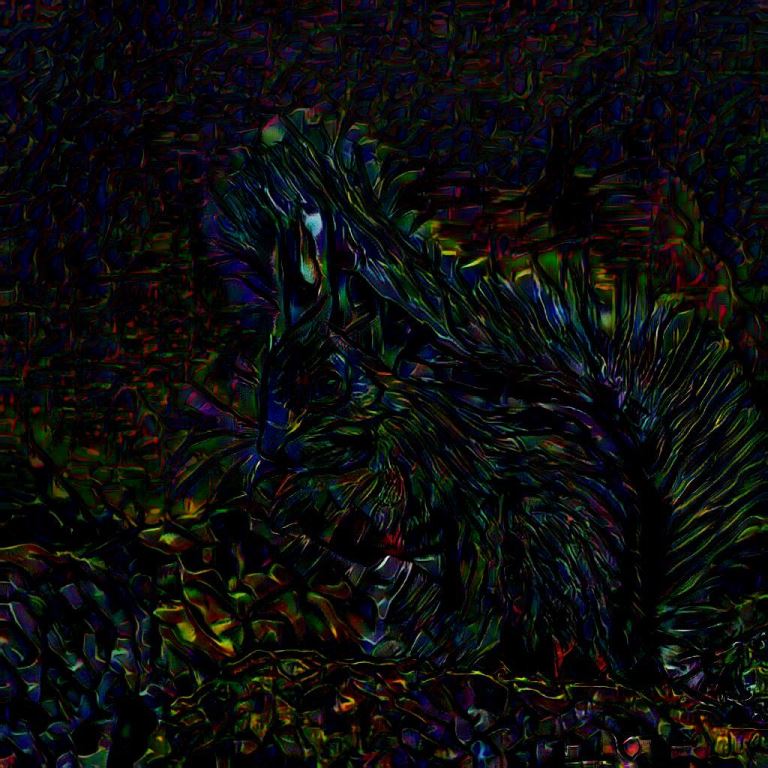} &\includegraphics[width=\e\textwidth]{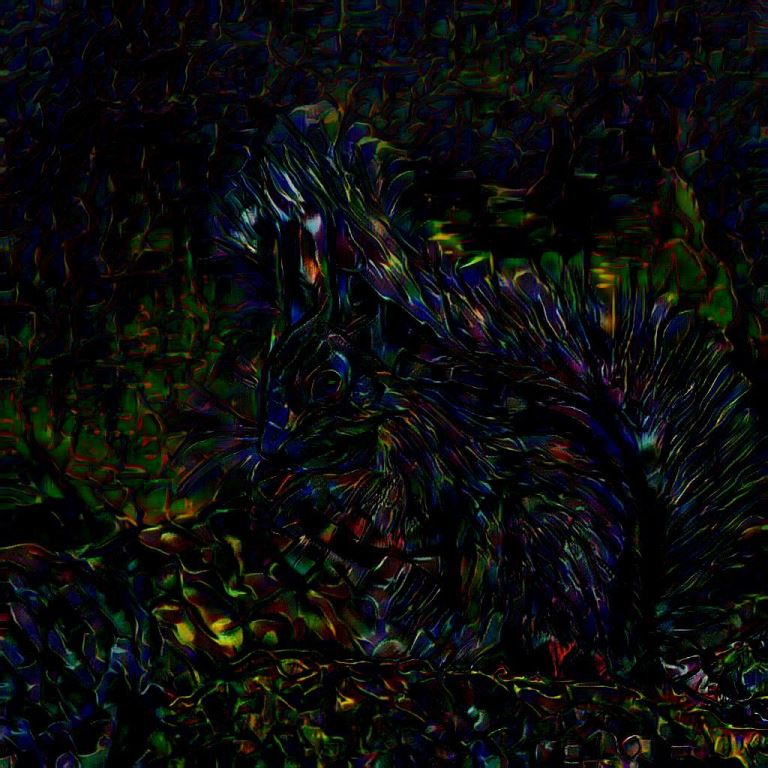} & \includegraphics[width=\e\textwidth]{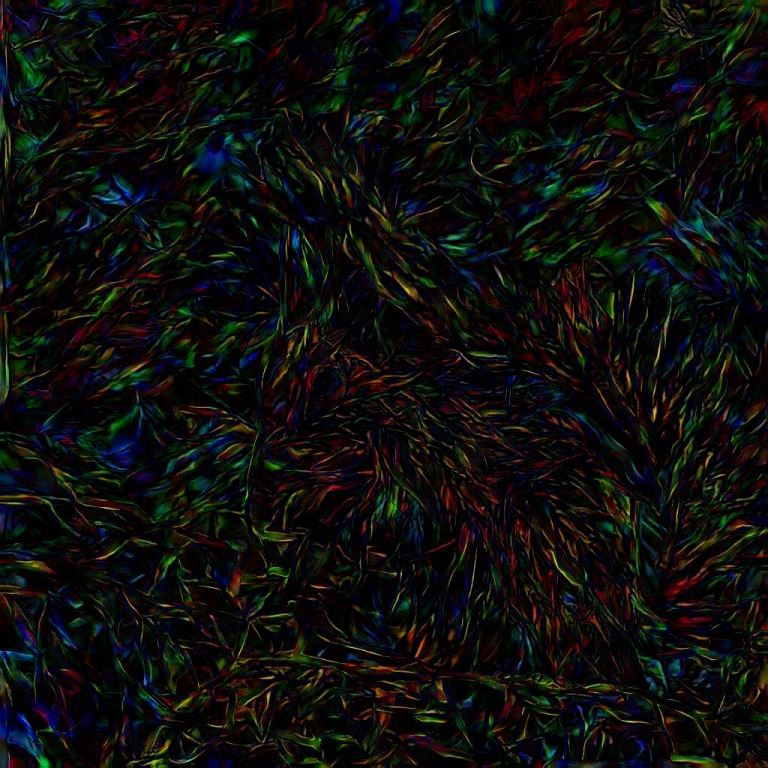}& \includegraphics[width=\e\textwidth]{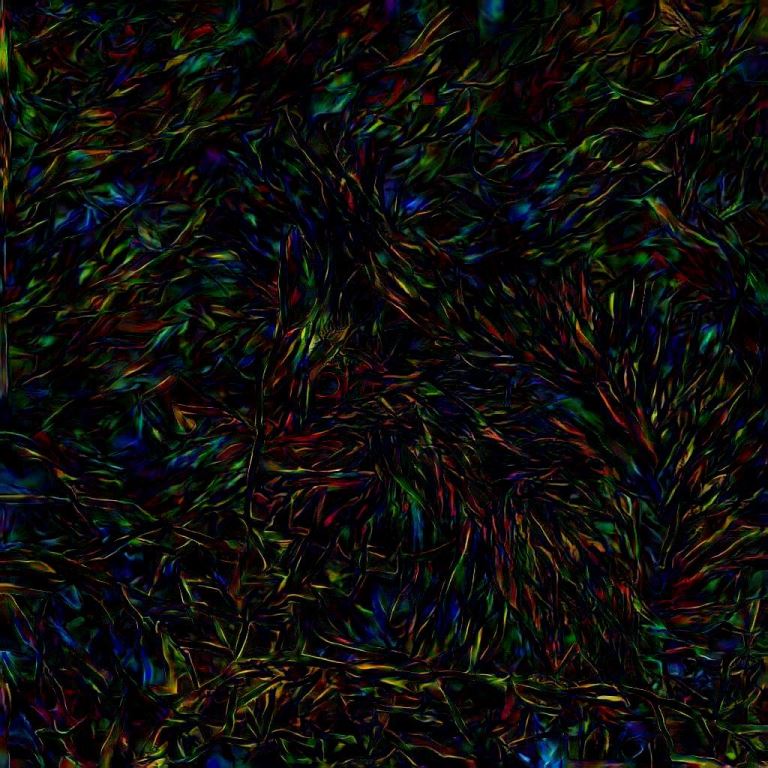}& \includegraphics[width=\e\textwidth]{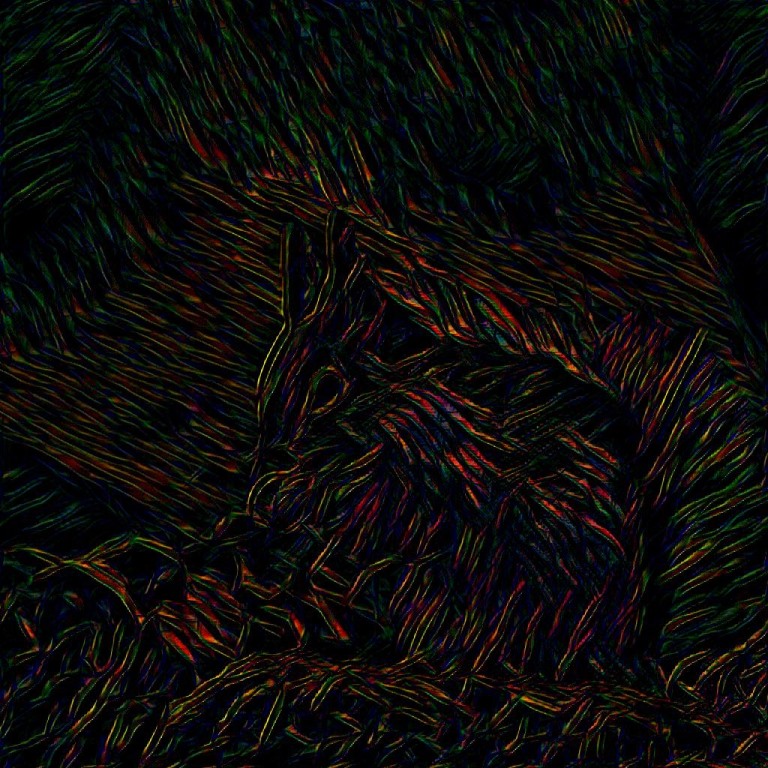}& \includegraphics[width=\e\textwidth]{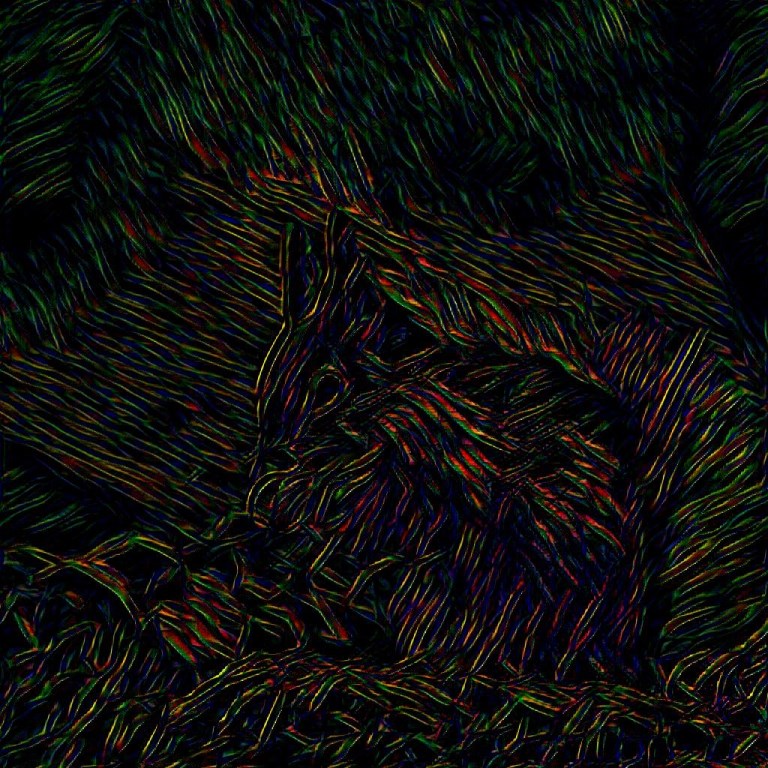}\\

&\multicolumn{2}{c?{0.2mm}}{\scriptsize{\cite{huang2017arbitrary}'s differential images}} &\multicolumn{2}{c?{0.2mm}}{\scriptsize{\cite{li2017universal}'s differential images}} &\multicolumn{2}{c}{\scriptsize{Our differential images}}\\

\end{tabular}

}

%\vspace{-0.18cm}
%\vspace{-0.01cm}
\caption{Results of the absolute differences of adjacent images in each row of Figure~\ref{fig:strokeinterpolation}.}
\label{fig:difference} %% label for entire figure
\end{figure}

\begin{figure}[!t]
\setlength\tabcolsep{0.5pt}
{\renewcommand{\arraystretch}{0.8}
%\begin{tabular}{>{\centering}n{\p} >{\centering}n{\p} >{\centering\arraybackslash}n{\p}}
\begin{tabular}{>{\centering}m{3cm} >{\centering}m{3cm} >{\centering}m{3cm} >{\centering\arraybackslash}m{3cm}}

\includegraphics[width=0.24\textwidth]{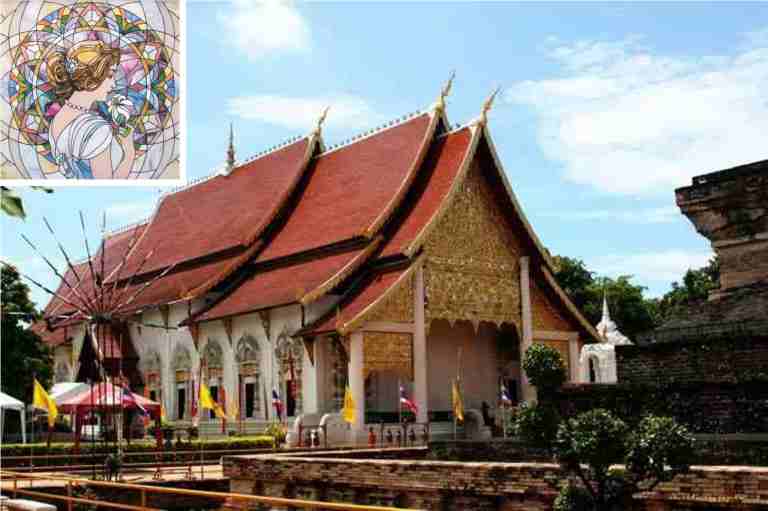}& \includegraphics[width=0.24\textwidth]{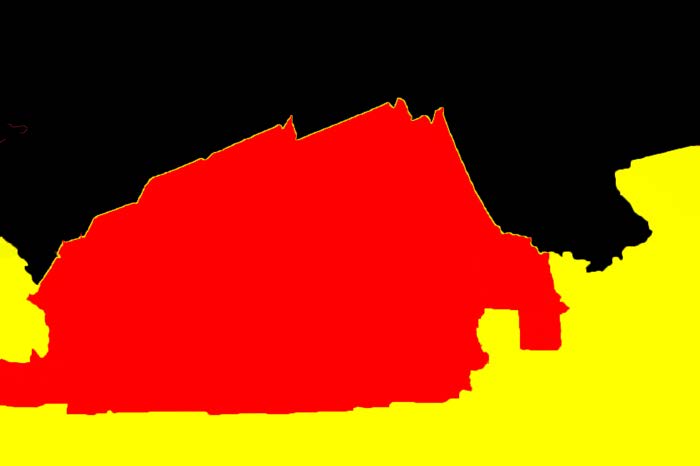} &\includegraphics[width=0.24\textwidth]{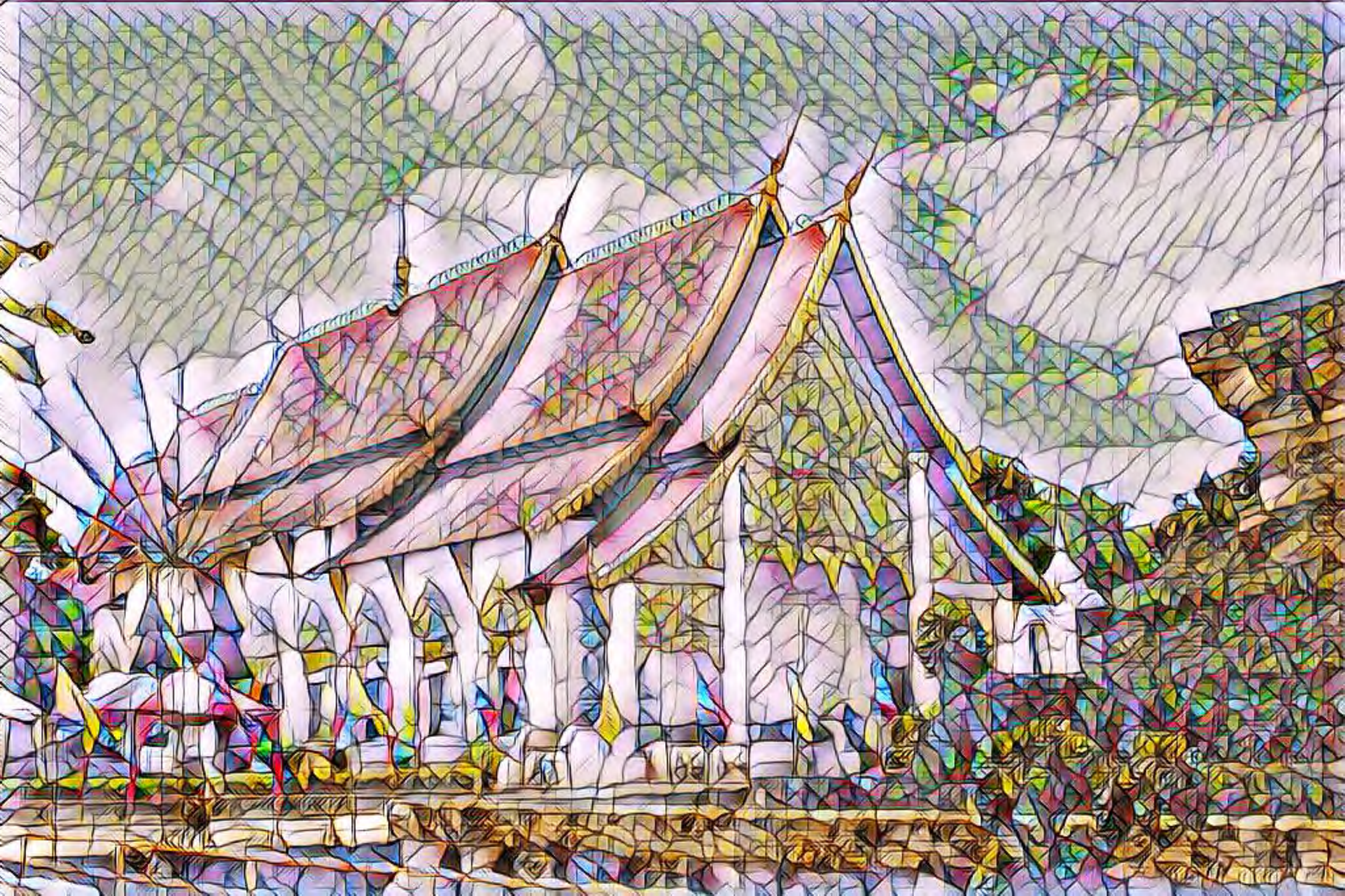} & \includegraphics[width=0.24\textwidth]{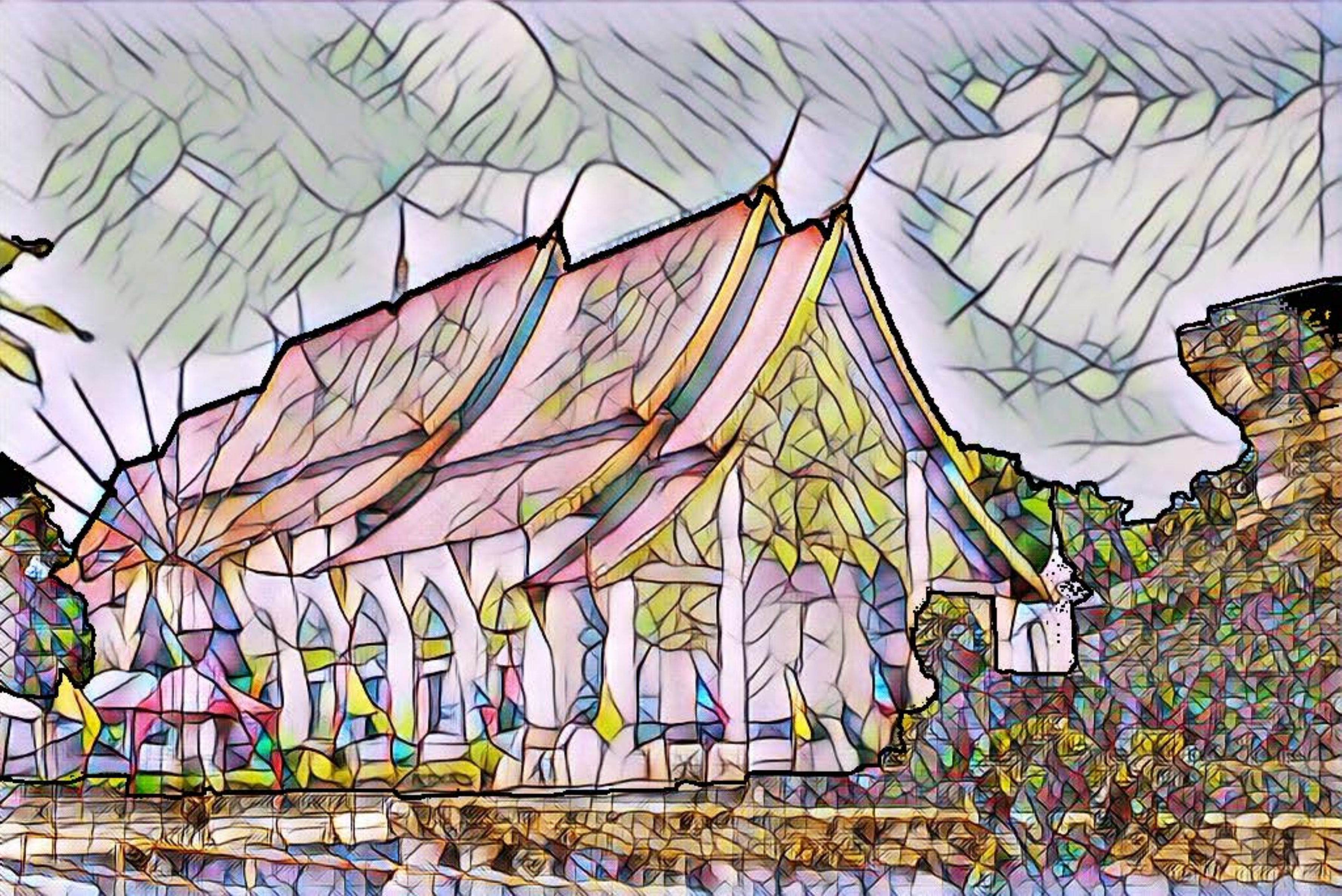} \\

\includegraphics[width=0.24\textwidth]{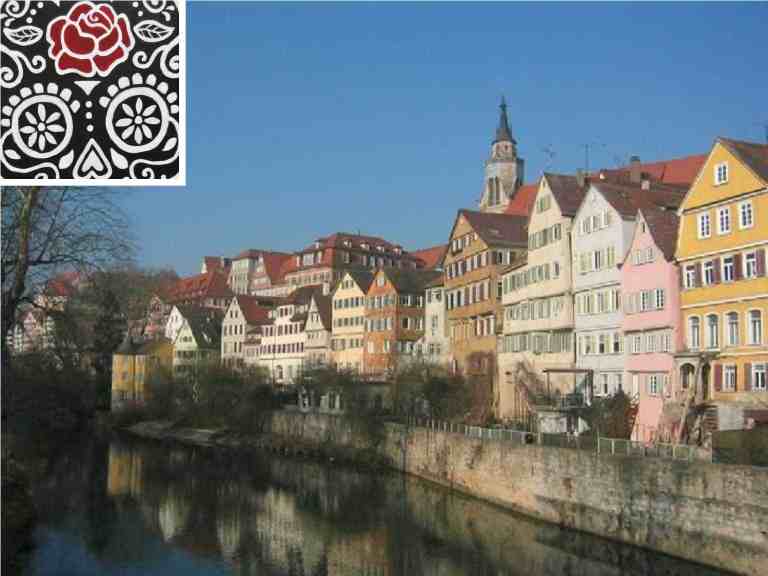}& \includegraphics[width=0.24\textwidth]{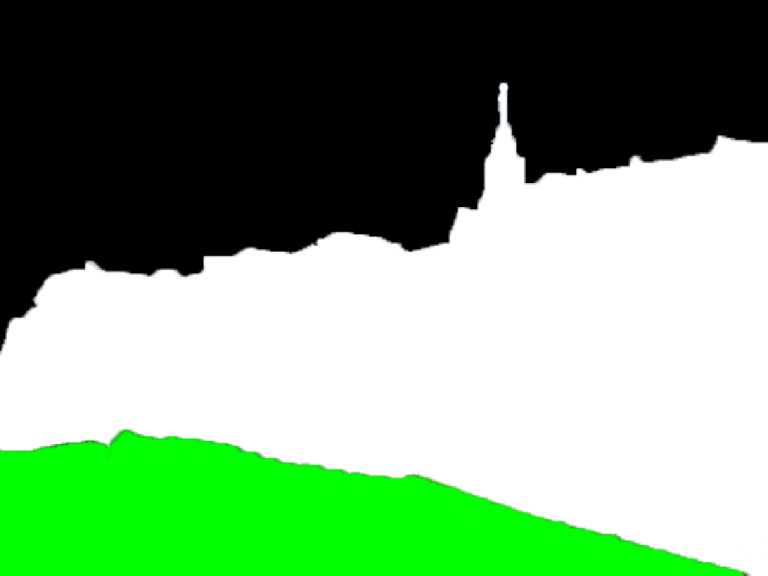} &\includegraphics[width=0.24\textwidth]{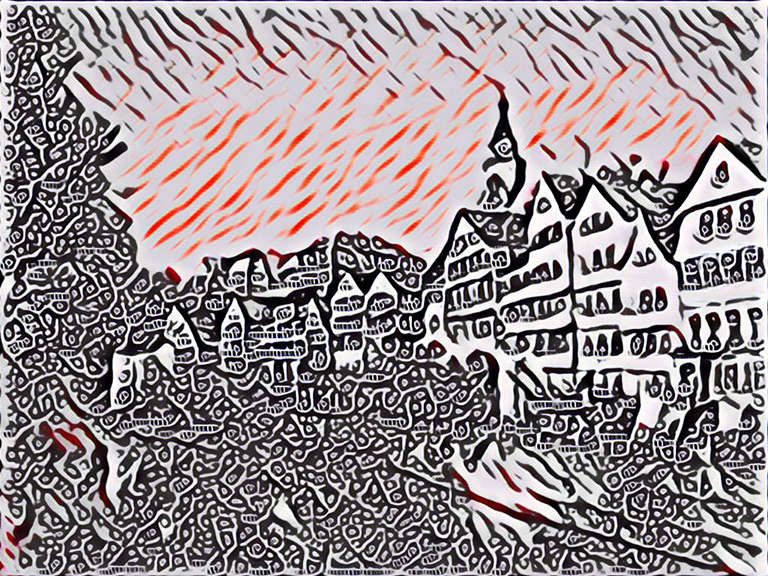} & \includegraphics[width=0.24\textwidth]{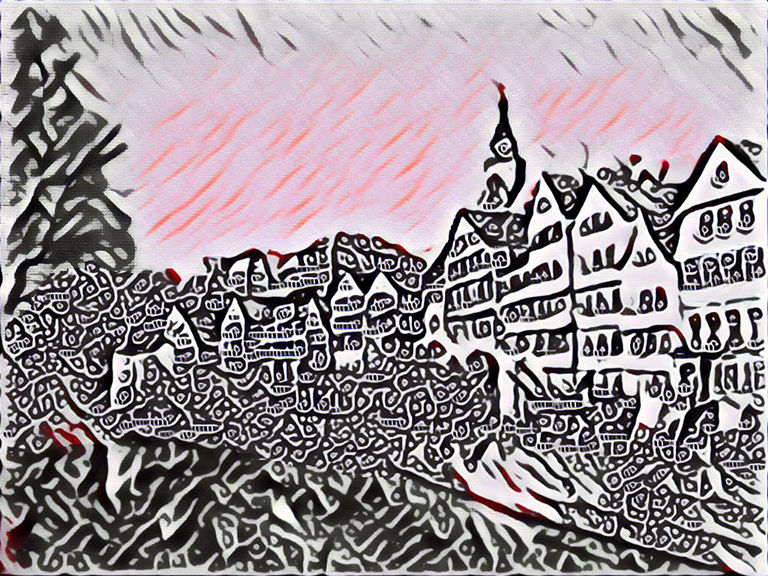} \\

\includegraphics[width=0.24\textwidth]{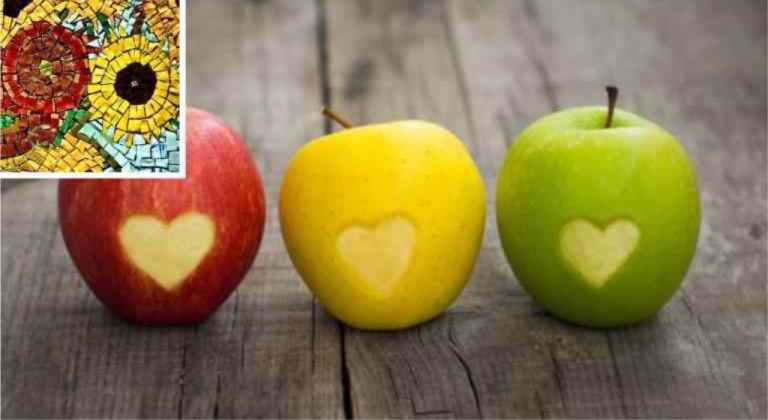}& \includegraphics[width=0.24\textwidth]{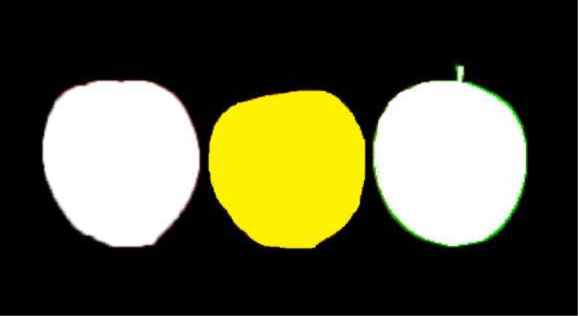} &\includegraphics[width=0.24\textwidth]{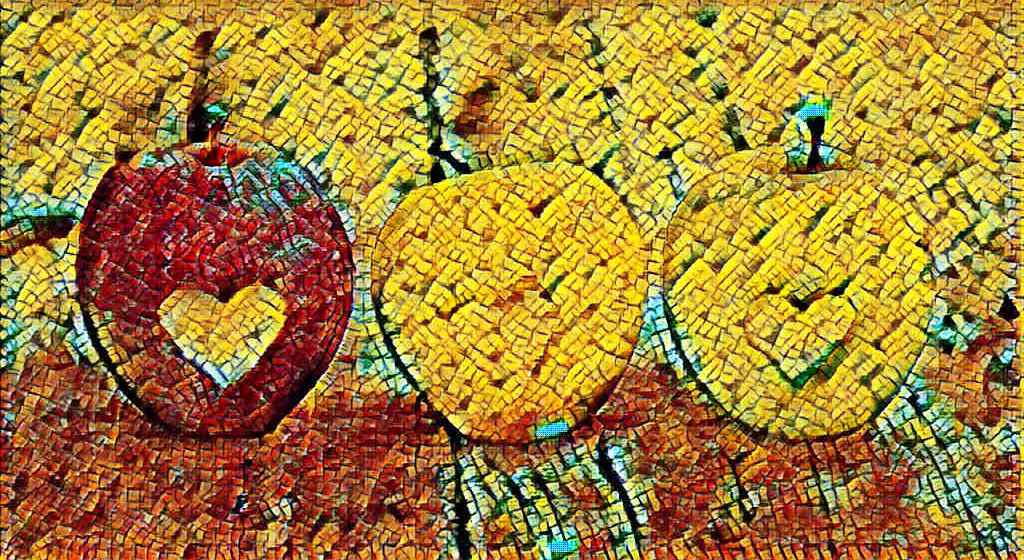} & \includegraphics[width=0.24\textwidth]{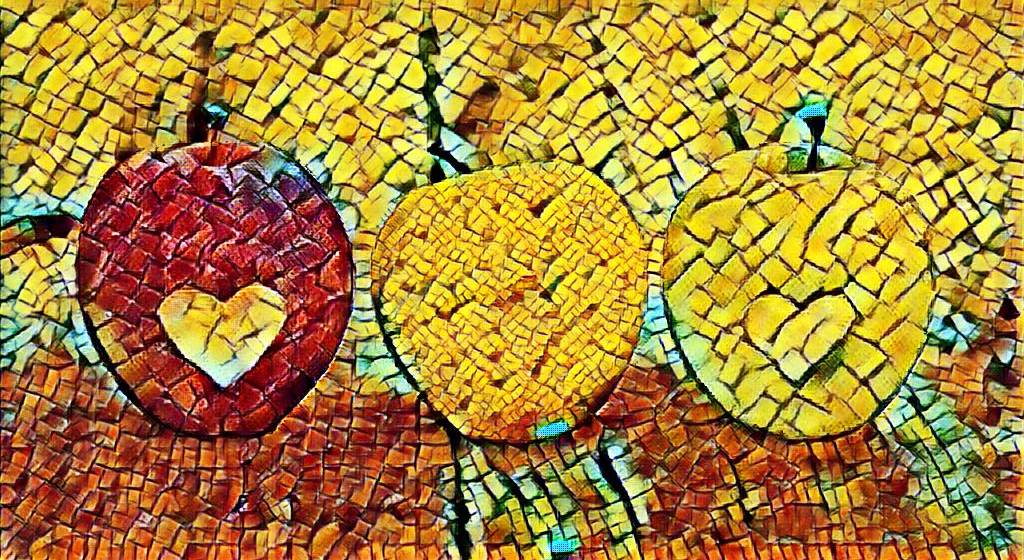} \\

\scriptsize{(a) Content \& Style\\} & \scriptsize{(b) Content mask\\} &\scriptsize{(c) Same stroke size across image}& \scriptsize{(d) Our spatial stroke size control}\\

% \footnotesize{(a) Training a \\separate generator}&\footnotesize{(b) Image resizing + forwarding + SR} & \footnotesize{(c) Our proposed approach}\\

\end{tabular}
}
%\smallskip
\caption{Our algorithm allows flexible spatial stroke size control during stylization. The result produced by our single model can have mixed stroke sizes, which is more consistent with an artist's artwork in reality.}
%\myLcomment{draw the difference graph}
\label{fig:mix} %% label for entire figure
\end{figure}

% * <brooksong@ieee.org> 2017-11-09T08:51:28.410Z:
%
% > results (Figure~\ref{fig:qualitativeresult}
% 这个地方请注意，原来是参考文献15.
%
% ^ <398372213@qq.com> 2017-11-12T01:30:28.070Z:
%
% 好的，之前写错了..
%
% ^ <brooksong@ieee.org> 2017-11-13T01:42:25.411Z.

%Our algorithm is capable of producing appealing results while preserving flexibility for stroke size control.

%Let $\mathcal{T}_i \in \mathbb{T}$ denote different stroke sizes, $\mathbb{T}$ denote the set of all stroke sizes, and $I^{\mathcal{T}_i}$ represent an image $I$ composed of lots of stroke textons with size $\mathcal{T}_i$. The aforementioned relation between statistical distribution $p(I_o)$ and style image $I_s$ with different $\mathcal{T}_i$ can be then presented as:

%With the same amount of parameters withthe previous Fast Style Transfer algorithms \cite{Johnson2016perceptual,ulyanov2017improved}, our work demonstrates that one single generator is capable of learning multiple stroke sizes.
%%%%%

%\begin{figure}
%%  \centering
%  \subfigure[Content loss]{
%    %\label{fig:subfig:a} %% label for first subfigure
%    \includegraphics[width=0.23\textwidth]{figs/finalcontentloss_new.pdf}}
%  %\hspace{0.1in}
%  \subfigure[Style loss]{
%    %\label{fig:subfig:a} %% label for first subfigure
%    \includegraphics[width=0.23\textwidth]{figs/finalstyleloss_new.pdf}}
%  %\hspace{0.1in}
%
%  \caption{Comparisons of the average content and style loss of our algorithm with state-of-the-art Neural Style Transfer algorithms.}
%  \label{fig:finalloss} %% label for entire figure
%\end{figure}

\subsection{Quantitative Evaluation}
%\myLcomment{compare the loss curve. Demonstrate that different branches improve each other. compare the final loss with other algorithms. verify the architecture chosen?}

\begin{figure}[!tp]
\centering
%\begin{tabular}{>{\centering}n{\p} >{\centering}n{\p} >{\centering\arraybackslash}n{\p}}
\begin{minipage}{\textwidth}
\begin{tabular}{>{\centering}m{5.7cm} >{\centering\arraybackslash}m{2cm}}
%\begin{tabular}{cc}
%\begin{minipage}{0.1\textwidth}
%\end{minipage}

\begin{minipage}{0.5\textwidth}
\centering
\begin{minipage}{\textwidth}
\includegraphics[width=0.95\textwidth]{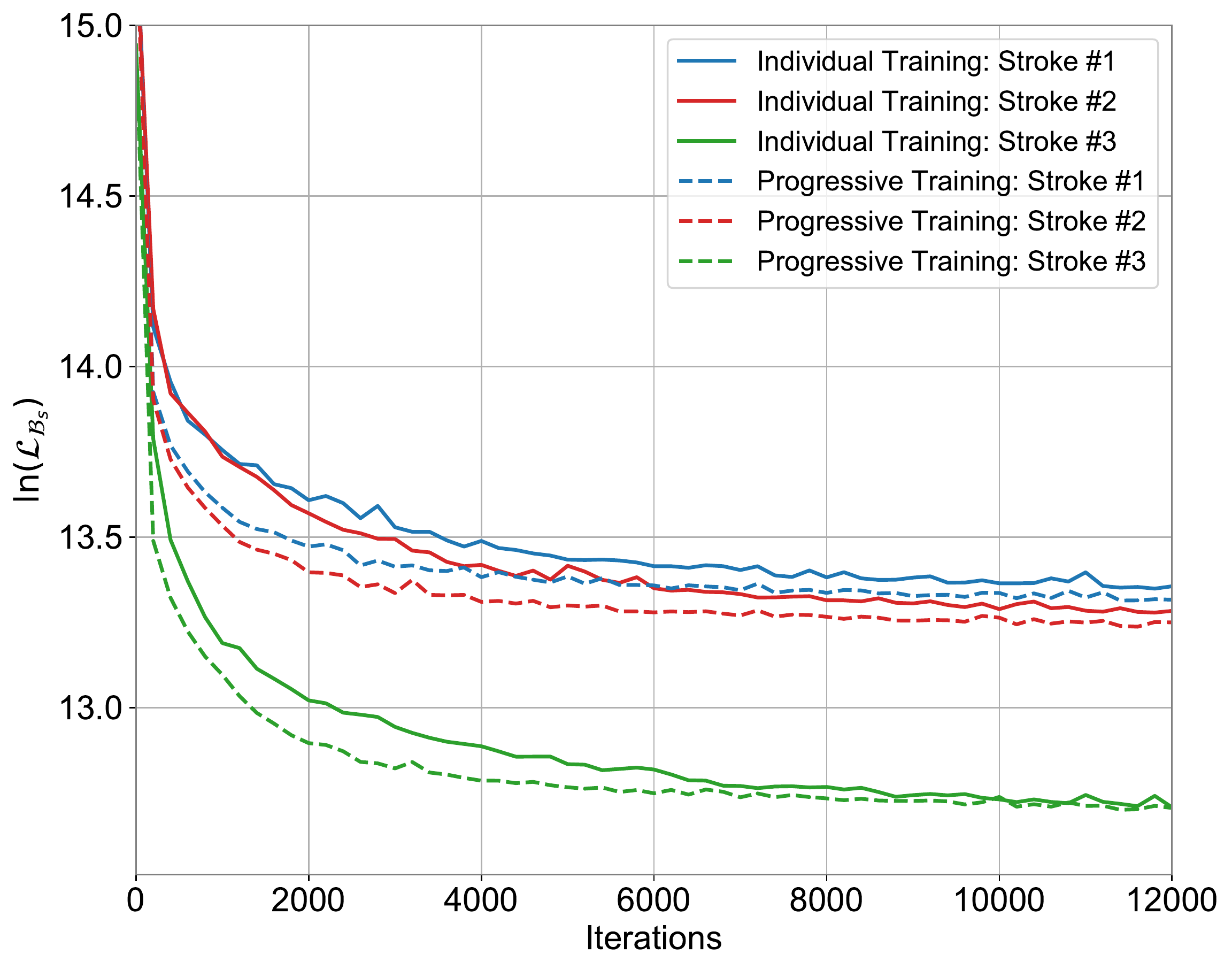}
\end{minipage}

\begin{minipage}{\textwidth}
\centering
\footnotesize{(a) Progressive training.}
\end{minipage}

%\footnotesize{(a) Progressive learning.}
\end{minipage}

&
\begin{minipage}{0.5\textwidth}

\begin{minipage}{\textwidth}
\centering
\includegraphics[width=0.9\textwidth]{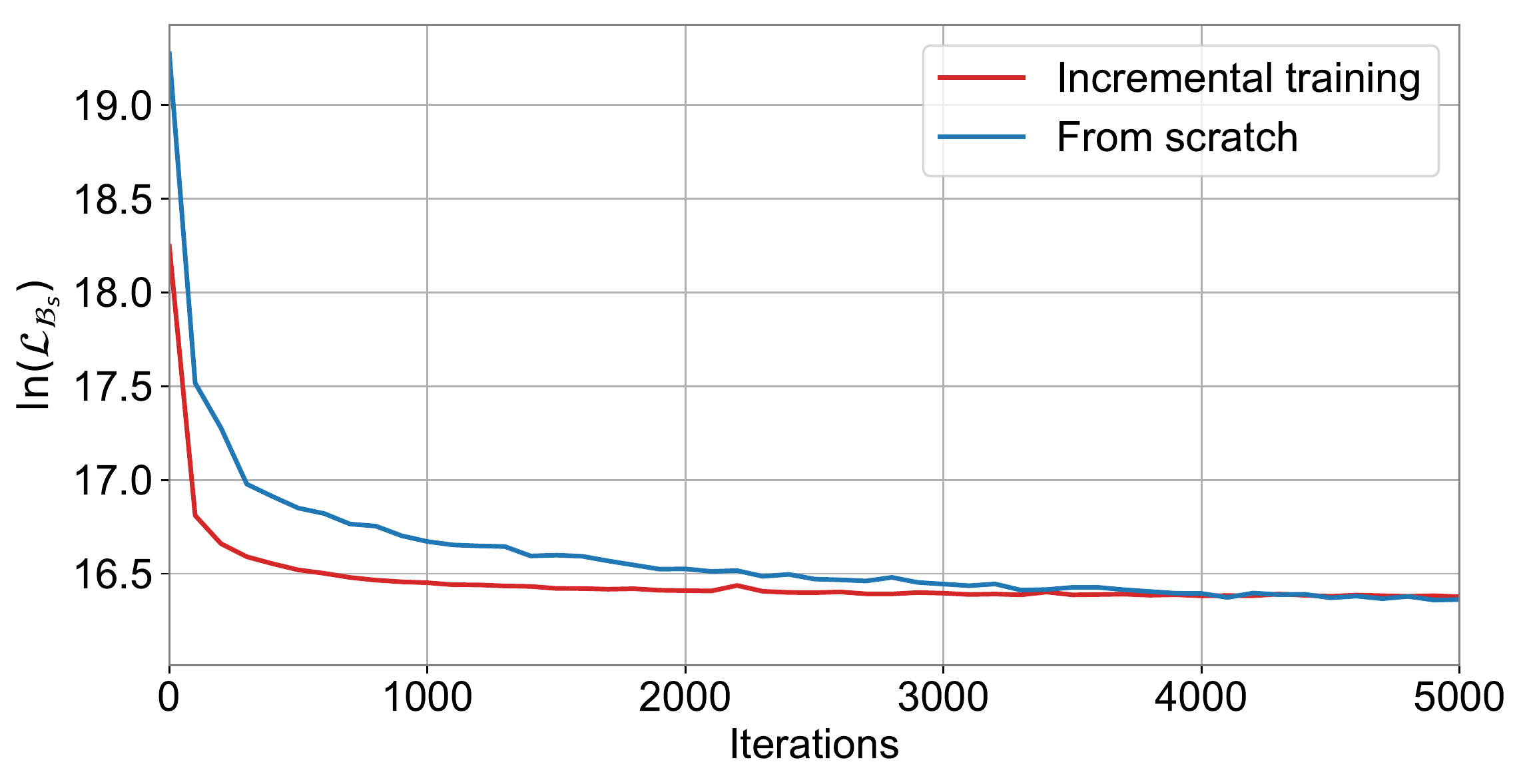}
\end{minipage}

\begin{minipage}{\textwidth}
\begin{tabular}{ccc}
\includegraphics[width=0.128\textwidth]{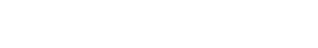}& \includegraphics[width=0.375\textwidth]{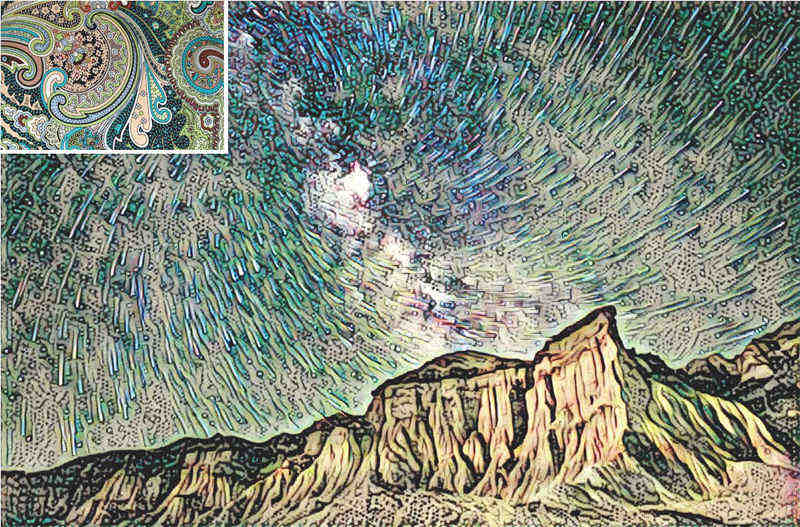} & \includegraphics[width=0.375\textwidth]{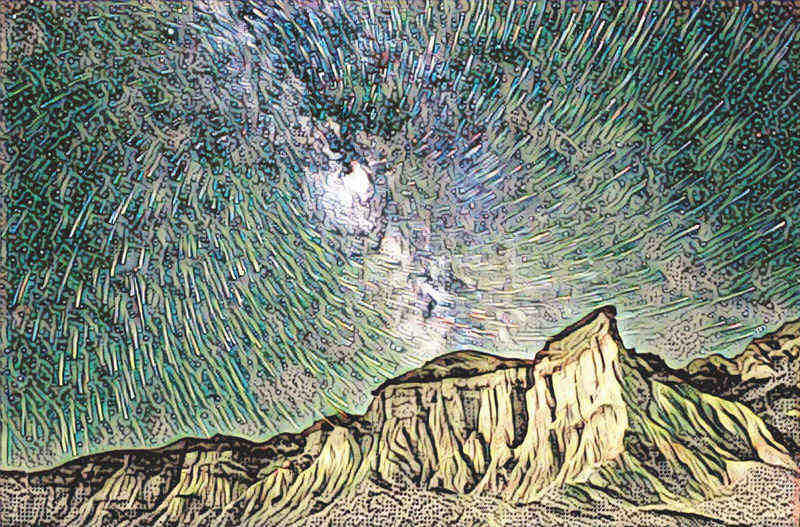}
\end{tabular}
\end{minipage}

\begin{minipage}{\textwidth}
\centering
%\smallskip
%\vspace{0.1cm}
\footnotesize{(b) Incremental training.}
\end{minipage}
\end{minipage}

\end{tabular}
\end{minipage}
%\vspace{-0.50cm}
\caption{Training curve comparisons of the training strategies. The bottom left and right images in (b) are the results of incremental training and training from scratch.}
%\myLcomment{draw the difference graph}
\label{fig:trainingcurve} %% label for entire figure
\end{figure}

%The bottom left image in (b) is the result of incremental training, and the bottom right image is the result of training from scratch.

For the quantitative evaluation, we focus on three evaluation metrics, which are: training curves during progressive training and incremental training; average content and style loss for test content images; training time for our single model and corresponding generating time for results with different stroke sizes.

\textbf{Training curve analysis.} To demonstrate the effectiveness of our progressive training strategy, we record the stroke losses when learning several sizes of strokes progressively and learning different strokes individually. The result is shown in Figure~\ref{fig:trainingcurve}(a). The reported loss values were averaged over 15 randomly selected batches of content images. It can be observed that the network which progressively learns multiple stroke sizes converges relatively faster than the one which learns only one single stroke size individually. The result indicates that during progressive training, the latter stroke branch benefits from the learned knowledge of the previous branches, and can even improve the training of previous branches through a shared network component in turn. To validate our stroke incremental training strategy, we present both the training curves of the incremental training and training from scratch in Figure~\ref{fig:trainingcurve}(b). While achieving comparable stylization quality, incrementally learning a stroke can significantly speed up the training process compared to learning from scratch.

\begin{figure}[!tp]
\setlength\tabcolsep{0.6 pt}
{\renewcommand{\arraystretch}{1}
%\begin{tabular}{>{\centering}n{\p} >{\centering}n{\p} >{\centering\arraybackslash}n{\p}}
\begin{tabular}{>{\centering}m{6.3cm}  >{\centering\arraybackslash}m{5cm}}
\centering

\includegraphics[width=0.35\textwidth]{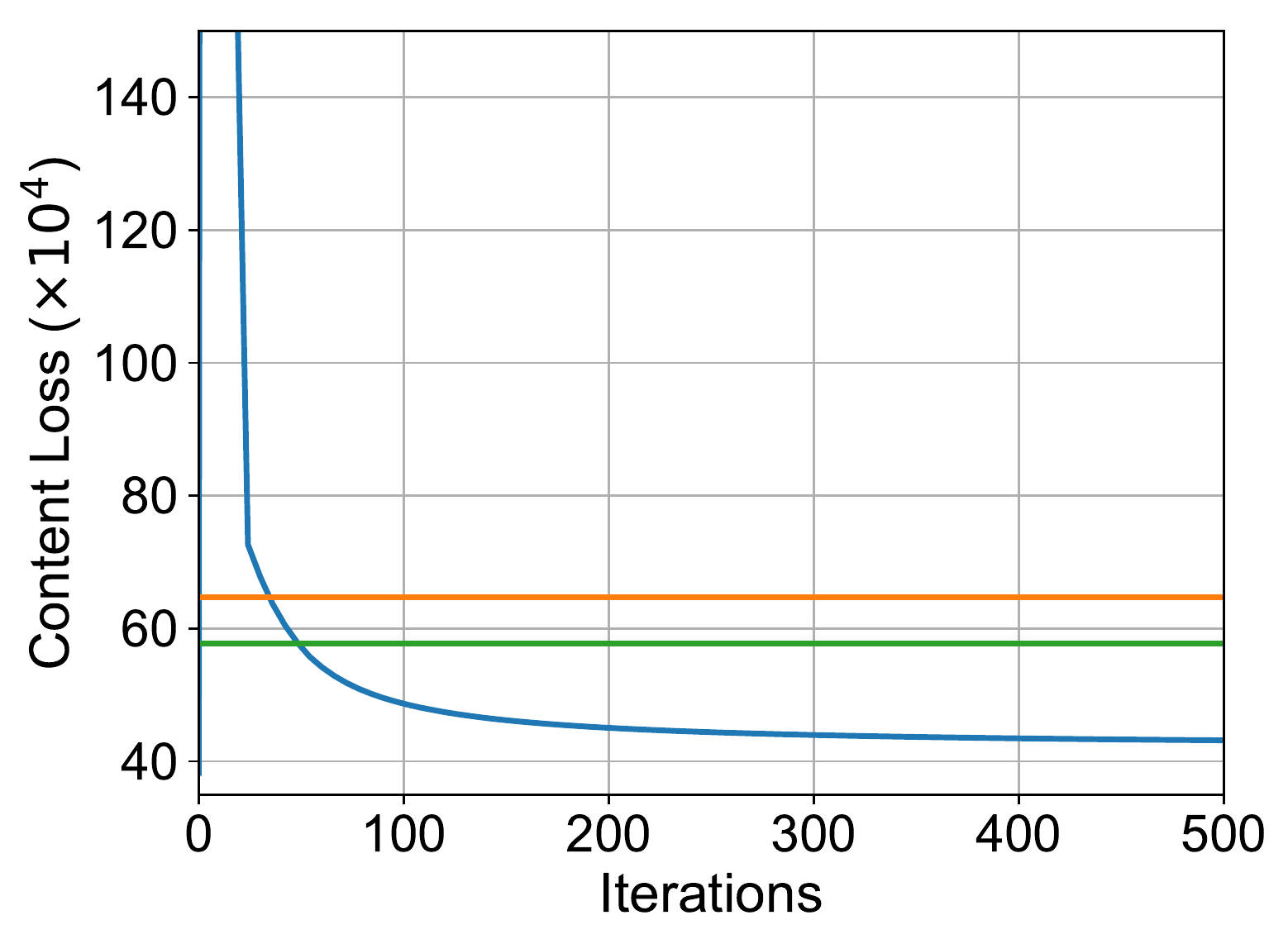}  & \includegraphics[width=0.35\textwidth]{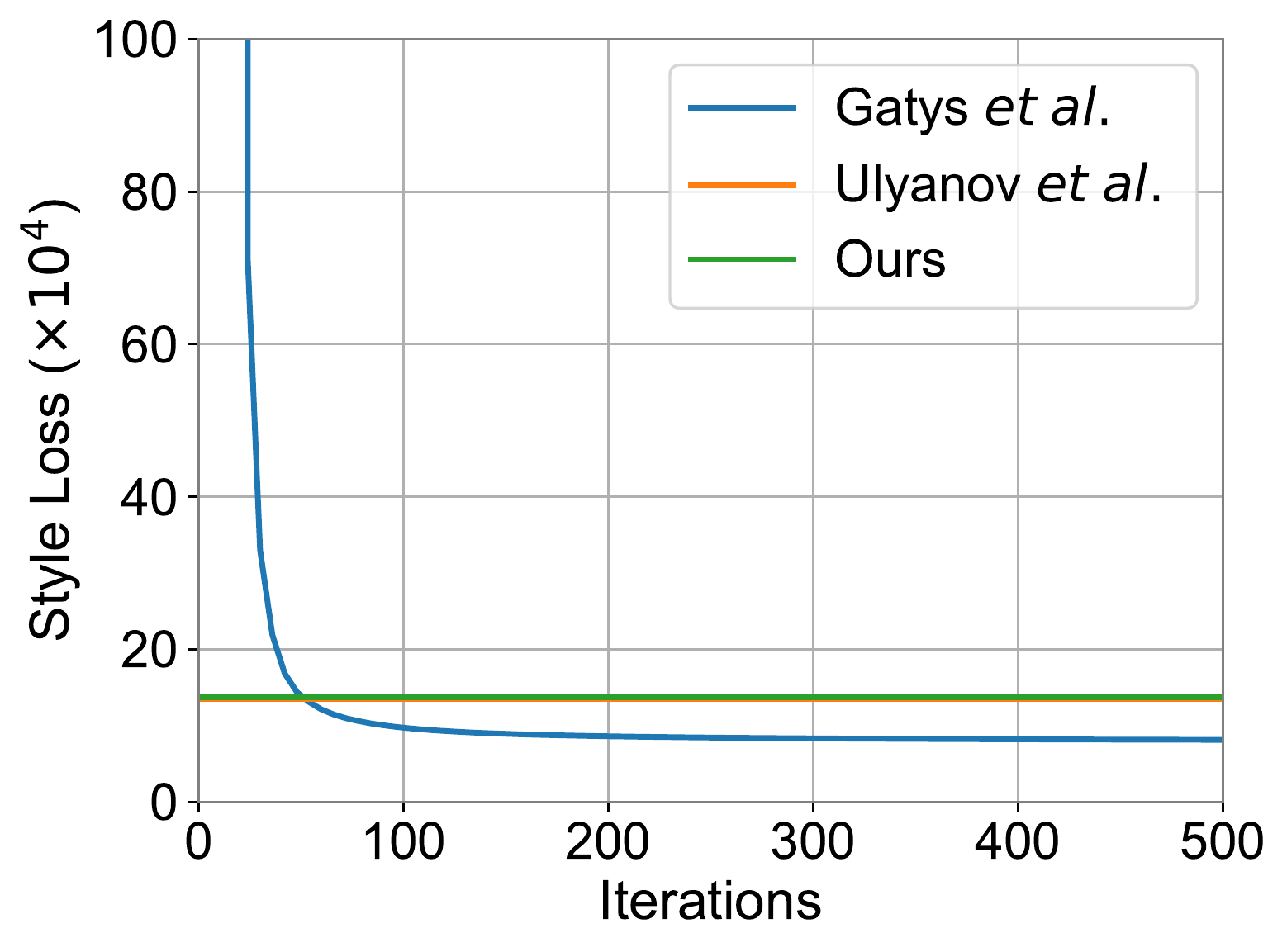}\\

\footnotesize{(a) Content loss} & \footnotesize{(b) Style loss} %\\
\end{tabular}
}
%\smallskip
  \caption{Comparisons of the average content and style loss of our algorithm with state-of-the-art Neural Style Transfer algorithms.}
  \label{fig:finalloss} %% label for entire figure
\end{figure}

\textbf{Average loss analysis.} To measure how well the loss function is minimized, we compare the average content and style loss of our algorithm with other style transfer methods. The recorded values are averaged over 100 content images and 5 style images. For each style, we calculate the average loss of the three stroke sizes. As shown in Figure~\ref{fig:finalloss}, the average style loss of our algorithm is similar to \cite{ulyanov2017improved}, and our average content loss is slightly lower than \cite{ulyanov2017improved}. This indicates that our algorithm achieves comparable or slightly better performance than \cite{ulyanov2017improved} regarding the ability to minimize the loss function.

\textbf{Speed and model size analysis.} Fully training one single model with three stroke sizes takes about 2 hours on a single NVIDIA Quadro M6000. For generating time, it takes averagely 0.09 seconds to stylize an image with size $1024 \times 1024$ on the same GPU using our algorithm. Since our network architecture is similar with \cite{Johnson2016perceptual,ulyanov2017improved} but with a shorter path for some stroke sizes, our algorithm can be on average faster than \cite{Johnson2016perceptual,ulyanov2017improved}, and further faster than Wang \etal's algorithm, Huang and Belongie's algorithm and Li \etal's algorithm according to the speed analysis in \cite{wang2016multimodal,huang2017arbitrary,li2017universal}. The size of our model on disk is 0.99 MB.

%The results of our continuous stroke size control are demonstrated in Figure~\ref{}.

\section{Discussion and Conclusion}
\label{sect:conclusion}

%In this paper, we introduce a fine and flexible stroke size control approach for Fast Style Transfer. Without trading off quality and efficiency, our algorithm is the first to achieve continuous and spatial stroke size control with one single model. Our experimental results demonstrate that by controlling the stroke size using our algorithm, AI-Created Art can be much closer to Human-Created Art. Our idea can also be directly applied to MSPM methods. For the application in the real world, our work provides a new tool for practitioners to inject their own artistic preferences into style transfer results, which can be directly applied in the production software and entertainment. Regarding the significance of our work for the larger vision community beyond style transfer, our work takes one step in the direction of learning adaptive receptive fields in the human vision system and primarily validates its significance in style transfer. In the future, we hope to further explore the use of learning adaptive receptive fields to benefit the larger vision community, \eg, multi-scale deep image aesthetic assessment, deep image compression, deep image colorization, \etc.

%Our experimental results demonstrate that by controlling the stroke size using our algorithm, AI-Created Art can be much closer to Human-Created Art.
%which can be directly applied in the production software and entertainment.
In this paper, we introduce a fine and flexible stroke size control approach for Fast Style Transfer. Without trading off quality and efficiency, our algorithm is the first to achieve continuous and spatial stroke size control with one single model. Our idea can also be directly applied to MSPM methods. For the application in the real world, our work provides a new tool for practitioners to inject their own artistic preferences into style transfer results, which can be directly applied in the production software and entertainment. Regarding the significance of our work for the larger vision community beyond style transfer, our work takes one step in the direction of learning adaptive receptive fields in the human vision system and primarily validates its significance in style transfer. In the future, we hope to further explore the use of learning adaptive receptive fields to benefit the larger vision community, \eg, multi-scale deep image aesthetic assessment, deep image compression, deep image colorization, \etc.

% in this field
%We believe t
Our work is only the first step towards the finer and more flexible stroke size control, and there are still some issues remaining to be addressed. The most interesting one is probably the automatic spatial stroke size control in one shot. The process of spatial stroke size control will be more efficient and user-friendly if the semantic segmentation network can be incorporated as a module in our network, so as to support the automatic determination of the stroke sizes for different spatial regions. Besides, the relations among the style representations of different scales of the same style image still remains unclear. The transformation from the style representation of one scale to that of another is the key to a more flexible stroke size control.
\\
\\
\textbf{Acknowledgments. }The first two authors contributed equally. Mingli Song is the corresponding author. This work is supported by National Key Research and Development Program (2016YFB1200203), National Natural Science Foundation of China (61572428, U1509206), Fundamental Research Funds for the Central Universities (2017FZA5014) and Key Research, Development Program of Zhejiang Province (2018C01004) and ARC FL-170100117, DP-180103424 of Australia.

\bibliographystyle{splncs04}
\bibliography{MYRE}
\end{document}